\documentclass[12pt,authoryear]{elsarticle}  

\usepackage[table]{xcolor}
\definecolor{lightgray}{gray}{0.93}

\usepackage[utf8]{inputenc}
\usepackage[T1]{fontenc}
\usepackage{lmodern}

\usepackage{geometry}
\usepackage{amsmath,amssymb}
\usepackage{mathrsfs}

\usepackage{subcaption}
\usepackage[font=scriptsize]{caption}
\usepackage{placeins}  

\usepackage{array}
\usepackage{booktabs}
\usepackage{multirow}
\usepackage{makecell}
\usepackage{tabularx}
\usepackage{threeparttable}
\usepackage{siunitx}
\usepackage{rotating}
\usepackage{pdflscape}

\usepackage{algorithm}
\usepackage{algpseudocode}

\usepackage{pgfplots}
\pgfplotsset{compat=newest}
\usepackage{tikz}
\usetikzlibrary{patterns,positioning}

\usepackage{hyperref}
\hypersetup{colorlinks=true, linkcolor=blue, citecolor=blue, urlcolor=blue}

\journal{Expert Systems with Applications}

\begin{document}
	
	\begin{frontmatter}
		
		\title{Hybrid Imbalanced Regression Through Unified Data-Level and Algorithm-Level Balancing}
		
		\author[label1]{Shermin Shahbazi}
		\ead{sh.shahbazi@znu.ac.ir}
		
		\author[label1]{Hossein Mohammadi\corref{cor1}}
		\ead{hosm@znu.ac.ir}
		
		\author[label1]{Mohsen Afsharchi}
		\ead{afsharchi@znu.ac.ir}
		
		\affiliation[label1]{organization={Department of Computer Engineering, University of Zanjan},
			state={Zanjan},
			country={Iran}}
			
		\cortext[cor1]{Corresponding author: Hossein Mohammadi (email: hosm@znu.ac.ir)}

		\begin{abstract}
			Imbalanced learning is a critical challenge in machine learning, where underrepresented target values can lead to biased models that underperform on rare but high-impact phenomena. While extensively studied in classification tasks with skewed class distributions, imbalanced regression (characterized by non-uniform densities in continuous target spaces) remains underexplored. Existing works in imbalanced regression predominantly focus on either \emph{data-level} methods, which synthesize samples to enrich sparse regions but risk introducing noise or overfitting, or \emph{algorithm-level} methods, which adjust loss functions to prioritize minority areas but often falter with complex distributions and severe imbalances, thereby neglecting the complementary advantages of the other approach. To address this gap, we propose a \emph{hybrid framework} that synergistically integrates data- and algorithm-level strategies into a unified, end-to-end pipeline. This regressor-agnostic framework employs a five-phase workflow: \emph{adaptive bin partitioning} to emulate class-like structures by dynamically segmenting the target space based on local linear coherence; target-conditioned \emph{representation learning} via a Conditional Variational Autoencoder for enhanced latent modeling; a novel multistage \emph{data-level balancing} that incorporates feature-level clustering followed by oversampling in minority clusters to minimize bias; an \emph{algorithm-level balancing} with a new Latent-Density Weighted Loss (LDWL) to emphasize joint rarity in latent and target spaces; and attention-based gated \emph{fusion} to integrate balanced representations for final regression. Empirical evaluations demonstrate that integrating this framework upstream of various regressors, enhances predictive performance compared to standalone regressors.
		\end{abstract}
		
		\begin{highlights}
			
			\item Proposes a hybrid framework that combines data-level and algorithm-level methods to tackle imbalanced regression problems effectively.
			\item Features adaptive bin partitioning to dynamically segment continuous target spaces based on local data patterns.
			\item Introduces multistage balancing with feature clustering and oversampling to enrich sparse regions without adding noise.
			\item Develops a new latent-density weighted loss to prioritize rare samples in both latent and target spaces.
			\item Demonstrates improved predictive performance on 16 benchmark datasets, outperforming standalone approaches.
			
		\end{highlights}
		
		
	\end{frontmatter}
	
	\section{Introduction}
	\label{sec:Introduction}
	Imbalanced learning refers to a class of machine learning problems in which certain target values are significantly underrepresented compared to others \citep{Chen2024, REZVANI2023110415}. While this challenge has been extensively addressed in classification (typically involving skewed class distributions), it is regarded differently in regression, where the target variable is continuous. In this context, imbalance is characterized by a non-uniform density of samples across the continuous target spectrum, with specific regions (particularly extreme or rare target values) being sparsely populated \citep{Ribeiro2020, BELHAOUARI2024}. Consequently, some ranges of the target variable, often associated with rare but critical phenomena, may be severely underrepresented or nearly absent from the training data \citep{DOLAR2025tuba, ZHU2019140}.
	
	This skewed distribution introduces unique challenges for learning algorithms. Standard regression models, driven by global error minimization, tend to prioritize the majority or over-represented regions of the target space, leading to poor performance in low-density minority regions \citep{Kaur2019Har}. Such limitations are particularly problematic in real-world domains where rare outcomes are of high importance \citep{Ribeiro2020}, including extreme weather forecasting \citep{ZHEN2025120952, Scheepens2023}, anomaly detection \citep{2017Candelieri, Hajjami2020}, and rare event prediction in medical \citep{Miao2017ieee, YEW2025min} or financial domains \citep{2020Lucas, LI2025Financial}. In these applications, the primary objective is often to accurately predict rare or extreme cases which, although rare, are often of paramount practical importance. As a result, conventional models often fail to generalize in these critical regions, highlighting the necessity for specialized regression techniques that explicitly address imbalance in continuous targets.

	Addressing the challenges of imbalanced regression outlined above, the field of imbalanced learning emerged in the late 1990s \citep{kubat1997addressing}, initially tackling skewed class distributions in classification  \citep{Chen2024, REZVANI2023,  Goswami2024}. Over the past decade, focus has slightly shifted to regression, where continuous target imbalances pose distinct challenges, yet this area remains scarcely investigated. A closer examination of the literature reveals two primary approaches: \emph{data-level methods}, which use resampling to synthesize samples or modify datasets to balance underrepresented minority regions, and \emph{algorithm-level methods}, which preserve data originality by adjusting loss functions to prioritize rare target ranges \citep{Chen2024, Aguiar2024, Kaur2019Har}. Data-level methods are intuitive, model-agnostic, and enhance dataset diversity, improving robustness in sparse regions, but risk introducing synthetic noise or overfitting. Algorithm-level approaches offer flexible, principled loss reweighting without oversampling artifacts, yet struggle with complex distributions and severe imbalances. Individually, these methods are insufficient, as resampling lacks algorithmic guidance, and loss adjustments falter without balanced data. Although data-level and algorithm-level methods are expected to complement and reinforce each other, their tight integration into truly hybrid approaches, to the best of our knowledge has received almost no attention in imbalanced regression. Whereas a few studies have combined resampling with ensemble strategies \citep{IRDA2024, Camacho2024, Chen2019Lalor, 2018REBAGG}, none has integrated data-level sample synthesis with explicit, sample-wise cost-sensitive or reweighting losses during training. To fill this gap, we propose an end-to-end hybrid framework that tightly unifies data-level and algorithm-level methods across five phases:
		
	\begin{itemize}
		\item \emph{Phase 0: Adaptive Target-Space Bin Partitioning}, which emulates class-like structures by adaptively segmenting the continuous target space using data-driven $R^2$ coherence scores to quantify local linear consistency, enabling precise imbalance identification across diverse data complexities.
		
		\item \emph{Phase I: Representation Learning}, which employs a Conditional Variational Autoencoder (CVAE), conditioned on the target variable, to project data into a latent space that captures predictive patterns for underrepresented regions.
		
		\item \emph{Phase II: Data-Level Balancing}, which balances label ($Y$) and feature ($X$) spaces through a multistage strategy to augment underrepresented regions while minimizing bias.
		
		\item \emph{Phase III: Algorithm-Level Balancing}, which applies a Latent-Density Weighted Loss (LDWL) to prioritize minority samples using joint density weighting in latent and target spaces.
		
		\item \emph{Phase IV: Feature Fusion}, which integrates balanced latent representations via attention-based Gated Fusion to ensure robust predictions across all target regions.
	\end{itemize}
	
	The key contributions are the four novel components above (Phases 0--III); Phase IV provides principled fusion. The proposed hybrid framework ultimately yields a feature vector that simultaneously benefits from data- and algorithm-level balancing, ready for application by any regressor. This study sets out to investigate the following research questions:
	
	\begin{itemize}
		\item \textbf{RQ1:} Does the integration of data-level and algorithm-level balancing strategies within a unified hybrid framework improve predictive performance and generalization in imbalanced regression tasks compared to using either strategy in isolation?
		
		\item \textbf{RQ2:} How does the proposed hybrid pipeline perform in large-scale imbalanced regression compared to standard regression baselines without imbalance handling?
		
		\item \textbf{RQ3:} How does the degree and nature of target variable skewness across datasets influence the effectiveness and robustness of the proposed hybrid pipeline?
		
		\item \textbf{RQ4:} What are the limitations of the proposed hybrid framework?
		
		\item \textbf{RQ5:} How does each phase of the proposed framework (adaptive bin partitioning, representation learning, data-level balancing, algorithm-level balancing, and fusion) contribute to overall performance in large-scale imbalanced regression?
	\end{itemize}
	
	The remainder of this paper is organized as follows. Section~\ref{sec:literature-review} reviews the relevant literature on imbalanced learning in regression, categorizing existing methods into data-level, algorithm-level, and hybrid approaches. Section~\ref{sec:methods} presents the proposed five-phase hybrid framework in detail, outlining each component from adaptive bin partitioning to final regression, with time complexity analyses for each phase provided to clarify their computational demands. Section~\ref{sec:resultAndDiscussion} reports the experimental results, including comparisons with baseline models and ablation studies to assess the contribution of each phase. Finally, Section~\ref{sec:conclusion} concludes the paper and outlines directions for future research. Additional analyses are provided in the appendix.

	\section{Literature Review}
	\label{sec:literature-review}
	Imbalanced learning has been widely studied in classification, where discrete class labels enable clear distinction of majority and minority classes. This has concluded a rich ecosystem of strategies, including resampling, cost-sensitive learning, and ensemble methods \citep{REZVANI2023, Chen2024, Goswami2024, ArafIm2024}. In contrast, imbalanced regression (where the target is continuous and lacks natural class boundaries) remains significantly underexplored, despite its critical role in domains such as extreme event prediction, anomaly detection, and precision engineering. This review focuses on imbalanced regression and organizes prior work into three primary categories:  \emph{Data-level balancing}, which rebalance the training distribution via resampling or sample generation; \emph{Algorithm-level balancing}, which modify the loss function, model architecture, or optimization process to prioritize rare target regions; and \emph{Hybrid approaches}, which combine data- and algorithm-level strategies for complementary benefits. Each category is further subdivided by core mechanism.

	\subsection{Data-Level Balancing}
	\label{sec:data-level-balancing-techniques}
	\emph{Data-level methods} (also known as \emph{preprocessing}) address imbalance by directly modifying the training data distribution prior to upstream model learning. These approaches operate under the principle that, in data-driven systems, the quality and representativeness of the input data are foundational to predictive performance \citep{Avelino2024, Carvalho2025, GONG2023Yoidu}. This is more challenging in regression, where minority regions are continuous and not easily distinguishable \citep{BELHAOUARI2024}. The core idea is to restructure the dataset into a more balanced form through resampling strategies, either by \emph{oversampling} underrepresented regions, \emph{undersampling} dominant ones, or employing \emph{combinations of both}.

	\subsubsection{Oversampling}
	Oversampling methods aim to alleviate regression imbalance by increasing the sample density in underrepresented (minority) regions of the target space. While effective on small datasets, these methods struggle to scale to large, complex target distributions without adaptive binning and structural validation. The central objective is to synthetically enrich these sparse areas until they become statistically comparable to majority regions, thus reducing learning bias and enhancing model generalization \citep{BELHAOUARI2024}. These techniques focus explicitly on the \textit{minority regions}, ensuring that critical yet under-sampled patterns are adequately represented during training. From the literature, oversampling methods for imbalanced regression can be categorized into the following groups:

	\paragraph{A. Synthetic Minority Oversampling} these methods adapt the core idea of \textit{SMOTE} \citep{chawla2002smote} from classification to regression by generating synthetic samples in sparse target intervals via interpolation between a sample and its nearest neighbors in feature space. Target values are computed using smooth interpolation or noise-augmented local modeling to preserve statistical coherence and continuity, thereby increasing density in underrepresented regions without discrete class boundaries. The \textit{SMOGN (Synthetic Minority Oversampling Technique for Regression with Gaussian Noise)} algorithm \citep{SMOGN2017} extends SMOTE by identifying rare target regions using a relevance function, synthesizing features through neighbor-based interpolation, and generating targets via linear interpolation augmented with Gaussian noise, ensuring smooth transitions, natural variability, and robustness against overfitting. In another study, the authors proposed \textit{SMOTER} \citep{SmoteR2013} that adapts SMOTE for regression by using a relevance function to map continuous targets onto a scale that identifies rare (often extreme) regions for oversampling. Synthetic samples are generated via interpolation between instances of similar relevance, preserving smooth target variation. This relevance-driven approach offers precise control over which target regions are emphasized, making it particularly effective for tasks requiring accurate prediction of extreme values. 	\textit{WSMOTER} \citep{Camacho2024} extends SMOTER by incorporating distance- and variance-based weighting during SMOTE-style interpolation, enabling finer control over the relevance and quality of synthetic samples in sparse target regions. This emphasizes informative neighbors, reduces noise, and improves generalization in minority intervals. Similarly, \textit{Importance-SMOTE} \citep{ImportanceSMOTE2022} enhances robustness to noise by selectively oversampling only high-relevance, high-density minority instances, using local density and relevance scores to avoid amplifying unreliable or borderline points. Both methods advance data-level balancing by prioritizing quality and informativeness in sample generation, making them particularly effective in noisy, highly skewed regression settings.

	\paragraph{B. Generative Deep Learning-Based Methods} unlike classical SMOTE-style approaches that rely on local interpolation between neighboring samples in the feature space, generative model-based methods, such as Generative Adversarial Networks (GANs) and Variational Autoencoders (VAEs), learn the underlying data distribution using deep architectures. By sampling from a latent space, they produce diverse and coherent synthetic samples, better capturing nonlinear patterns and effectively addressing sparsity in imbalanced regression. In an empirical study the authors propose a CVAE conditioned on rare instances (identified via relevance function) to generate statistically valid samples, enhancing learning in sparse regions without distorting normal data \citep{HUANG2022}. In a similar experiment, the authors use autoencoders to learn the data manifold, generating synthetic samples via latent-space interpolation and decoding to preserve global structure and local smoothness, outperforming SMOTE variants \citep{BELHAOUARI2024}. In another investigations, the authors combine VAEs with smoothed bootstrapping to improve oversampling on tabular data, with strong empirical results \citep{Stocksieker2024, Ohno2020, TIAN2023119157}. Though GANs are widespread in classification, their use in regression is emerging \citep{2024Hssayeni, xu2019modeling, Ohno2020}. These methods use adversarial training (a generator creates realistic rare samples, while a discriminator ensures they are faithful to the true data) to better capture complex minority distributions.

	\paragraph{C. Other Oversampling Methods} beyond SMOTE and deep generative models, alternative oversampling techniques use statistical, distributional, or noise-based mechanisms to create flexible, often model-agnostic synthetic samples. \emph{Non-parametric methods} like Kernel Density Estimation (KDE) \citep{Steininger2021, Zhang2025yu} model the data distribution to identify low-density regions and sample new instances from the estimated density, adapting well to complex or multimodal patterns. \emph{Noise-based approaches} \citep{BRANCO201976, FENG2025103192} add controlled Gaussian or uniform perturbations to minority samples, efficiently increasing local density while preserving structure. \emph{Parametric fitting methods} \citep{PAN20201214, Scrucca2025} fit distributions like Gaussian Mixture Models (GMMs), to rare regions and sample accordingly, offering interpretability but assuming a specific form. Hybrid strategies are also appearing: in a study the authors combine Random Oversampling (ROS) with stacking ensembles, showing that simple rebalancing can significantly boost performance in imbalanced regression \citep{Avelino2025}.

	\subsubsection{Undersampling}
	Undersampling reduces majority dominance by selectively removing samples from high-density target regions, thereby rebalancing the distribution and sharpening focus on rare values \citep{aleksic2025}. Unlike oversampling, it prunes redundant majority instances rather than synthesizing new ones. Methods are classified as \emph{random} (indiscriminate removal) or \emph{informed} (e.g., Selective Undersampling-SUS), which use relevance functions and density scoring to retain only the most valuable majority samples.

	\paragraph{A. Random Undersampling} random undersampling indiscriminately removes majority samples to balance the distribution, offering simplicity but risking the loss of informative instances \citep{Komsrimorakot2025, YangCy2024}. Though widely used in classification \citep{Komsrimorakot2025, TAHIR20123738}, it has (to the best of our knowledge) not been adopted in imbalanced regression due to the field’s limited attention and the method’s neglect of data relevance.
	
	\paragraph{B. Informed Undersampling} 
	In contrast, informed undersampling aims to preserve valuable training information by selectively discarding only the least relevant or most redundant majority samples based on statistical or relevance-driven criteria. Informed undersampling has been explored through several strategies. Examples include SUS \citep{aleksic2025}, which uses relevance- and density-based scores to preserve critical samples; an approach that employs mutual class potential functions to guide radial pruning of majority instances \citep{KOZIARSKI2}; and a method that constructs spherical neighborhoods to retain local structural patterns while reducing the majority class \citep{yan2023Yuan}

	\subsubsection{Combined Over/Under sampling}
	Combined over/undersampling integrates the strengths of both approaches (enriching minority regions and selectively pruning redundant majority samples) to achieve better balance and generalization in imbalanced regression. Research in this area remains scarce. A notable exception is \emph{WERCS} \citep{BRANCO201976}, which uses a relevance function to assign sampling probabilities, enabling targeted duplication or removal of instances across the continuous target space without binning or thresholds.

	\subsection{Algorithm-Level Balancing}
	\emph{Algorithm-level} (also known as \emph{model-level} or \emph{cost-sensitive}) methods address imbalance by adapting the learning objective, loss function, or architecture (rather than modifying the data) to prioritize underrepresented target regions \citep{Araf2024Ima, fernandez2018learning}. These approaches embed imbalance awareness directly into training and fall into three main categories: \textit{cost-sensitive and loss reweighting}, \textit{ensemble learning}, and \textit{distribution-aware methods}.

	\subsubsection{Cost-Sensitive and Loss Reweighting}
	Cost-sensitive and loss reweighting methods address imbalance by prioritizing rare samples during training, without modifying the data. \emph{Cost-sensitive} approaches impose higher penalties for errors in minority regions, while \emph{loss reweighting} adjusts sample contributions based on relevance, rarity, or density using heuristic or statistical criteria. \textit{IRDA} \citep{IRDA2024} combines relevance-driven loss reweighting with embedded data augmentation in deep networks, implicitly prioritizing rare samples without binning or synthetic generation. In a similar experiment, the authors introduce a cost-sensitive loss that boosts rare-sample influence using label-feature consistency for robust machinery health prognosis \citep{CAO2024123930}. In another study, the authors applies target-aware reweighting with deep feature transfer to penalize errors on low-RUL (remaining useful life) instances, enhancing generalization in bearing life prediction \citep{DING2022ff}.
	
	Several advanced reweighting and regularization methods prioritize rare targets during training. \textit{DenseWeight} \citep{LI2023119541} uses KDE to assign higher weights to low-density (minority) samples in a model-agnostic way. \textit{Distribution Smoothing Loss (DSL)} \citep{2021Delving} smooths the empirical target distribution via a regularization loss to improve generalization across imbalanced regions. \textit{RankSim} \citep{RankSim2022} enforces input-output ranking consistency to preserve structural relationships under skew. Building on this, \textit{Rank-N-Contrast} \citep{Kaiwen2023zha} applies contrastive learning with ranking and distributional constraints to better focus on continuous, underrepresented targets in regression.

	\subsubsection{Ensemble Learning}
	Ensemble learning improves robustness in imbalanced regression by combining predictions from multiple base learners, often trained on resampled or weighted subsets, without modifying the base loss function \citep{KHAN2024122778, 2017Evaluation}. This promotes diversity and reduces majority overfitting. Methods fall into three classes: \emph{bagging} reduces variance via balanced bootstrapping; \emph{boosting} sequentially emphasizes poorly predicted (typically minority) instances; and \emph{stacking} aggregates diverse models using a meta-learner, optionally with relevance-aware weighting \citep{zhou2025ensemble, Mor2012Mendes}. \textit{REBAGG (REsampled BAGGing)} \citep{2018REBAGG} adapts bagging for skewed continuous targets by resampling subsets using relevance-weighted probabilities to prioritize rare values, thereby improving ensemble focus and performance on underrepresented target regions.
	
	In an investigation into boosting-based ensembles, the authors proposed a Random Forest and Least Squares Boosting hybrid with Bayesian tuning to improve extrapolation in sparsely sampled regions of 5G throughput prediction \citep{Isabona2022}. In another study, the authors introduced \textit{SMOTEBoost}, which applies SMOTE before each boosting iteration to augment rare samples, enhancing focus on extreme target values without removing majority instances \citep{2018MonizSMOTEBoost}. Additionally, in another experiment the authors evaluated bagging and boosting on benchmark imbalanced regression tasks, showing that relevance-aware resampling significantly improves performance in underrepresented target regions \citep{2017Evaluation}.
	
	In a recent work, the authors proposed a stacking-based density estimation approach with an oversampling variant to enhance prediction performance in rare target regions \citep{Xin2025Stacking}. Using heterogeneous ensembles to estimate density and generate synthetic samples, SDE-OS outperforms kernel density and SMOGN-based approaches. In another contribution, the researchers presented a hybrid ensemble that first employs a classifier to detect high-relevance (minority) samples, then trains a dedicated regressor on these instances, improving accuracy in sparse regions without altering the data distribution \citep{2023vchdteg}.

	\subsubsection{Distribution-Aware Techniques}
	Distribution-aware techniques, a subset of algorithm-level methods, model target or feature distributions to prioritize sparse regions without altering the data. Unlike cost-sensitive reweighting or ensemble aggregation, they use statistical properties (such as label or feature density) to smooth predictions across the target space. For example, \emph{Label Distribution Smoothing (LDS)} and \emph{Feature Distribution Smoothing (FDS)} adjust outputs to respect target continuity, enhancing performance in minority regions, though typically requiring deep architectures. In a very recent study, a distribution-aware strategy for predicting continuous refractive error from retinal images was introduced, combining \emph{LDS} and \emph{FDS} within a deep regression model to enhance generalization in rare target ranges \citep{YEW2025min}. In another investigation, the hybrid use of \emph{LDS} and \emph{FDS} was proposed to address long-tailed regression in vision, language, and healthcare, improving age estimation without dataset or loss modification \citep{2021Delving}. In a similar experiment, the authors presented \textit{SMOTEBoost} for Regression (\textit{SB-R}), integrating SMOTE-like oversampling into boosting iterations to dynamically balance the target space and boost sensitivity to extreme values \citep{2018MonizSMOTEBoost}. In their excellent study, the authors proposed a contrastive framework preserving ranking and distributional continuity in image-based regression, promoting robust representations of rare targets \citep{Kaiwen2023zha}. Another research, introduced an adaptive loss adjustment prioritizing extreme values to improve minority region performance in imbalanced regression \citep{Ribeiro2020}.

	\subsection{Hybrid Approaches}
	\label{sec:Hybrid-Approaches}
	Hybrid approaches in imbalanced regression integrate data-level and algorithm-level techniques to better handle skewed target distributions \citep{Chen2024, fernandez2018learning}. These methods combine synthetic sample generation, relevance-aware resampling, or distribution-aware augmentation with customized loss functions, reweighting, or model adjustments. By simultaneously improving data representation and learning dynamics, such frameworks provide a more comprehensive solution for enhancing prediction of rare but critical target values. Hybrid approaches are well-explored in imbalanced classification \citep{KHAN2024122778, Dablain2024, Sun2018Hui, SAGLAM2022117023}, yet remain extremely rare in regression. The few existing regression studies combine data-level methods only with ensemble techniques, excluding cost-sensitive or loss-reweighting strategies. To our knowledge, no prior work in imbalanced regression has integrated data-level oversampling with an explicit, sample-wise cost-sensitive or reweighting loss applied during gradient-based training of a single neural regressor. Existing hybrid attempts either rely on ensemble strategies \citep{Avelino2025,2018REBAGG} or are limited to non-tabular domains \citep{IRDA2024,Chen2019Lalor}.
	
	In a more recent study, the authors introduced a stacking-based ensemble for imbalanced regression, where diverse base learners are trained on oversampled data to better capture minority regions, and a meta-learner combines their predictions without modifying individual loss functions \citep{Avelino2025}. In another contribution, the authors proposed \emph{REBAGG (REsampled BAGGing)}, a bagging variant that resamples subsets using relevance-weighted probabilities to prioritize rare target values, thereby enhancing ensemble performance in underrepresented regions of continuous target spaces \citep{2018REBAGG}. Beyond tabular data, hybrid approaches have been applied to vision and clinical natural language processing.  \emph{IRDA (Implicit Reweighting with Data Augmentation)} \citep{IRDA2024} was introduced for age and head pose estimation, combining relevance-driven reweighting (algorithm-level) with mixup-based augmentation (data-level) to improve representation of rare image targets while preserving generalization. In another clinical text analysis, the researchers proposed a hybrid method for hypoglycemia detection, pairing SMOTE oversampling of rare cases with cost-sensitive support vector machine (SVM) to increase sensitivity to critical events in highly imbalanced patient messages \citep{Chen2019Lalor}.

	\subsection{Research Gap}
	\label{subsec:research_gap}
	As reviewed in \hyperref[sec:literature-review]{Literature Review}, to the best of our knowledge no work has combined data-level methods with algorithm-level methods that directly modify learning. Additionally, adaptive binning responsive to both feature and label distributions, along with targeted representation learning for sparse regions, remains unexplored. Our framework fills these gaps via a unified hybrid pipeline featuring adaptive target-space bin partitioning, CVAE-based representation learning, multistage data-level balancing with feature clustering and minority oversampling, and Latent Density Weighted Loss (LDWL) with gated fusion, delivering robust performance in imbalanced regression.

	\section{Methods}
	\label{sec:methods}

	\subsection{Datasets}
	We evaluate the proposed hybrid framework on widely used benchmark regression datasets commonly adopted in imbalanced learning studies, as summarized in Table \ref{tab:datasets}. These datasets span diverse domains (housing, energy, materials, finance, etc.) and exhibit significant target skewness and imbalance, making them ideal for assessing imbalanced regression methods. The detailed target distribution plots are provided in \ref{appen:datasets-target-lable-distrib}.
	
	\begin{table}[h!]
		\tiny
		\centering
		\caption{List of benchmark regression datasets used in this study and their basic characteristics.}
		\label{tab:datasets}
		\begin{tabular}{lcc}
			\hline
			\textbf{Dataset} & \textbf{No. of Samples} & \textbf{No. of Features} \\
			\hline
			california & 20,640 & 8 \\
			compactive & 8,192 & 21 \\
			cpu\_small & 8,192 & 12 \\
			heat & 7,400 & 11 \\
			wine\_quality & 4,898 & 11 \\
			abalone & 4,177 & 8 \\
			space\_ga & 3,107 & 6 \\
			debutanizer & 2,394 & 7 \\
			available\_power & 1,802 & 15 \\
			maximal\_torque & 1,802 & 32 \\
			fuel\_consumption\_country & 1,764 & 37 \\
			acceleration & 1,732 & 14 \\
			airfoild & 1,503 & 5 \\
			mortgage & 1,049 & 15 \\
			treasury & 1,049 & 15 \\
			concreteStrength & 1,030 & 8 \\
			\hline
		\end{tabular}
	\end{table}

	\subsection{Framework Architecture}
	To bridge the gaps identified in the \hyperref[subsec:research_gap]{Research Gap} section and to address the research questions stated in the \hyperref[sec:Introduction]{Introduction}, we propose a five-phase interconnected architecture, illustrated in Fig. \ref{fig:general_Architecture}. This framework is designed to demonstrate that a hybridization of data-level with algorithm-level methods  can yield substantial performance gains in imbalanced regression. Each phase is designed with a distinct objective, collectively contributing to a more balanced and robust regression model:
	
	\begin{figure}[ht]
		\centering
		\includegraphics[width=1.0\textwidth]{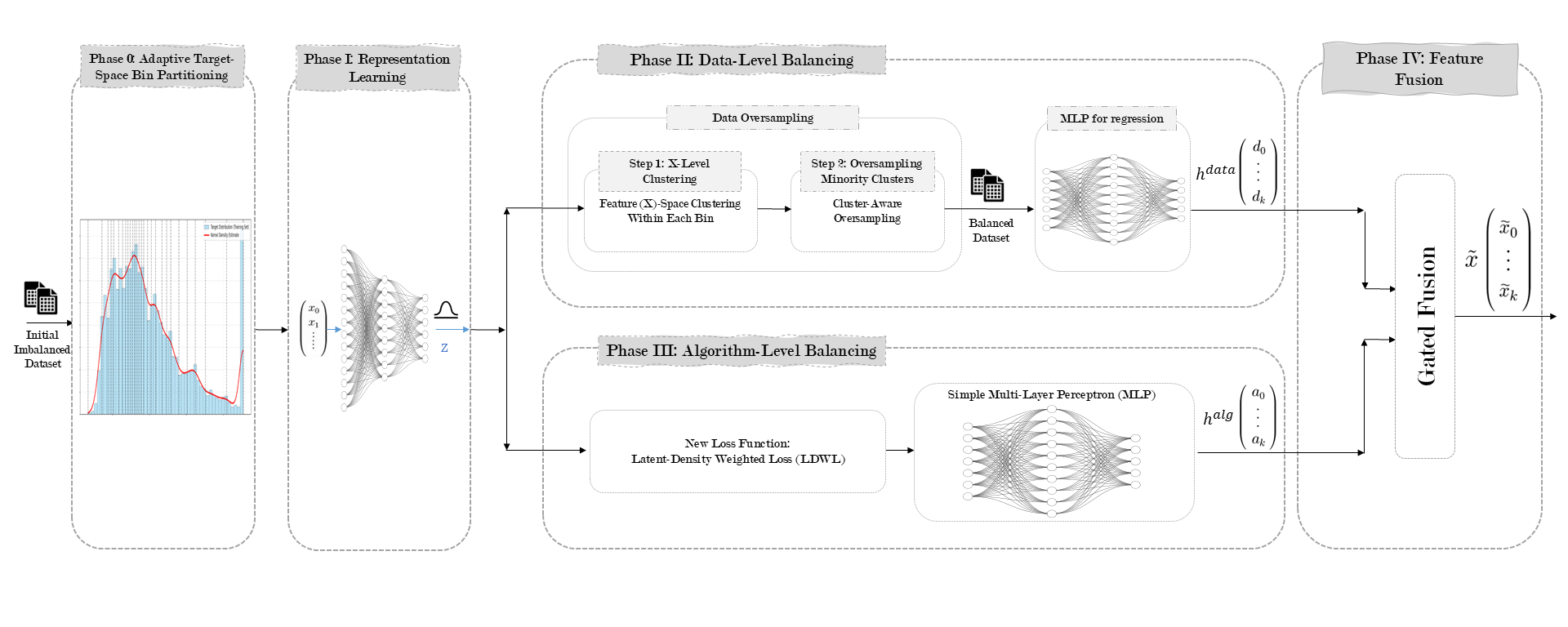} 
		\caption{The general architecture of the proposed imbalanced learning system}
		\label{fig:general_Architecture}
	\end{figure}

	\begin{itemize}
		\item \textbf{Phase 0 (Adaptive Target-Space Bin Partitioning)\footnote{Designated as Phase 0 to reflect its role as a preprocessing step that transforms the continuous target into a discrete, class-like structure before the main pipeline begins.}} adaptively transforms the continuous regression problem into a class-like structure by dividing the target space into statistically significant bins, enabling effective imbalance handling across diverse regions of the target distribution, similar to classification tasks.
		
		\item \textbf{Phase I (Representation Learning)} establishes the system’s foundation by learning expressive data representations to capture meaningful patterns, enhancing balancing and regression in later phases.
		
		\item \textbf{Phase II (Data-Level Balancing)} introduces a multi-step strategy that jointly balances label ($Y$) and feature ($X$) spaces. It adaptively analyzes target distribution, clusters features, and oversamples underrepresented regions, enhancing data diversity while preserving original structure and minimizing synthetic bias.
		
		\item \textbf{Phase III (Algorithm-Level Balancing)} focuses on adapting the learning algorithm to address imbalances without modifying the data, ensuring data integrity in domains where preserving the original distribution is critical.
		
		\item \textbf{Phase IV (Final Fusion)} fuses data- and algorithm-level balancing into a unified pipeline, leveraging their complementary strengths to produce a hybrid-enhanced representation vector ready for use by any regressor.
		
	\end{itemize}

	\subsection{Phase 0: Adaptive Target-Space Bin Partitioning}
	\label{sec:phase0}
	In imbalanced regression, where the target variable $y \in \mathbb{R}$ is continuous, the absence of discrete class labels complicates both balancing and minority region identification. To overcome this challenge, we introduce \emph{Adaptive Target-Space Bin Partitioning}, a preprocessing phase that emulates class-like structure by segmenting the target space into statistically coherent bins. This transformation enables later phases to operate with localized awareness of sparsity, thus improving performance in underrepresented regions. The term “\emph{adaptive}” signifies that the binning is not fixed-width or quantile-based, but rather \emph{data-driven}. Crucially, this phase serves as the foundation for the entire imbalance-aware learning process, by defining coherent bin partitions.
	
	Let $D = \{(x_i, y_i)\}_{i=1}^N$, denote the dataset, $x_i \in \mathbb{R}^d$ and $y_i \in \mathbb{R}$. To enable adaptive bin partitioning, we leverage the \emph{coefficient of determination} $R^2$ as a statistical criterion for evaluating local linear coherence within candidate target intervals. The $R^2$ metric quantifies the proportion of target variance explained by a model, making it particularly suitable for identifying regions where a simple linear relationship holds between features and targets. \emph{Higher} $R^2$ indicates greater internal consistency and justifies bin retention, whereas \emph{lower} $R^2$ suggests the need for further subdivision. To this end, we fit a simple \textit{linear} regressor within each candidate bin to compute the $R^2$ score, deliberately using lightweight models to prevent expressive capacity from inflating $R^2$ and masking true local coherence.
	
	Our objective is to partition the continuous target space $y$ into bins $\{b_i\}_{i=1}^B$, where $b_i$ contains samples with similar target structure. For each candidate subset $b_i$, a linear model $f_{b_i}(x_i) = \beta^T x_i + \epsilon_i$ is fitted, and the coefficient of determination (coherence score) is computed as Eq. \ref{eq:R2}:
	\begin{equation}
		R_{b_i}^2 = 1 - \frac{\sum_{i \in b_i} (y_i - \hat{y}_i)^2}{\sum_{i \in b_i} (y_i - \bar{y}_b)^2}
		\label{eq:R2}
	\end{equation}
	where $\hat{y}_i = f_{b_i}(x_i)$ is the model's prediction, and $\bar{y}_b$ is the mean target value in $b_i$. The bin $b_i$ is accepted if: $R_{b_i}^2 \geq \tau \quad \text{and} \quad |b_i| \geq M$. Here $M \in \mathbb{N}$ is \emph{minimum bin size} that declares the minimum number of samples required for a bin to be considered statistically reliable. Setting $M$ too low risks \emph{over-partitioning}, the algorithm may produce an excessive number of small bins many of which lack sufficient data to support meaningful local analysis. Such fragmented bins can disrupt downstream processes by amplifying noise, destabilizing local density estimation, and causing unreliability in next phases (unreliable oversampling in Phase II or incorrect loss weighting in Phase III). Conversely, if $M$ is set too \emph{high}, it may suppress the creation of fine-grained bins necessary to capture localized structure in sparse regions, thereby obscuring rare patterns or minority subspaces crucial to achieving balance. Adaptive selection of $M$ is essential for robust, generalizable performance in imbalanced regression. A fixed $M$ fails across datasets of varying size and sparsity (too restrictive for small data, too permissive for large) leading to poor binning. We propose a data-aware strategy that dynamically sets $M$ based on each dataset’s statistical properties, ensuring reliable bins with sufficient samples while preserving flexibility across diverse target distributions. To this end, we define $M$ as Eq. \ref{eq:M_minimum_Number_of_Samples}:
	\begin{equation}
		M = \max\left( \alpha \cdot N, M_{\text{KDE}} \right)
		\label{eq:M_minimum_Number_of_Samples}
	\end{equation}
	where $\alpha \in (0,1)$ is a \emph{proportion coefficient} indicating the minimum allowable bin size relative to dataset size $N$, and $M_{\text{KDE}}$ is computed from a \textit{kernel density estimation (KDE)} over the target variable $y$ as Eq. \ref{eq:M_KDE} shows:
	\begin{equation}
		M_{\text{KDE}} = \left\lceil \frac{1}{\max_y \, \hat{p}_{\text{KDE}}(y)} \right\rceil
		\label{eq:M_KDE}
	\end{equation}	
	where $\hat{p}_{\text{KDE}}(y)$ denotes the estimated probability density function of the continuous target variable $y$, obtained using KDE. KDE is a non-parametric method for estimating the probability density function of a random variable by smoothing observed data points using a kernel function. The value $\max_y \, \hat{p}_{\text{KDE}}(y)$ reflects the peak density in the target space, and its inverse indicates the minimum number of samples required to statistically support a coherent region under that density. Thereby, high density yields smaller $M$ for finer bins, while low density enforces larger $M$ to avoid unstable fragmentation.
	
	KDE-based estimation excels at uncovering hidden structure in imbalanced data, providing smoother and higher-resolution binning than histogram or quantile methods. However, in small datasets or low-sample regions, KDE may yield unreliable \( M_{\text{KDE}} \) due to noise overfitting or density underestimation from poor bandwidth selection. To address this, we incorporate a fallback term, \( \alpha \cdot N \), which activates when \( M_{\text{KDE}} < \alpha \cdot N \), ensuring a conservative minimum bin size to prevent over-fragmentation. This hybrid approach ensures robustness, scalability, and generality, enabling flexible yet principled binning across diverse datasets.
	
	If a candidate bin $b_i$ fails to meet the coherence and support criteria, i.e., when the coefficient of determination $R_{b_i}^2 < \tau$, the subset is deemed structurally unreliable for modeling. Rather than accepting such a bin, we recursively partition it to seek more locally coherent subregions. Specifically, $b_i$ is split at the mean target value according to Eq. \ref{eq:mu_y}:
	\begin{equation}
		\mu_y = \frac{1}{|b_i|} \sum_{(x_i, y_i) \in b_i} y_i
		\label{eq:mu_y}
	\end{equation}
	resulting in two child bins: $b_i^{\text{left}} = \{(x_i, y_i) \in b_i : y_i < \mu_y\}$ and $b_i^{\text{right}} = \{(x_i, y_i) \in b_i : y_i \geq \mu_y\}$. This recursive partitioning strategy ensures that, over successive iterations, the target space is adaptively segmented into bins that are both statistically coherent (high $R^2$) and sufficiently supported (large enough $M$).
	
	The input to this phase is the raw training dataset $D = \{(x_i, y_i)\}_{i=1}^N$, and the output is a bin-partitioned version of $D$, namely $\mathcal{B} = \{b_1, b_2, \ldots, b_B\}$, where each sample is assigned to a statistically coherent target-space bin $\{b_i\}_{i=1}^B$, serving as a foundation for subsequent balancing operations. The full procedure is formalized in Algorithm \ref{alg:adaptive-binning}, which outlines the iterative bin refinement process, from initialization to final convergence.
	
	\begin{algorithm}
		\footnotesize
		\caption{Adaptive target-space bin partitioning (phase 0)}
		\label{alg:adaptive-binning}
		\begin{algorithmic}[1]
			\Require Training dataset $D = \{(x_i, y_i)\}_{i=1}^N$, where $x_i \in \mathbb{R}^d$ and $y_i \in \mathbb{R}$; threshold $\tau \in [0,1]$ the minimum acceptable $R^2$; adaptive minimum bin size $M = \max(\alpha \cdot N, M_{\text{KDE}})$
			\Ensure Set of accepted bins $\mathcal{B} = \{b_1, b_2, \ldots, b_B\}$ with bin assignments for each sample
			\newline
			\State Initialize queue $Q \gets \{D\}$ the queue of candidate subsets
			\State Initialize final bin set $\mathcal{B} \gets \emptyset$
			\While{$Q$ is not empty}
			\State $b_i \gets$ Pop a subset from $Q$
			\If{$|b_i| < M$}
			\State Discard $b_i$ and \textbf{continue}
			\EndIf
			\State Fit linear regression model $f_{b_i}(y_i) = \beta^T x_i + \epsilon_i$ on $b_i$
			\State Compute $R_{b_i}^2 = 1 - \frac{\sum_{i \in b_i} (y_i - \hat{y}_i)^2}{\sum_{i \in b_i} (y_i - \bar{y}_b)^2}$
			\If{$R_{b_i}^2 \geq \tau$}
			\State Accept $b_i$ and append to $\mathcal{B}$
			\Else
			\State Compute mean target value: $\mu_y = \frac{1}{|b_i|} \sum_{(x_i, y_i) \in b_i} y_i$
			\State Let $b_i^{\text{left}} = \{(x_i, y_i) \in b_i \mid y_i < \mu_y\}$
			\State Let $b_i^{\text{right}} = \{(x_i, y_i) \in b_i \mid y_i \geq \mu_y\}$
			\State Push $b_i^{\text{left}}$ and $b_i^{\text{right}}$ to $Q$
			\EndIf
			\EndWhile
			\State \Return final bin list $\mathcal{B}$ and bin assignment labels for all $(x_i, y_i) \in D$
		\end{algorithmic}
	\end{algorithm}

	
	\subsection{Phase I: Representation Learning}
	Effective learning in imbalanced regression tasks relies not only on data balance, but also on the quality and structure of feature representations. Raw input features, especially in high-dimensional or sparsely populated regions of the feature space, often fail to capture latent structures and predictive semantics necessary for generalization \citep{2021representation}. This limitation is exacerbated in minority regions, where the lack of sufficient data makes it difficult for standard learning algorithms to form reliable hypotheses \citep{2021Korycki}. Phase I addresses this via \emph{representation learning}, projecting data into a compact, target-aware latent space to support robust downstream balancing and regression. To this end, we adopt a \emph{Conditional Variational Autoencoder (CVAE)} \citep{NIPS2015}, a deep generative model that extends the standard Variational Autoencoder (VAE) by incorporating supervision through conditioning. In our setup, the CVAE is conditioned on the continuous target variable $y \in \mathbb{R}$, allowing the model to learn latent embeddings that are both structurally and \textit{predictively relevan}t. This conditioning is achieved by concatenating the target y with the input $x$ for the encoder and with the latent variable $z$ for the decoder, enabling the model to capture target-specific patterns. This target-awareness is particularly beneficial in imbalanced regression settings, as it encourages the latent space to preserve variations aligned with the target distribution, including sparsely populated regions.

	Formally, given an input-target pair $(x, y)$, where $x \in \mathbb{R}^d$ denotes the feature vector and $y \in \mathbb{R}$ is the corresponding target value, the CVAE consists of: An \emph{encoder} (or \emph{inference model}) $q_\phi(z \mid x, y)$, parameterized by $\phi$, which estimates the posterior distribution over the latent variable $z \in \mathbb{R}^k$; A \emph{decoder} $p_\theta(x \mid z, y)$, parameterized by $\theta$, which reconstructs the input $x$ from $z$ and $y$; and a \emph{prior distribution} over $z$, typically chosen as a standard multivariate Gaussian: $p(z) = \mathcal{N}(0, I)$, where $I$ denotes the identity covariance matrix, indicating that each latent dimension is independently normally distributed with unit variance and zero covariance. The model is trained by \emph{maximizing} the \emph{conditional evidence lower bound (ELBO)} for each data point, which serves as a tractable surrogate objective to approximate the intractable true conditional log-likelihood $\log\, p(x \mid y)$ in variational inference as shown in Eq. \ref{eq:L_CVAE}:
	
	\begin{equation}
		\mathcal{L}_{\text{CVAE}} = \mathbb{E}_{q_\phi(z \mid x, y)} \left[ \log\, p_\theta(x \mid z, y) \right] - \mathrm{KL}\left( q_\phi(z \mid x, y) \,\|\, p(z \mid y) \right)
		\label{eq:L_CVAE}
	\end{equation}
	
	The first term is the expected reconstruction log-likelihood, which encourages accurate reconstruction of the input $x$ from the latent variable $z$ and condition $y$. Assuming a Gaussian output distribution for $p_\theta(x \mid z, y)$, this term is approximated via Monte Carlo sampling and implemented as a mean squared error loss: 
	\[
	\mathbb{E}_{q_\phi(z \mid x, y)} \left[ \log\, p_\theta(x \mid z, y) \right] \approx -\|x - \hat{x}\|^2
	\]
	where $\hat{x} = p_\theta(x \mid z, y)$ is the reconstructed input. The second term, the Kullback-Leibler (KL) divergence, regularizes the encoder by minimizing the difference between the learned posterior $q_\phi(z \mid x, y)$ and the prior $p(z)$, promoting smoothness and continuity in the latent space.
	
	The CVAE provides a powerful representation learning mechanism by embedding input features into a structured latent space conditioned on the target variable. This conditioning enables \emph{target-aware latent embeddings} that capture predictive structure aligned with the output space, which is particularly critical in imbalanced regression settings. This conditioning encourages the formation of a semantically smooth latent manifold, enabling \emph{generalization in sparse target regions} by aligning nearby target values with similar latent representations, even when such regions lack sufficient data. Moreover, the probabilistic nature of the CVAE, which models each input-target pair as a distribution over latent variables, facilitates \emph{uncertainty-aware representations}. These capture latent variability and epistemic uncertainty, which are especially valuable for downstream modules such as density-aware sampling or cost-sensitive learning.
	
	The input to this phase is the bin partitioned original dataset in which each sample has already been assigned to a bin $\{b_i\}_{i=1}^B$ in Phase 0, namely $\mathcal{B} = \{b_1, b_2, \ldots, b_B\}$. The output is a learned probabilistic latent representation $z_i \sim \mathcal{N}(\mu_i, \sigma_i^2) \in \mathbb{R}^k$ for each training sample, where $\mu_i$ and $\sigma_i$ are the mean and standard deviation estimated by the encoder. These representations will be passed to both the data-level (Phase II) and the algorithm-level (Phase III) balancing modules, serving as a structured and informative basis for the remainder of the framework.

	\subsection{Phase II: Data-Level Balancing}
	In supervised learning, model generalization is inherently constrained by the availability and quality of training samples \citep{GONG2023Yoidu}. This dependency is amplified in imbalanced regression, where the target variable exhibits unequal density across its domain \citep{BELHAOUARI2024}. In such settings, conventional learners tend to bias predictions toward majority regions while underperforming in minority regions. To address this challenge and tackle data lack, we propose a structured, \textit{two-step data-level balancing} strategy that explicitly addresses imbalances in both the target variable $y \in \mathbb{R}$ and the latent feature space $z \in \mathbb{R}^k$. The pipeline is designed to enhance training coverage in sparse regions while preserving the statistical and geometric structure of the data, and consists of the following steps: in \emph{step 1}, feature-space clustering is performed within each bin using DBSCAN to identify sparse, minority clusters in each bin; \emph{step 2} involves performing localized oversampling of identified minority clusters using an adapted SMOGN technique, guided by cluster-level relevance to prioritize rare target values within each minority cluster, ensuring targeted augmentation without relying on bin-wide distributional alignment.

	\subsubsection{Step 1: Feature-Space Clustering Within Each Bin}
	Within each target-space bin, the distribution of samples in latent feature space can still be uneven, with certain regions remaining sparsely populated  as target-space partitioning alone does not guarantee uniform sample distributions across the latent feature space. To locate these sparse and informative feature regions that merit oversampling, we apply \emph{DBSCAN (Density-Based Spatial Clustering of Applications with Noise)} \citep{ester1996density} within each bin $\{b_i\}_{i=1}^B$. DBSCAN is chosen for its ability to identify non-convex clusters and detect noise points, making it suitable for capturing complex structures \citep{Bushra2021} in high-dimensional feature spaces typical of imbalanced regression tasks. The main idea is that imbalanced learning in regression is not only a function of label distribution but also of how well the feature space is covered across different target intervals.
	
	Let $\{z_i\}_{i=1}^{N_b}$ denote the set of latent feature vectors in bin $b$, where $z_i \in \mathbb{R}^k$, ${N_b}$ is the number of samples in bin $b$. Using latent representations leverages the target-aware, compact embeddings learned in Phase I, which enhance clustering by reducing dimensionality and aligning features with the target distribution. DBSCAN defines a sample $z_i$ as a core point if:
	\[
	|\{z_j \in X_b : \|z_j - z_i\| \leq \varepsilon\}| \geq \text{minPts}
	\]
	where $\varepsilon > 0$ is the neighborhood radius and $\text{\textit{minPts}} \in \mathbb{N}$ is the density threshold. We set $\textit{minPts}$ following the common heuristic $\textit{minPts} = k+1$, where $k$ is the dimensionality of the latent space, ensuring robustness across varying densities. We compute the distance from each point $z \in b$ to its $\text{minPts}$-th nearest neighbor. The radius $\varepsilon_{b}$ is then set automatically as the median of these distances:
	\[
	\varepsilon = \text{median}\left(\{d_{\text{minPts}}(z)\}_{z \in b}\right).
	\]
	
	Phase I's dimensionality reduction helps mitigate noise in the $k$-distance curve, improving the stability of $\varepsilon$ selection in high-dimensional data. Only clusters that are both small relative to other clusters within the same bin (minority clusters) and well-formed (non-noise) are retained for oversampling. This enables targeted data generation in informative low-density regions without introducing synthetic bias.
	
	The input to this step is the latent representation of the set of samples already assigned to each bin $b_i$, and the output is the cluster assignment for each sample within that bin. These clusters provide the structural foundation for the next step, where oversampling is applied only to clusters identified as minority regions, ensuring that synthetic data is introduced with maximal relevance and minimal distortion.

	\subsubsection{Step 2: Oversampling Minority Cluster}
	\label{subsubsec:step2}
	Following the clustering procedure in Step 1, each bin $\{b_i\}_{i=1}^B$ is partitioned into a set of clusters $C_{b_i}$ based on localized density in the latent feature space. In this step, we identify minority clusters within each bin and apply targeted oversampling to augment these regions aiming to enhance minority region coverage. To perform targeted data augmentation, we first identify minority clusters within each bin and then apply localized oversampling using SMOGN. These two interrelated substeps are described below.
	
	\textbf{A. Minority cluster identification via progressive size matching:} As a prerequisite to oversampling, we systematically detect and prioritize minority clusters within each bin using a progressive balancing strategy. For each bin $b_i$, the discovered clusters $C_{b_i}$ are first sorted in ascending order of their size (number of samples). Let $c_{b_i}^{(1)}$ be the smallest cluster, $c_{b_i}^{(2)}$ the second smallest, and so on, with $|c_{b_i}^{(1)}| \leq |c_{b_i}^{(2)}| \leq \cdots \leq |c_{b_i}^{(n)}|$. We then perform \emph{progressive size matching} as follows:
	\begin{itemize}
		\item Select the smallest cluster $c_{b_i}^{(1)}$ and oversample its samples, leveraging its structural information, until its size matches the size of $c_{b_i}^{(2)}$.
		\item $c_{b_i}^{(1)}$ and $c_{b_i}^{(2)}$ are then oversampled equally until they reach the size of $c_{b_i}^{(3)}$.
		\item This process continues iteratively during substep \emph{B} until the total bin size $|b_i|$ (original plus synthetic) reaches the size of the largest bin in the dataset, which serves as the criterion for having sufficient samples in that bin.
	\end{itemize}
	This strategy ensures a balanced increase in sample density across sparse regions while respecting the structural properties of the data.
	
	\textbf{B. SMOGN:} For localized oversampling in each identified minority cluster, we employ SMOGN \citep{SMOGN2017}, a method tailored for continuous targets in imbalanced regression. SMOGN augments underrepresented minority regions by generating synthetic samples, using a \textit{relevance function} $\phi(y) \in [0,1]$ to prioritize rare target values and combining interpolation with controlled noise to ensure target smoothness. In our framework, SMOGN operates on latent representations $z_i \in \mathbb{R}^k$ from Phase I's CVAE, paired with targets $y_i \in \mathbb{R}$. Since we have decided to perform oversampling in a localized manner post-clustering to better target sparse regions, we modify the relevance function to be calculated within each cluster rather than across the entire bin. This adjustment enhances precision by focusing on local target rarity, addressing the unique distribution within minority clusters while reducing the risk of dilution from bin-wide patterns, as supported by the cluster-specific structural insights from DBSCAN. Accordingly, the new cluster-based relevance function $\phi_C(y) \in [0,1]$, defined as:
	\begin{equation}
		\phi_C(y) = 1 - \text{rank}_C(y) / \text{max}\textunderscore \text{rank}_C
		\label{eq:relevance_fi}
	\end{equation}
	evaluates the rarity of each sample within the specific cluster $C$, where $\text{rank}_{C}(y)$ is the rank of target value $y$ based on its frequency within the cluster, and $\text{max}\textunderscore \text{rank}_C$ is the maximum rank within that cluster (equal to the number of unique $y$ values in $C$). Higher $\phi_C(y)$ values indicate underrepresented targets within the cluster. For a seed sample $(z_i, y_i)$ selected from the current minority cluster and prioritized by $\phi_C(y_i)$, SMOGN identifies $k$ nearest neighbors $(z_j, y_j)$ within the same cluster’s latent space $Z_C$. A synthetic sample $(z_{\text{new}}, y_{\text{new}})$ is generated in Eq. \ref{eq:z_new} and Eq. \ref{eq:y_new}:
	
	\begin{equation}
		z_{\text{new}} = z_i + \lambda (z_j - z_i), \quad \lambda \sim \mathcal{U}(0, 1)
		\label{eq:z_new}
	\end{equation}
	
	\begin{equation}
		y_{\text{new}} = y_i + \lambda (y_j - y_i) + \epsilon, \quad \epsilon \sim \mathcal{N}(0, \sigma_{\text{SMOGN, C}}^2)
		\label{eq:y_new}
	\end{equation}
	where $\lambda$ is a random interpolation factor from the uniform distribution $\mathcal{U}(0,1)$, $\epsilon$ is Gaussian noise added to the target for variability, and $\sigma_{\text{SMOGN, C}}^2 = \alpha \cdot \bar{\sigma}_{C}^2 + (1 - \alpha) \cdot \sigma_{\text{base}}^2$ is the noise variance specific to cluster $C$, with $\alpha \in [0,1]$ (default 0.5) weighting the cluster’s average variance $\bar{\sigma}_{C}^2$ against a global baseline $\sigma_{\text{base}}^2$ (default 0.01). To ensure coherence, $z$ vectors are standardized within the cluster before interpolation. This process generates synthetics that align with the local cluster structure, effectively balancing the minority cluster while preserving the continuous nature of the target variable. 
	
	After oversampling, a shallow 1-layer MLP is trained on the full augmented dataset (original + synthetic samples). Although synthetic samples are used during training to expose the network to improved minority-region geometry and push the hidden representations of original samples toward more discriminative directions, only the hidden activations of the original samples are extracted at the end of this phase. These representations, denoted $h^{\text{data}}$, implicitly benefit from the presence of synthetic neighbors during training while preserving exact correspondence with the original instances processed in Phase III. This ensures perfect sample-wise alignment with the algorithm-level features $h^{\text{alg}}$ in the final fusion stage (Phase IV). The Algorithm \ref{alg:data-level-balancing} demonstrates the proposed data-level balancing step-by-step.

	
	\begin{algorithm}
		\caption{Data-Level Balancing (Phase II)}
		\label{alg:data-level-balancing}
		\scriptsize  
		
		\begin{algorithmic}[1]
			\Require
			\Statex ${\text{\{}}X_{b_i}\text{\}}_{i=1}^B$: The set of $(z, y)$ pairs acquired after applying CVAE to the samples in each bin $b_i$, where $z_i \in \mathbb{R}^k$
			\Statex DBSCAN parameters: $\varepsilon$, minPts (set as $k+1$, where $k$ is latent space dimensionality)
			\Statex SMOGN parameters: $k$ (number of nearest neighbors), $\alpha$ (noise weighting), $\sigma_{\text{base}}^2$ (default global noise variance for synthetic sample generation (set to 0.01))
			
			\Ensure
			\Statex $h^{\text{data}}$: data-level hidden representations extracted from the original samples only
			
			\State $target\_bin\_size \gets \arg\max_{|b_i|} \text{where } i \in {1, \ldots, B}$ 
			
			\For{each bin $b_i, \text{where } i \in {1, \ldots, B}$}
			
			\State Let $X_{b_i} = \text{ set of all samples } (z, y) \text{ in bin } b_i$ (after CVAE)
			\State $Clusters$ $\gets$ Apply DBSCAN on ${z_i} \text{ in } X_{b_i}$ 
			\textcolor{gray}{\Comment{Step 1: Feature-Space Clustering Within Each Bin}}

			\State Sort the clusters by ascending size, namely: $|c_{b_i}^{(1)}| \leq |c_{b_i}^{(2)}| \leq \cdots \leq |c_{b_i}^{(n)}|$ 
			\textcolor{gray}{\Comment{Step 2(A): minority clusters identification ($c_{b_i}^{(n)} \text{denotes $n^{th}$ min cluster in bin $b_i$}$)}}
			
			\State Initialize $X_{b_i}^{\text{aug}} \gets X_{b_i}$ \textcolor{gray}{\Comment{Initializing the \textit{oversampled set} with the original samples}}
			\If{$|Clusters| > 1$}
			\textcolor{gray}{\Comment{When DBSCAN yields more than 1 clusters in current bin}}

			\For{$j = 2$ \textbf{to} $|Clusters|$} 
			\textcolor{gray}{\Comment{Handling the progressive manner of oversampling}}
			\For{$l = 1$ \textbf{to} $j$}
			\State Compute $\phi_{c_{b_i}^{(l)}}(y)$ for $c_{b_i}^{(l)}$ using $\phi_{c_{b_i}^{(l)}}(y) = 1 - \text{rank}_{c_{b_i}^{(l)}} / \text{max\_rank}_{c_{b_i}^{(l)}}$ 
			\textcolor{gray}{\Comment{Step 2(B): Oversampling}}
			
			\State Standardize $z$ vectors within $c_{b_i}^{(l)}$ (zero-mean, unit-variance per feature)
			\While{$|c_{b_i}^{(l)}| < |c_{b_i}^{(j)}|$ \textbf{and} $|X_{b_i}^{\text{aug}}| < target\_bin\_size$}
			\State Select seed $(z_i, y_i) \in c_{b_i}^{(l)}$ with probability proportional to $\phi_{c_{b_i}^{(l)}}(y_i)$
			\State Find $k$ nearest neighbors ${(z_j, y_j)}$ in $c_{b_i}^{(l)}$
			\State Sample $\lambda \sim \mathcal{U}(0,1)$
			\State $z_{\text{new}} \gets z_i + \lambda (z_j - z_i)$
			\State $y_{\text{new}} \gets y_i + \lambda (y_j - y_i) + \epsilon$, $\epsilon \sim \mathcal{N}(0, \sigma_{\text{SMOGN},c_{b_i}^{(l)}}^2)$
			\State $\sigma_{\text{SMOGN},c_{b_i}^{(l)}}^2 = \alpha \cdot \bar{\sigma}_{c_{b_i}^{(l)}}^2 + (1 - \alpha) \cdot \sigma_{\text{base}}^2$, where $\bar{\sigma}_{c_{b_i}^{(l)}}^2$ is cluster variance
			\State Add $(z_{\text{new}}, y_{\text{new}})$ to $X_{b_i}^{\text{aug}}$ and update $|c_{b_i}^{(l)}|$
			\EndWhile
			
			\EndFor
			
			\EndFor
			\Else
			\textcolor{gray}{\Comment{When DBSCAN yields just 1 cluster or no clusters, treat the entire bin as a single cluster}}
			\State Compute $\phi_{C_{b_i}}(y)$ for $X_{b_i}$ using $\phi_{C_{b_i}}(y) = 1 - \text{rank}_{C_{b_i}}(y) / \text{max\_rank}_{C_{b_i}}$ within the bin
			\textcolor{gray}{\Comment{Step 2(B): Oversampling}}
			\State Standardize $z$ vectors within $X_{b_i}$ (zero-mean, unit-variance per feature)
			\While{$|X_{b_i}^{\text{aug}}| < target\_bin\_size$}
			\State Select seed $(z_i, y_i) \in X_{b_i}$ with probability proportional to $\phi_{C_{b_i}}(y_i)$
			\State Find $k$ nearest neighbors ${(z_j, y_j)}$ in $X_{b_i}$
			\State Sample $\lambda \sim \mathcal{U}(0,1)$
			\State $z_{\text{new}} \gets z_i + \lambda (z_j - z_i)$
			\State $y_{\text{new}} \gets y_i + \lambda (y_j - y_i) + \epsilon$, $\epsilon \sim \mathcal{N}(0, \sigma_{\text{SMOGN},C_{b_i}}^2)$
			\State $\sigma_{\text{SMOGN},C_{b_i}}^2 = \alpha \cdot \bar{\sigma}_{C_{b_i}}^2 + (1 - \alpha) \cdot \sigma_{\text{base}}^2$, where $\bar{\sigma}_{{C_{b_i}}}^2$ is bin variance
			\State Add $(z_{\text{new}}, y_{\text{new}})$ to $X_{b_i}^{\text{aug}}$
			\EndWhile
			\EndIf
			\EndFor
			\State $S \gets \bigcup_i X_{b_i}^{\text{aug}}$ \textcolor{gray}{\Comment{Final augmented dataset}}
			\State Train a shallow MLP on $S$
			\State Extract the hidden activations of only the original (non-synthetic) samples from the trained MLP
			\State \Return  $h^{\text{data}}$  
		\end{algorithmic}
	\end{algorithm}
	

	\subsection{Phase III: Algorithm-Level Balancing}
	To complement the claimed hypothesis, this phase tackles imbalance from an algorithmic perspective by directly modifying the loss function used for model optimization. Rather than augmenting the data, we adapt the loss surface to emphasize underrepresented regions in both the feature and target spaces. This strategy is especially valuable when data modification may introduce bias or noise \citep{Tarawneh2022}. 
	
	In regression settings with imbalanced target distributions, conventional loss functions such as MSE are inherently biased toward the majority regions of the target space \citep{XuLei2024Li, Aryan2024Jadon}. This leads to suboptimal performance in underrepresented regions, areas where samples are scarce but often the most important. To rectify this, we propose the \emph{Latent-Density Weighted Loss (LDWL)}, a cost-sensitive loss that upweights minority samples based on their joint rarity in: the latent feature space $Z$, learned via the CVAE (Phase I); and the target space $Y$, discretized into statistically coherent bins through adaptive target-space bin partitioning (Phase 0).
	
	Let the training dataset be $D = \{(x_i, y_i)\}_{i=1}^N$, where $x_i \in \mathbb{R}^d$ and $y_i \in \mathbb{R}$, which is fed into Phase~0. As a result, $\mathcal{B} = \{b_1, b_2, \ldots, b_B\}$ is obtained, representing the adaptive bin partitioning of the original dataset. From Phase I, each $x_i$ is passed through a CVAE, producing:  
	\[
	z_i \sim \mathcal{N}(\mu_i, \sigma_i^2), \quad z_i \in \mathbb{R}^k
	\]  
	
	Let $f(z_i) \in \mathbb{R}$ denote the model’s prediction, and define the base loss as Eq. \ref{eq:loss}:
	
	\begin{equation}
		\ell(f(z_i), y_i) = (f(z_i) - y_i)^2
		\label{eq:loss}
	\end{equation}
	
	We now define the bin-aware LDWL objective as Eq. \ref{eq:LDWL}:
	
	\begin{equation}
		\mathcal{L}_{\text{LDWL}} = \frac{1}{N} \sum_{i=1}^{N} \left[ w_i \cdot \ell(f(z_i), y_i) + \lambda w_i \right]
		\label{eq:LDWL}
	\end{equation}	
	where, $\lambda > 0$ is a regularization coefficient, penalizing large weights to prevent overfitting and $w_i \in [0, 1]$ is a \textit{density-aware rarity weight} defined by Eq. \ref{eq:weighting}, designed to emphasize samples that occur in low-density regions of both the feature and target spaces, regarding the bin structure:
	
	\begin{equation}
		w_i = \frac{1}{p(z_i \mid \theta) \cdot p(y_i \mid b_i)}
		\label{eq:weighting}
	\end{equation}
	
	Each component of this formulation captures a distinct notion of sample rarity:
	\begin{itemize}
		\item The latent density term $p(z_i \mid \theta)$ measures how rare a sample $z_i$ is \emph{within the latent feature space} by applying KDE over \emph{all latent representations} $\{z_j\}_{j=1}^N$ (Eq. \ref{eq:KDEz_i}): 
		
		\begin{equation}
			p(z_i \mid \theta) = \frac{1}{N} \sum_{j=1}^N K_{h_{\text{latent}}}(z_i - z_j),
			\label{eq:KDEz_i}
		\end{equation}
		where $\theta = \{K, h_{\text{latent}}\}$ denotes the KDE parameters (kernel function and bandwidth).

		\item The bin-conditioned target density $p(y_i \mid b_i)$ quantifies how rare the target value $y_i$ is \emph{within its assigned adaptive bin} $\{b_i\}_{i=1}^B$. This bin-conditioned target density is also estimated using KDE applied \emph{only to the target values within the same bin} (Eq. \ref{eq:KDEy_i}). Specifically,
		\begin{equation}
			p(y_i \mid b_i) = \frac{1}{n_{b_i}} \sum_{j : b_j = b_i} K_{h_{\text{target}}}(y_i - y_j)
			\label{eq:KDEy_i}
		\end{equation}
		
		where $K_{h_{\text{target}}}$ is the Gaussian kernel with bandwidth $h_{\text{target}}$; $n_{b_i}$ is the number of samples in bin $b_i$. This ensures that target density is computed locally within the bin, preserving adaptive resolution across dense and sparse regions.
	\end{itemize}

	To ensure stable training and prevent numerical explosion from extremely small joint densities, a normalization step is applied afterward:
	\[
	p(y_i \mid b_i) \leftarrow \frac{p(y_i \mid b_i)}{\max_j p(y_j \mid b_j)}, \quad 
	p(z_i \mid \theta) \leftarrow \frac{p(z_i \mid \theta)}{\max_j p(z_j \mid \theta)}
	\]
	And then the final weight normalization is as Eq. \ref{eq:weightNormalization} shows:
	
	\begin{equation}
		w_i \leftarrow \frac{\frac{1}{p(z_i \mid \theta) \cdot p(y_i \mid b_i)}}{\max_j \left[ \frac{1}{p(z_j \mid \theta) \cdot p(y_j \mid b_j)} \right]}
		\label{eq:weightNormalization}
	\end{equation}
	
	This normalization constrains all weights to the range [0,1], preserving their relative magnitude while ensuring bounded gradients during loss optimization. Building on this foundation, the LDWL, as defined in Eq. \ref{eq:LDWL}, comprises two core components:
	\begin{itemize}
		\item \textbf{Weighted Loss} ($w_i \cdot \ell(f(z_i), y_i)$): This part scales the prediction error to focus on underrepresented regions (by upweighting errors in minority regions). The weight $w_i$ is high for rare samples, ensuring the model prioritizes these areas over common ones.
		\item \textbf{Regularization} ($\lambda w_i$): This part adds a penalty for high weights to prevent overfitting to rare outliers. It keeps the model stable by limiting excessive focus on very sparse samples.
	\end{itemize}
	It can be construed that the weight $w_i$ in the LDWL, dynamically prioritizes minority regions by assigning higher values ($w_i \approx 1$) to samples in sparse, underrepresented areas of the target space, while assigning lower values ($w_i \approx 0$) to samples in dense, majority regions, thereby mitigating bias toward well-represented areas and enhancing performance in imbalanced regression tasks. This design enables LDWL to effectively address imbalanced regression challenges by emphasizing underrepresented regions through density-aware weighting, while the regularization term ensures robust optimization by mitigating overfitting to extreme outliers. By leveraging adaptive binning from Phase 0 and latent representations from Phase I, LDWL provides a flexible, data-driven loss formulation that enhances model performance across diverse target distributions, particularly in sparse regions, without requiring data modifications that could introduce bias. The Algorithm \ref{alg:ldwl} formalizes the LDWL computation and training process for Phase III:

	\begin{algorithm}
		\footnotesize
		\caption{Algorithm-level balancing (phase III)}
		\label{alg:ldwl}
		
		\begin{algorithmic}[1]
			
			\Require 
			\Statex Latent representations $\{z_i, \mu_i, \sigma_i\}_{i=1}^N$ from Phase I (CVAE)
			\Statex Target values $\{y_i\}_{i=1}^N$
			\Statex Adaptive bin assignments $\{b_i\}_{i=1}^B$
			\Statex Regression model $f$, base loss $\ell$
			\Statex Regularization parameter $\lambda$
			\Statex KDE bandwidths $h_{\text{latent}}$, $h_{\text{target}}$
			
			\Ensure 
			\Statex $h^{\text{alg}}$: algorithm-level hidden representations
			\newline
			
			\For{each sample $(z_i, y_i)$ in dataset $\{(z_i, y_i)\}_{i=1}^N$}
			\State Obtain latent representation $z_i \sim \mathcal{N}(\mu_i, \sigma_i^2)$ from CVAE
			\State Estimate latent density $p(z_i \mid \theta)$ using KDE with Gaussian kernel and bandwidth $h_{\text{latent}}$
			\State Estimate bin-conditioned target density $p(y_i \mid b_i)$ using KDE within bin $b_i$ with bandwidth $h_{\text{target}}$
			\State Normalize: $p(y_i \mid b_i) \gets \frac{p(y_i \mid b_i)}{\max_j p(y_j \mid b_j)}$ and $p(z_i \mid \theta) \leftarrow \frac{p(z_i \mid \theta)}{\max_j p(z_j \mid \theta)}$
			\State Compute joint weight: $w_i \gets \frac{1}{p(z_i \mid \theta) \cdot p(y_i \mid b_i)}$
			\State Normalize: $w_i \gets \frac{w_i}{\max_j w_j}$
			\EndFor
			\State Train a MLP by minimizing
			\textcolor{gray}{\Comment{Optimize regression model with LDWL loss}}
			
			\[
			\mathcal{L}_{\text{LDWL}} = \frac{1}{N} \sum_{i=1}^N \left[ w_i \cdot \ell(f(z_i), y_i) + \lambda w_i \right]
			\]
			\State Extract the 1activations from the last hidden layer 
			\State \Return $h^{\text{alg}}$
		\end{algorithmic}
	\end{algorithm}


	\subsection{Phase IV: Final Fusion}
	The final phase fuses the outputs of the data- and algorithm-level balancing phases into a single, enriched latent vector that integrates knowledge from both strategies. This phase receives as input two types of latent representations:
	
	\begin{itemize}
		\item From Phase II (data-level balancing), we extract $h^{\text{data}} \in \mathbb{R}^k$ from the last hidden layer of the MLP trained on the balanced dataset. Only representations of \emph{original} samples are retained (synthetic samples are discarded) to ensure that the fused features in Phase IV correspond exactly to the same original instances trained using LDWL in Phase III, guaranteeing consistent and aligned feature integration.
		
		\item From Phase III (algorithm-level balancing), the latent features $h^{\text{alg}} \in \mathbb{R}^k$ processed via the LDWL, extracted from the regressor’s final hidden layer.
	\end{itemize}

	To harness the complementary information captured by these two representations, we employ a \emph{gated fusion} mechanism \citep{Ren_2018_CVPR}, which dynamically integrates the two feature streams. Specifically, we adopt a gated fusion module, which dynamically learns how to integrate information from the two sources by assigning instance-specific importance weights to each representation stream. This gating network is a shallow neural network $g_w: \mathbb{R}^{2k} \to \mathbb{R}^k$ with parameter $w$, that takes the concatenation of both vectors:
	\[
	u_i = [h_i^{\text{data}} \, \| \, h_i^{\text{alg}}] \in \mathbb{R}^{2k}
	\]
	and produces a soft gating vector via sigmoid activation:
	\[
	\alpha_i = \sigma(g_w(u_i)) \in [0, 1]^k
	\]
	where $\sigma(\cdot)$ is applied element-wise. The final fused feature representation $\tilde{x}_i \in \mathbb{R}^k$ is computed as Eq. \ref{eq:fusion}:
	
	\begin{equation}
		\tilde{x}_i = \alpha_i \odot h_i^{\text{data}} + (1 - \alpha_i) \odot h_i^{\text{alg}}
		\label{eq:fusion}
	\end{equation}	
	where $\odot$ denotes element-wise multiplication. This allows the network to adaptively prioritize either the data-level or algorithm-level features depending on the complexity and distribution of each input sample. The fused vectors \(\tilde{x}_i\) thus yield an enhanced and enriched representation that leverages the complementary strengths of data-level and algorithm-level balancing. By employing an attention-based gated fusion mechanism, the model dynamically assigns instance-specific weights to each stream, ensuring adaptive integration that prioritizes the most informative features for individual samples. This hybrid approach remains regressor-agnostic, allowing any standard regression head to be appended at the pipeline's end, thereby seamlessly inheriting the robustness advantages of combined imbalanced learning strategies across the target distribution. Algorithm \ref{alg:final-hybrid-regression}, demonstrates the flow of the 4th phase.

	\begin{algorithm}
		\caption{Final Fusion (phase IV)}
		\label{alg:final-hybrid-regression}
		\footnotesize
		\begin{algorithmic}[1]
			
			\Require 
			
			\Statex $\{h_i^{\text{data}}\}_{i=1}^N$: Data-level latent representations (from Phase II, original samples only)
			\Statex $\{h_i^{\text{alg}}\}_{i=1}^N$: Algorithm-level latent representations (from Phase III)
			\Statex $\{y_i\}_{i=1}^N$: Ground-truth target values
			\Statex $g_w$: Gating network for feature fusion with parameters $w$

			\Ensure 
			\Statex The fused vector $\tilde{x}_i \in \mathbb{R}^k$ that embodies a principled hybridization of data-level and algorithm-level imbalanced learning methods.
			\newline
			\textcolor{gray}{\Comment{Step 1: Gated Feature Fusion}}
			\For{each sample $i = 1$ to $N$}
			\State Concatenate features: $u_i \gets [h_i^{\text{data}} \, \| \, h_i^{\text{alg}}]$
			\State Compute fusion gate: $\alpha_i = \sigma(g_w(u_i)) \in [0,1]^k$
			\State Fuse features: $\tilde{x}_i = \alpha_i \odot h_i^{\text{data}} + (1 - \alpha_i) \odot h_i^{\text{alg}}$
			\EndFor
			\newline

			\State \Return final fused vector $\tilde{x}_i \in \mathbb{R}^k$
			
			\textcolor{gray}{\Comment{Fused vector, enriched by data- and algorithm-level balancing, is regressor-agnostic and ready for any downstream regressor.}}
			
		\end{algorithmic}
	\end{algorithm}
	

	\subsection{Workflow Integration and Summary}
	This section summarizes the full hybrid imbalanced regression pipeline. It integrates all phases (adaptive bin partitioning, representation learning, data-level balancing, algorithm-level balancing, and final fusion) into a cohesive end-to-end workflow. Algorithm~\ref{alg:hybrid-framework} details the complete training and inference process, clarifying data flow and inter-phase dependencies to ensure clarity, reproducibility, and methodological unity.

	\begin{algorithm}
		\caption{Workflow of the hybrid imbalanced regression framework}
		\label{alg:hybrid-framework}
		\footnotesize
		\begin{algorithmic}[1]
			
			\Require 
			\Statex Original dataset
			\Statex Framework hyperparameters (see individual phases for specifics)
			
			\Ensure
			\Statex Enriched fused vector $\tilde{x}_i \in \mathbb{R}^k$ combining data- and algorithm-level balancing knowledge, ready for any downstream regressor.
			\newline
			\Statex \textbf{Phase 0: Adaptive Target-Space Bin Partitioning} 
			\State Partition the continuous target space $y$ into adaptive bins based on local explainability to capture statistical heterogeneity \textcolor{gray}{\Comment{Algorithm~\ref{alg:adaptive-binning}}}
			
			\Statex \textbf{Phase I: Representation Learning}
			\State Learn target-aware latent representations $z_i$ from original data using CVAE 
			
			\Statex \textbf{Phase II: Data-Level Balancing}
			\For{each target bin}
			\State a. Identify sparse feature regions via density-based clustering
			\State b. Apply localized oversampling (SMOGN) to minority clusters
			\State c. Train an MLP on the oversampled dataset, then extract the last hidden layer's vectors $h^{\text{data}}$ for the original samples only (synthetic samples excluded).
			\EndFor
			\textcolor{gray}{\Comment{Algorithm~\ref{alg:data-level-balancing}}} 
			
			\Statex \textbf{Phase III: Algorithm-Level Balancing}
			\State Train an MLP using the Latent-Density Weighted Loss (LDWL), then extract the last hidden layer's vector $h^{\text{alg}}$ for the original samples.
			\textcolor{gray}{\Comment{Algorithm~\ref{alg:ldwl}}}
			
			\Statex \textbf{Phase IV: Final Fusion}
			\State Fuse the hidden layer outputs extracted from Phases II and III (namely $h^{\text{data}}$ and $h^{\text{alg}}$) using a gated fusion mechanism.
			\State \Return Final fused vector $\tilde{x}_i \in \mathbb{R}^k$
			\textcolor{gray}{\Comment{Algorithm~\ref{alg:final-hybrid-regression}}}
			\newline
			\textcolor{gray}{\Comment{The fused vector is the final output of the proposed hybrid imbalanced regression framework and is regressor-agnostic, ready for use with any downstream regressor.}}
			
		\end{algorithmic}
	\end{algorithm}

	\subsection{Computational Complexity Analysis}
	To elucidate the computational demands of the proposed five-phase hybrid framework, this subsection analyzes the time complexity of each phase, providing insight into their scalability and practical applicability. Let $N$ denote the number of training samples and $d$ the original feature dimensionality. The computational cost of Phase 0 is dominated by the repeated linear regression fits performed on candidate bins during recursive partitioning. Fitting a linear model on a subset of size $|b|$ costs $O(d^2 |b| + d^3)$ with standard solvers. In the theoretical worst case, the recursion may visit $O(N/M)$ nodes (where $M \approx \alpha N$ is the minimum bin size), leading to $O(d^2 N^2)$ time. However, the strict acceptance criterion $R^2 \geq \tau$ causes early termination after only a few splits: across all benchmark datasets, Phase 0 consistently produced 18–28 final bins, resulting in empirical runtime scaling indistinguishable from $O(N d \log N)$. Computing $M_{\text{KDE}}$ via 1D kernel density estimation on the target variable adds only $O(N)$. Thus, Phase 0 remains highly efficient in practice despite its conservative worst-case bound.
	
	The computational complexity of Phase I, Representation Learning, is assessed for a dataset of $N$ samples, each with $d$ feature dimensions. The CVAE trains an encoder and decoder, each with $L$ layers and hidden size $H$, over $E$ epochs to produce $k$-dimensional latent representations. For each sample, a forward/backward pass through the encoder (input: $d+1$ dimensions, including target $y$) and decoder (input: $k+1$, output: $d$) incurs $O(d H L)$ due to matrix operations across layers. For $N$ samples over $E$ epochs, the total training cost is $O(N E d H L)$. The ELBO computation, including reconstruction (MSE, $O(d)$) and KL divergence ($O(k)$) per sample, is negligible, as it is subsumed by network operations. Thus, the overall complexity is $O(N E D)$, where $D = d H L$ encapsulates model complexity, ensuring scalability with the generation of target-aware latent representations.
	
	Let $k$ denote the latent dimensionality. Phase II consists of (i) DBSCAN clustering within each of the $B$ bins and (ii) progressive localized SMOGN oversampling followed by training a shallow MLP on the augmented data. DBSCAN over all samples costs $O(N k \log N)$ using standard implementations. Oversampling generates at most one synthetic sample per original sample in the worst case. Within each (progressively processed) minority cluster of size $m$, finding $k$-nearest neighbors and generating a new point costs $O(m k log m)$. Across all bins and iterations the total oversampling cost is bounded by $O(N k log N)$. Training the final MLP  on at most $2N$ samples costs $O(N k)$. Consequently, the overall time complexity of Phase II is $O(N k log N)$, which reduces to $O(N log N)$ for fixed $k$. In practice, Phase II is fast and easily scales to hundreds of thousands of samples.
	
	Phase III consists of (i) density-aware weight computation and (ii) training a small MLP with the LDWL loss, followed by hidden representation extraction. Global latent density estimation via KDE in $\mathbb{R}^k$ costs $O(N^2 k)$ naïvely, but is reduced to $O(N k \log N)$ using KD-tree acceleration. Bin-local 1D target density estimation costs $O(N^2 / B)$ in total. Computing and normalizing the $N$ weights is $O(N)$. Training the MLP over $E$ epochs costs $O(N E k)$. With fixed small $k$, fixed $E \leq 100$, and $B \gtrsim 20$, the dominant practical term is the bin-local KDE ($O(N^2 / B)$), which remains easily manageable.  The overall time complexity is therefore $O(N^2 / B + N k \log N + N E k)$. 
	
	Phase IV performs gated fusion of the two $k$-dimensional streams $h^{\text{data}}$ and $h^{\text{alg}}$, followed by end-to-end training of a lightweight gating network and a final regression head over $E$ epochs. Concatenation and gating (a shallow network with hidden size $H$) cost $O(N k H)$ once, while each training epoch on the fused $k$-dimensional features costs $O(N k H)$. The total complexity is therefore $O(N E k H)$. 
	
	The computational complexity of the proposed five-phase hybrid framework is summarised in Table~\ref{tab:time_complexity}. 
	Phase~0 (adaptive binning) scales empirically as $O(N d \log N)$ thanks to early termination. 
	Phase~I (CVAE representation learning) costs $O(N E d)$. 
	Phase~II (data-level balancing) is $O(N k \log N) \rightarrow O(N \log N)$ for fixed $k$. 
	Phase~III (algorithm-level balancing) contributes $O(N^{2}/B + N E k)$, where the bin-local 1D KDE term $O(N^{2}/B)$ dominates. 
	Phase~IV (final fusion) requires only $O(N E k)$.	
	With fixed small latent dimension $k$, bounded epochs $E$, and typically $B $ adaptive bins, the $N^{2}/B$ term from Phase~III is the largest across the entire pipeline. 
	All other terms involving $d$ (original feature dimension) disappear after the CVAE bottleneck, and the $N^{2}/B$ KDE operations are performed on cheap 1D targets within each bin. 
	Consequently, the overall time complexity of the complete framework is $O(N^{2}/B)$, or roughly $O(N^{2})$ with a very small constant.

	\begin{table}[t]
		\centering
		\tiny
		\caption{Time complexity of each phase and of the complete pipeline. $N$: number of samples, $d$: original feature dimension, $k$: latent dimension (fixed), $E$: training epochs, $B$: number of adaptive bins.}
		\label{tab:time_complexity}
		\begin{tabular}{lcc}
			\toprule
			\textbf{Phase}                  & \textbf{Time Complexity}               & \textbf{Dominant Operation(s)} \\
			\midrule
			Phase 0 – Adaptive Binning      & $O(N d \log N)$ (empirical)            & Repeated linear regression fits with early termination \\[4pt]
			Phase I – Representation Learning & $O(N E d)$                           & CVAE training \\[4pt]
			Phase II – Data-Level Balancing & $O(N k \log N) \rightarrow O(N \log N)$ & DBSCAN + localized SMOGN \\[4pt]
			Phase III – Algorithm-Level     & $O(N^2 / B + N E k)$                    & Bin-local KDE (dominant term) \\[4pt]
			Phase IV – Final Fusion         & $O(N E k)$                             & Gated fusion + final MLP training \\[4pt]
			\midrule
			\textbf{Overall Pipeline}       & $O(N^2 / B)$                           & Bin-local KDE in Phase III \\
			\bottomrule
		\end{tabular}
	\end{table}


	\section{Results and Analysis}
	\label{sec:resultAndDiscussion}

	\subsection{Evaluation Metrics}
	\label{subsec:evaluation_metrics}
	To evaluate our proposed framework for imbalanced regression, we employ three standard metrics, including:  
	
	\begin{itemize}
		\item \textit{Root Mean Squared Error (RMSE)} (Eq. \ref{eq:rmse}) that quantifies the square root of the average squared difference between predicted and actual values, emphasizing larger errors while preserving the original unit of the target:
		\begin{equation}
			\text{RMSE} = \sqrt{\frac{1}{N} \sum_{i=1}^{N} (y_i - \hat{y}_i)^2}
			\label{eq:rmse}
		\end{equation}
		
		\item \textit{Mean Absolute Error (MAE)} (Eq. \ref{eq:mae}) that measures the average absolute difference, offering robustness to outliers:
		\begin{equation}
			\text{MAE} = \frac{1}{N} \sum_{i=1}^{N} |y_i - \hat{y}_i|
			\label{eq:mae}
		\end{equation}
		
		\item \textit{Coefficient of determination ($R^2$)} (Eq. \ref{eq:r2}) that evaluates the proportion of target variance explained by the model, reflecting overall fit:
		\begin{equation}
			R^2 = 1 - \frac{\sum_{i=1}^{N} (y_i - \hat{y}_i)^2}{\sum_{i=1}^{N} (y_i - \bar{y})^2}
			\label{eq:r2}
		\end{equation}
	\end{itemize}
	
	In these equations, $y_i$ denotes the actual target value, $\hat{y}_i$ the predicted value, $N$ the number of samples, and $\bar{y} = \frac{1}{N} \sum_{i=1}^{N} y_i$ the mean of actual target values.

	\subsection{Parameter Setting}
	The proposed hybrid framework is implemented with the following phase-specific hyperparameters, set and kept fixed across all benchmark datasets and regressors to ensure fair and reproducible comparisons. The target variable is normalized to zero mean and unit variance prior to all phases and model training. All training procedures across the phases were conducted within a 10-fold cross-validation framework, with folds stratified by Phase 0 bins to ensure consistent evaluation and robust generalization. In Phase~0, adaptive bin partitioning uses a fixed $R^2$ acceptance threshold of $\tau = 0.85$, a proportion coefficient $\alpha = 0.02$ to determine the minimum bin size, and a Gaussian KDE with bandwidth $0.1$ applied on the standardized target variable to estimate $M_{\text{KDE}}$ for data-aware support enforcement. These values were selected because they provided stable partitions across datasets with different scales, and because bin sizes remained well-balanced under this parameterization. In future work, we plan to develop a fully data-driven strategy for automatically selecting $\tau$, $\alpha$, and KDE bandwidths based on the statistical properties of each dataset.
	
	In Phase~I, the CVAE uses a fixed architecture with input dimension $d$, hidden dimension 128, and latent dimension $k = 16$. The encoder and decoder each include two fully connected layers with SiLU activation, batch size 64, and 20\% dropout. An auxiliary regression head (latent $z \rightarrow y$) is trained jointly via MSE loss. Optimization uses AdamW with learning rate $10^{-3}$ and weight decay $10^{-4}$, trained for up to 100 epochs with gradient clipping (max norm 1.0). The KL weight $\beta$ is linearly annealed from 0 to $0.5$ over the first 50 epochs and held constant thereafter; the auxiliary loss weight is fixed at $\lambda_{\text{aux}} = 1.0$. Early stopping (patience 25) is applied based on validation loss. After training, the deterministic encoder mean $\mu_i \in \mathbb{R}^{16}$ (not stochastic samples) is used as the latent representation $z_i$ in all subsequent phases to improve stability and reduce variance across phases. The latent dimensionality $k = 16$ was chosen as a compromise between representation richness and computational efficiency; preliminary tests showed larger $k$ values did not yield performance gains.
	
	In Phase~II, DBSCAN clusters latent representations within each bin using a dynamic radius $\varepsilon$ equal to the median minPts-th nearest-neighbor distance, with minPts $= k+1$. Progressive oversampling aligns minority cluster sizes to the next largest, up to the size of the largest original bin in the dataset. SMOGN generates synthetic samples via latent-space interpolation with 5 nearest neighbors, cluster-specific relevance $\phi_C(y)$, and latent vectors standardized within each cluster before interpolation; the target noise variance is
	$ \sigma^2_{\text{SMOGN},C} = 0.5 \cdot \bar{\sigma}_C^2 + 0.5 \cdot 0.01$. A shallow 1-layer MLP ($k \rightarrow 16 \rightarrow 1$) with Tanh activation, batch normalization, batch size 64, and 20\% dropout is trained for 100 epochs on the oversampled dataset (original + synthetic samples) using AdamW (learning rate $10^{-3}$, weight decay $10^{-4}$) and MSE loss. The 16-dimensional hidden representations of only the original samples from this MLP are extracted as $h_i^{\text{data}}$ and used in Phase~IV. The oversampling settings were chosen to ensure stable synthetic generation independent of dataset size, preventing over-extrapolation in sparse regions.
	
	In Phase~III, LDWL weights are computed using Gaussian KDE (bandwidth 0.1) applied to the standardized target values locally within each bin and to the deterministic latent means $\mu_i$ globally. Densities are normalized by their respective maxima, and final instance weights $w_i$ are scaled to $[0,1]$ (as described in Eq.~\ref{eq:weightNormalization}). The training objective is the weighted MSE plus a regularization term $\lambda w_i$ with $\lambda = 0.001$. A 3-layer MLP ($k \rightarrow 32 \rightarrow 16 \rightarrow 1$) with LeakyReLU activations, batch size 64, 20\% dropout, and AdamW optimizer (learning rate $10^{-3}$, weight decay $10^{-4}$) is trained for 100 epochs with cosine annealing and gradient clipping (max norm 1.0). The 16-dimensional activations from the last hidden layer are extracted as $h_i^{\text{alg}}$ and forwarded to Phase~IV. KDE bandwidths were fixed across datasets to maintain consistent density estimation behavior; adaptive bandwidth selection will be investigated as future work.
	
	In Phase~IV, the 16-dimensional data-level features $h_i^{\text{data}}$ (from the Phase~II MLP) and algorithm-level features $h_i^{\text{alg}}$ (from the Phase~III MLP) are concatenated into $u_i \in \mathbb{R}^{32}$. A gating network with architecture $32 \rightarrow 64 \rightarrow 16$, using Tanh activations, batch normalization, and 20\% dropout, outputs a sigmoid-activated gate vector $\alpha_i \in [0,1]^{16}$. The fused representation is computed as $\tilde{x}_i = \alpha_i \odot h_i^{\text{data}} + (1-\alpha_i) \odot h_i^{\text{alg}} \in \mathbb{R}^{16}$. A final linear projection head ($16 \rightarrow 1$) predicts the standardized target using MSE loss. The fusion module and projection head are trained end-to-end with AdamW (learning rate $5 \times 10^{-4}$, weight decay $0.01$), batch size 64, for a maximum of 100 epochs using ReduceLROnPlateau (factor 0.5, patience 10) and early stopping (patience 25) on validation MSE. The gating architecture was selected to balance model expressiveness with computational efficiency; deeper fusion networks showed no additional benefit in preliminary experiments.
	
	The fused vectors $\tilde{x}_i$ are compatible with any downstream regressor. To isolate the effect of the proposed imbalance-handling pipeline, all regressors are used with default hyperparameters (Table~\ref{tab:regressors_default_hyperparams}). Performance is averaged over 10-fold cross-validation with folds stratified by Phase-0 bins to preserve minority regions.
	This evaluation strategy ensures fairness, prevents overfitting to specific splits, and maintains consistent representation of sparse target regions across folds.

	\begin{table}[htbp]
		\centering
		\caption{Default Hyperparameter Settings for Selected Regressors. 
			All models use scikit-learn defaults (with \emph{random state=42} 
			for reproducibility where applicable). Parameters are identical across datasets.}
		\label{tab:regressors_default_hyperparams}
		{\scriptsize
			\resizebox{\textwidth}{!}{
				\begin{tabular}{@{}ll||@{\hspace{1.2em}}ll||@{\hspace{1.2em}}ll@{}}
					\toprule
					\textbf{Regressor} & \textbf{Param = Value} &
					\textbf{Regressor} & \textbf{Param = Value} &
					\textbf{Regressor} & \textbf{Param = Value} \\
					\midrule
					
					\emph{MLP} & Hidden layers = (100,) &
					\emph{XGBoost} & Number of trees = 100 &
					\emph{Linear Reg.} & Fit intercept = True \\
					
					& Activation = ReLU &
					& Learning rate = 0.3 &
					& Copy X = True \\
					
					& Optimizer = Adam &
					& Max depth = 6 &
					& Jobs = None \\
					
					& L2 penalty = 0.0001 &
					& Subsample = 1.0 &
					& Positive = False \\
					
					& Initial LR = 0.001 &
					& Column subsample = 1.0 & \\[-0.5ex]
					
					& Max iterations (epochs) = 200 &
					& Random state = 42 & \\
					
					& Random state = 42 & & & & \\
					
					\midrule
					
					\emph{KNN} & Neighbors = 5 &
					\emph{SVR} & Kernel = RBF &
					\emph{Ridge} & Regularization = 1.0 \\
					
					& Weights = uniform &
					& Degree = 3 &
					& Fit intercept = True \\
					
					& Algorithm = auto &
					& Gamma = scale &
					& Copy X = True \\
					
					& Leaf size = 30 &
					& Coef0 = 0.0 &
					& Max iterations = auto \\
					
					& Distance = Euclidean &
					& Tolerance = 0.001 &
					& Tolerance = 0.0001 \\
					
					& Power = 2 &
					& C = 1.0 &
					& Solver = auto \\
					
					& & & Epsilon = 0.1 &
					& Random state = 42 \\
					
					& & & Shrinking = True & & \\
					
					& & & Cache size = 200 & & \\
					
					\bottomrule
				\end{tabular}
			}
		}
	\end{table}

	\subsection{Results and Comparisons}

	\subsubsection{Baseline Methods}
	\label{sec:baseline-methods}
	Despite growing interest in imbalanced regression, hybrid approaches that combine \textit{data-level} and \textit{algorithm-level} strategies remain extremely scarce, particularly those integrating resampling with cost-sensitive or loss-reweighting mechanisms that directly alter the optimization objective. Existing works (like \citep{Avelino2025, 2018REBAGG}) primarily combine data manipulation with ensemble aggregation but do not modify individual model loss functions to prioritize underrepresented target regions. To the best of our knowledge, no prior study has proposed a hybrid pipeline that unifies data-level balancing with cost-sensitive learning at the algorithm level. This absence underscores the novelty of our approach. Nevertheless, to ensure a rigorous and fair evaluation, we compare our hybrid framework against some widely used individual regressors (MLP, XGBoost, Linear Regression, KNN, SVR, Ridge) without any imbalance handling. These models serve as standard, non-imbalanced baselines in regression literature and allow us to demonstrate that our hybrid strategy not only addresses imbalance but also considerably outperforms standalone regressors across benchmark datasets. The detailed comparison is presented in the \hyperref[sec:discussion]{discussion} when answering RQ2.

	\subsubsection{Adaptive Target-Space Bin Partitioning}
	To overcome the lack of discrete class-like structure in regression, we introduce a novel adaptive bin partitioning strategy driven by local $R^2$ coherence scores. Unlike fixed quantiling (which imposes arbitrary) our method leverages $R^2$ to quantify the linear relationship between input features and the target variable in localized intervals. This allows for data-driven segmentation of the continuous target space, such that each bin captures regions of distinct predictive relevance. The $R^2$ statistic is particularly advantageous here as it simultaneously accounts for variance explanation and alignment with model-learnable structures, ensuring that bins are not only statistically consistent but also meaningful for subsequent phases of representation learning and resampling. Fig. \ref{fig:adaptive_bin_Partitioning_results} illustrates the results of the proposed adaptive bin partitioning method applied to the target variable distribution across the evaluated datasets. As shown, the proposed method transforms skewed and unevenly populated target ranges into more structured and analyzable segments, enabling more effective handling of minority regions in downstream phases. 	Complementing these visual insights, Table \ref{tab:bin-partitioning-stats} summarizes the bin partitioning statistics for each dataset, including the number of bins generated, bin boundaries, and sample counts per bin.
	
	The proposed adaptive bin partitioning method sometimes yields bins with zero width, as shown in Table~\ref{tab:bin-partitioning-stats} (datasets with zero minimum bin size). Such bins arise when many instances share an identical target value, forming a plateau in the target distribution. In these regions, local linear regression achieves near-perfect fit (namely, $R^2 \approx 1$), so the algorithm correctly assigns a dedicated bin, treating the plateau as a homogeneous, class-like segment for targeted oversampling, thereby emulating imbalance handling in classification. In addition in certain datasets, the method produces just a single bin for the entire training set. This occurs when the global linear model achieves a high $R^2$ value, indicating that a single linear relationship effectively explains the target-feature association across the entire training set, making further bin subdivision unnecessary and statistically unjustified. Therefore, the proposed method appropriately avoids unnecessary fragmentation, preserving the data as one well-represented predictive regime.

	
	\begin{figure}[htbp]
		\centering
		\subfloat[\centering california]{
			\includegraphics[width=0.45\textwidth]{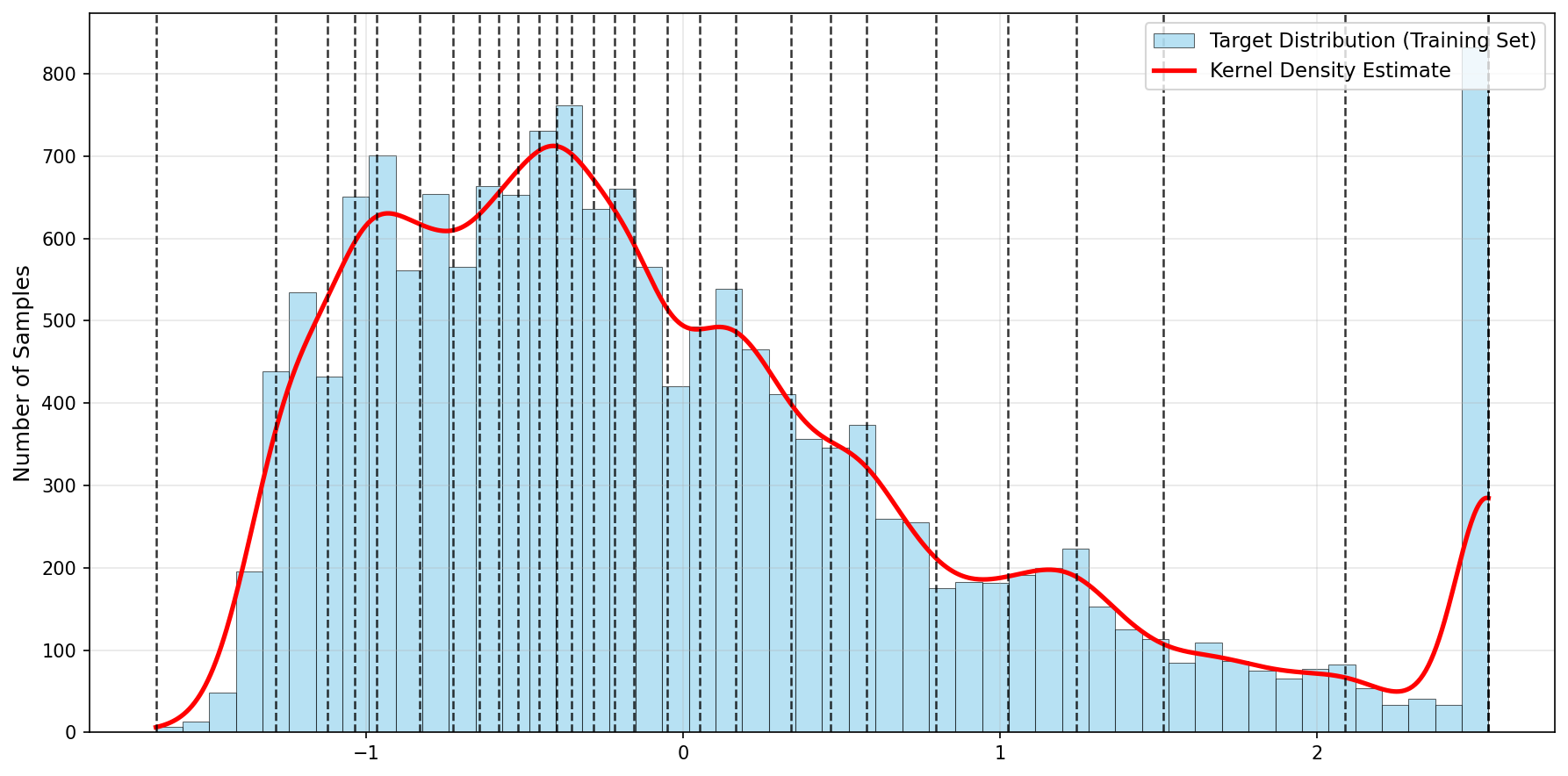}
			\label{fig:raw-California}
		}
		\hfill
		\subfloat[\centering compactive]{
			\includegraphics[width=0.45\textwidth]{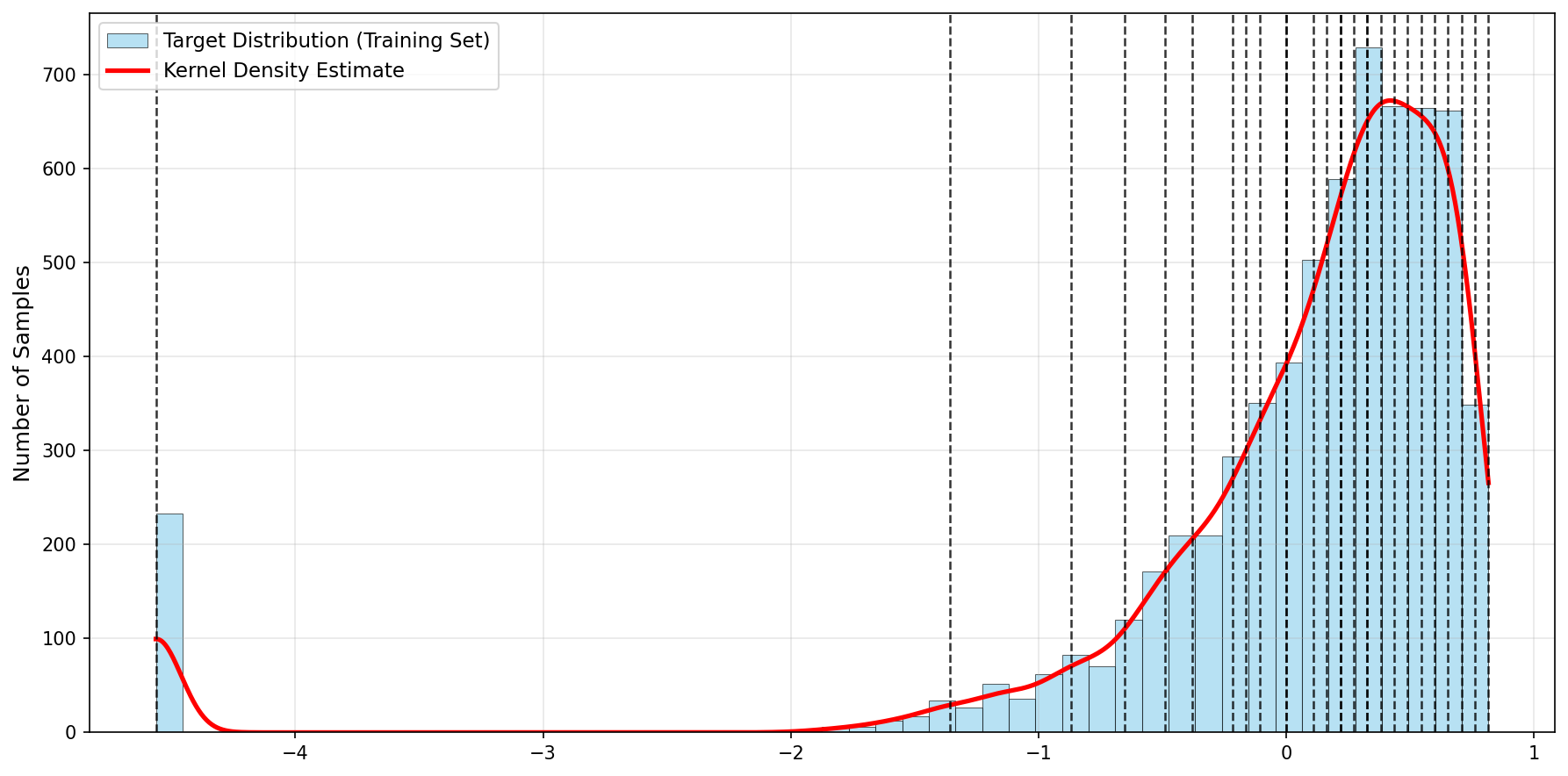}
			\label{fig:binned_compactive}
		}
		
		\subfloat[\centering cpu\_small]{
			\includegraphics[width=0.45\textwidth]{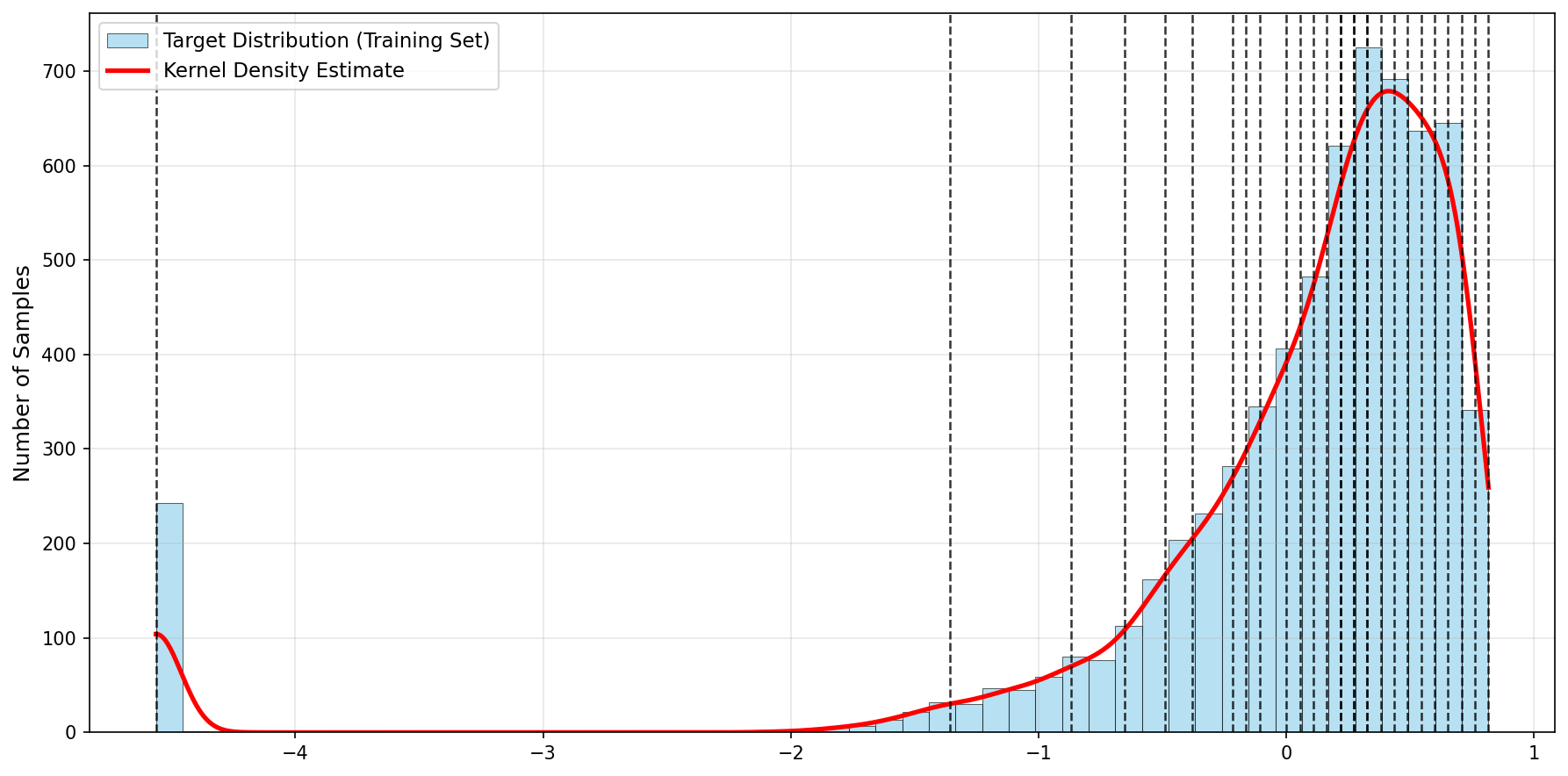}
			\label{fig:raw-cpuSmall}
		}
		\hfill
		\subfloat[\centering heat]{
			\includegraphics[width=0.45\textwidth]{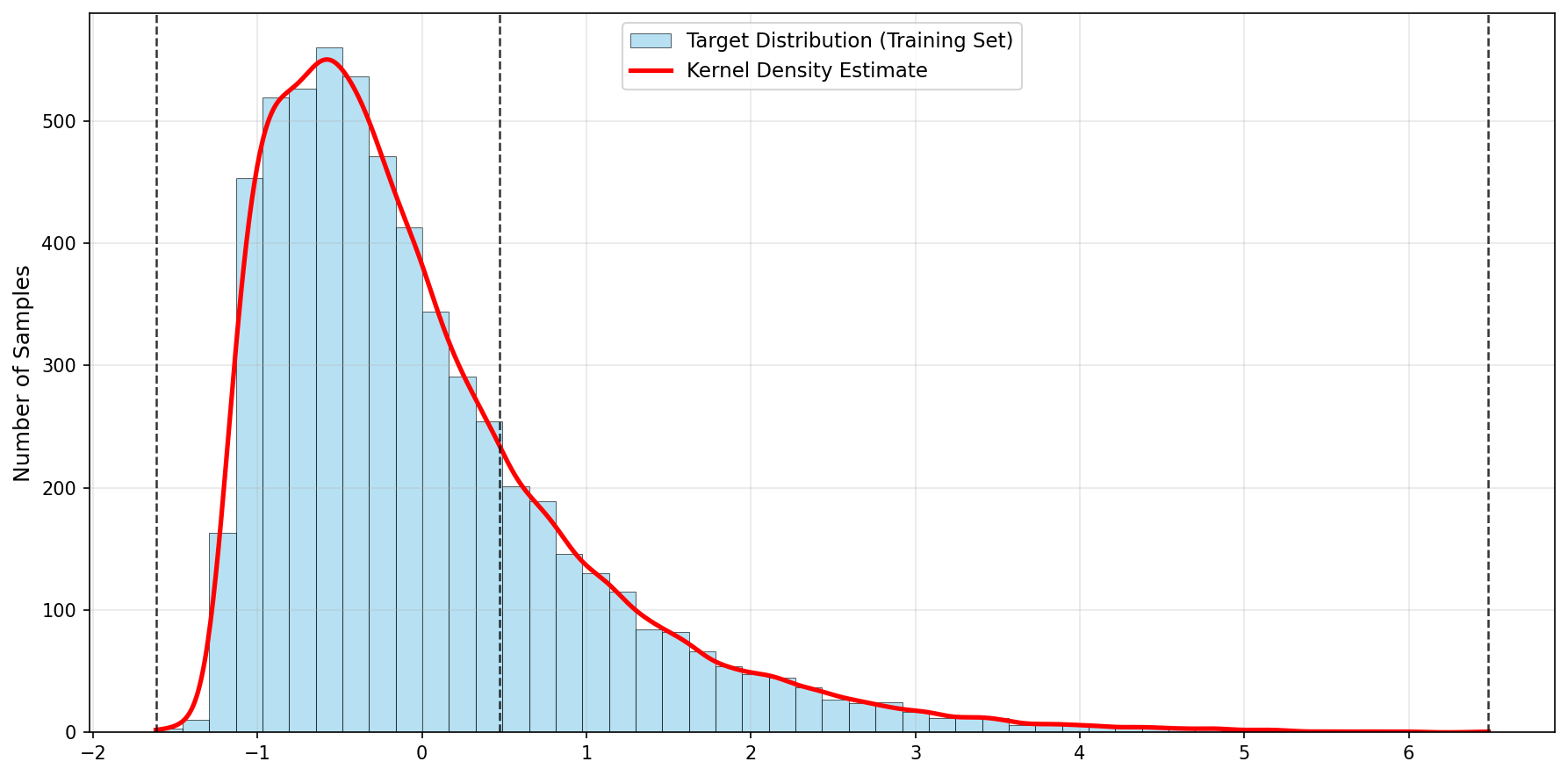}
			\label{fig:binned_heat}
		}
		
		\subfloat[\centering wine\_quality]{
			\includegraphics[width=0.45\textwidth]{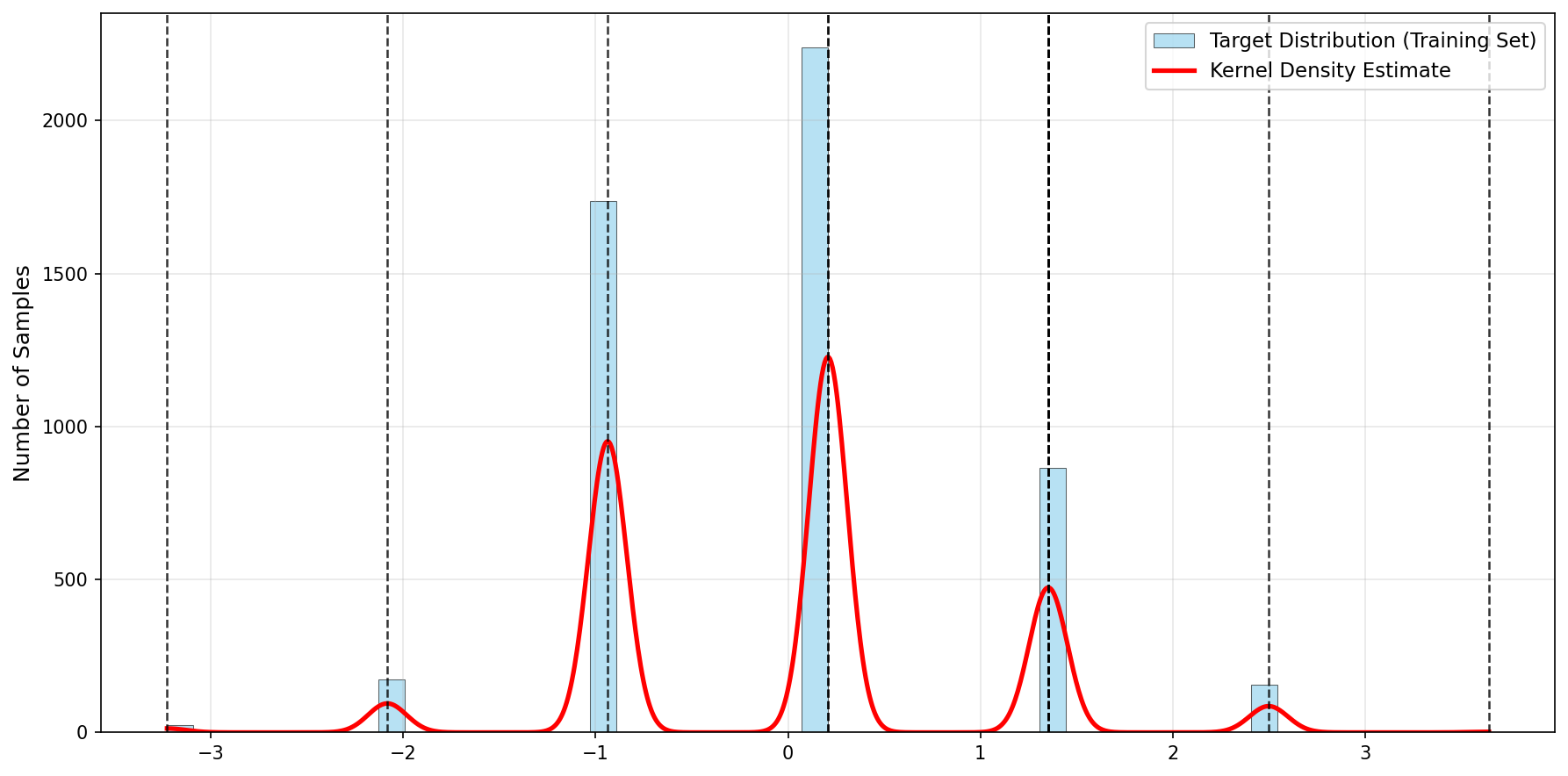}
			\label{fig:winequality}
		}
		\hfill
		\subfloat[\centering abalone]{
			\includegraphics[width=0.45\textwidth]{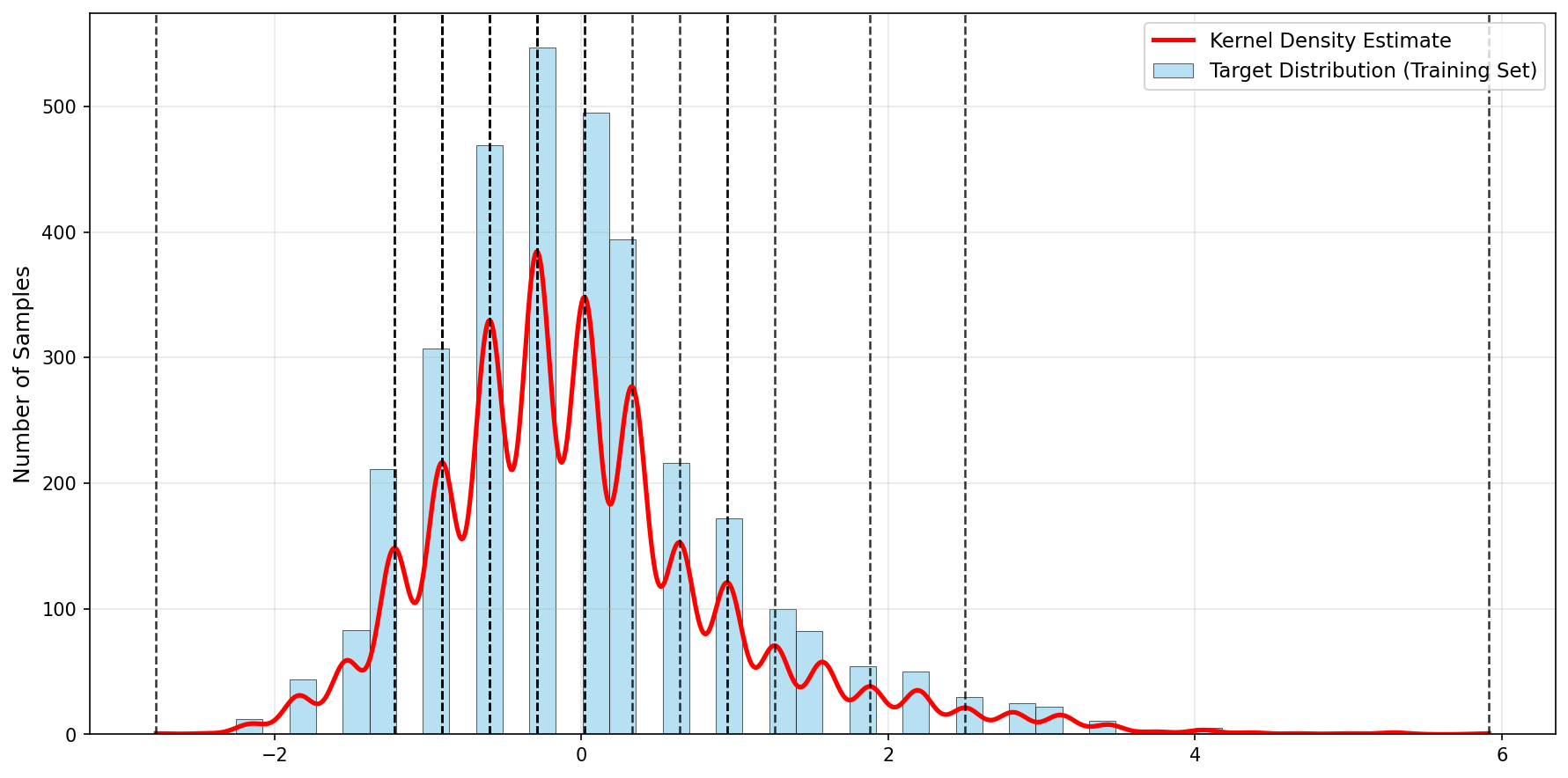}
			\label{fig:binned_abalone}
		}
		
		\subfloat[\centering space\_ga]{
			\includegraphics[width=0.45\textwidth]{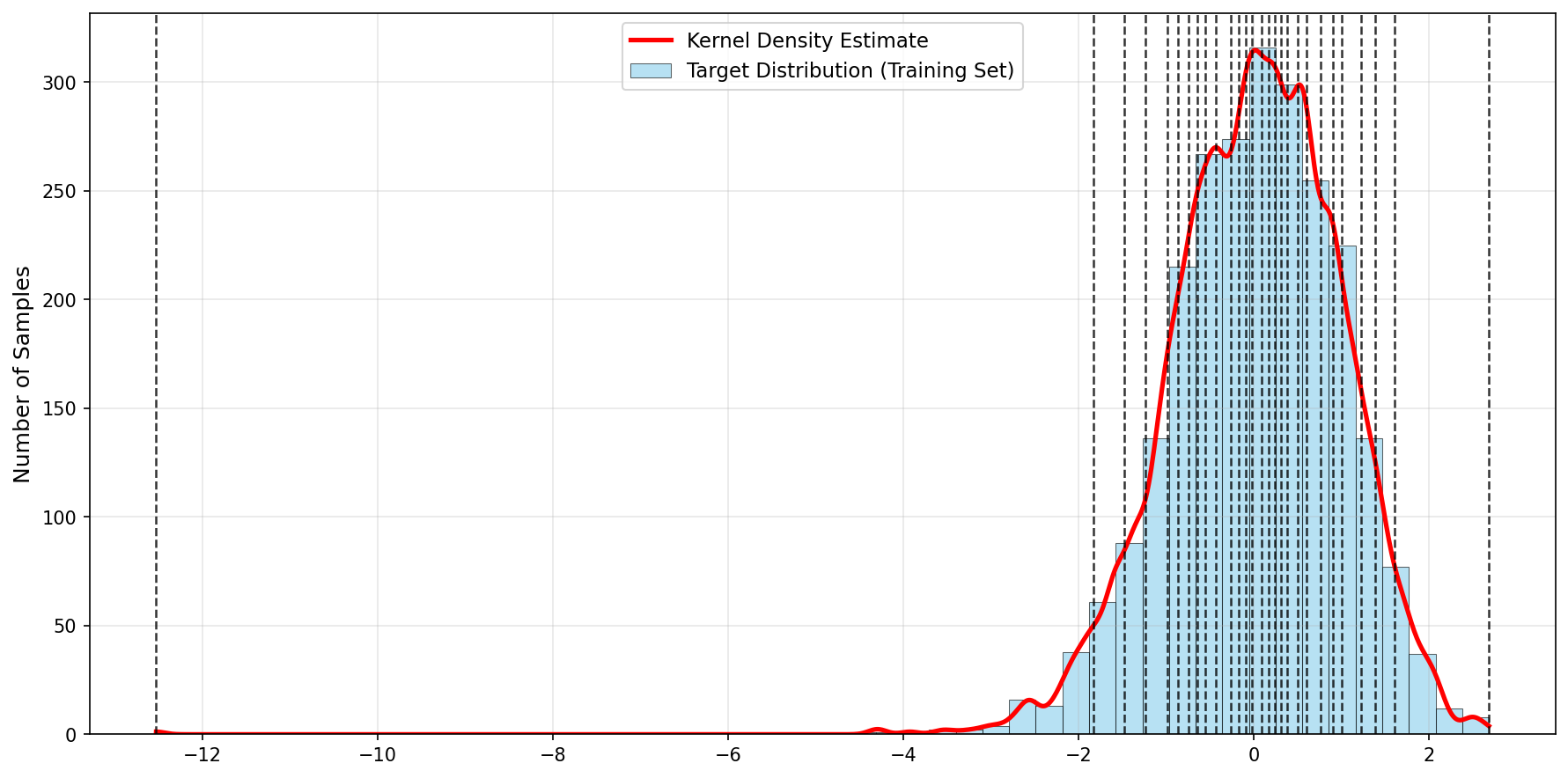}
			\label{fig:binned_spaceGa}
		}
		\hfill
		\subfloat[\centering debutanizer]{
			\includegraphics[width=0.45\textwidth]{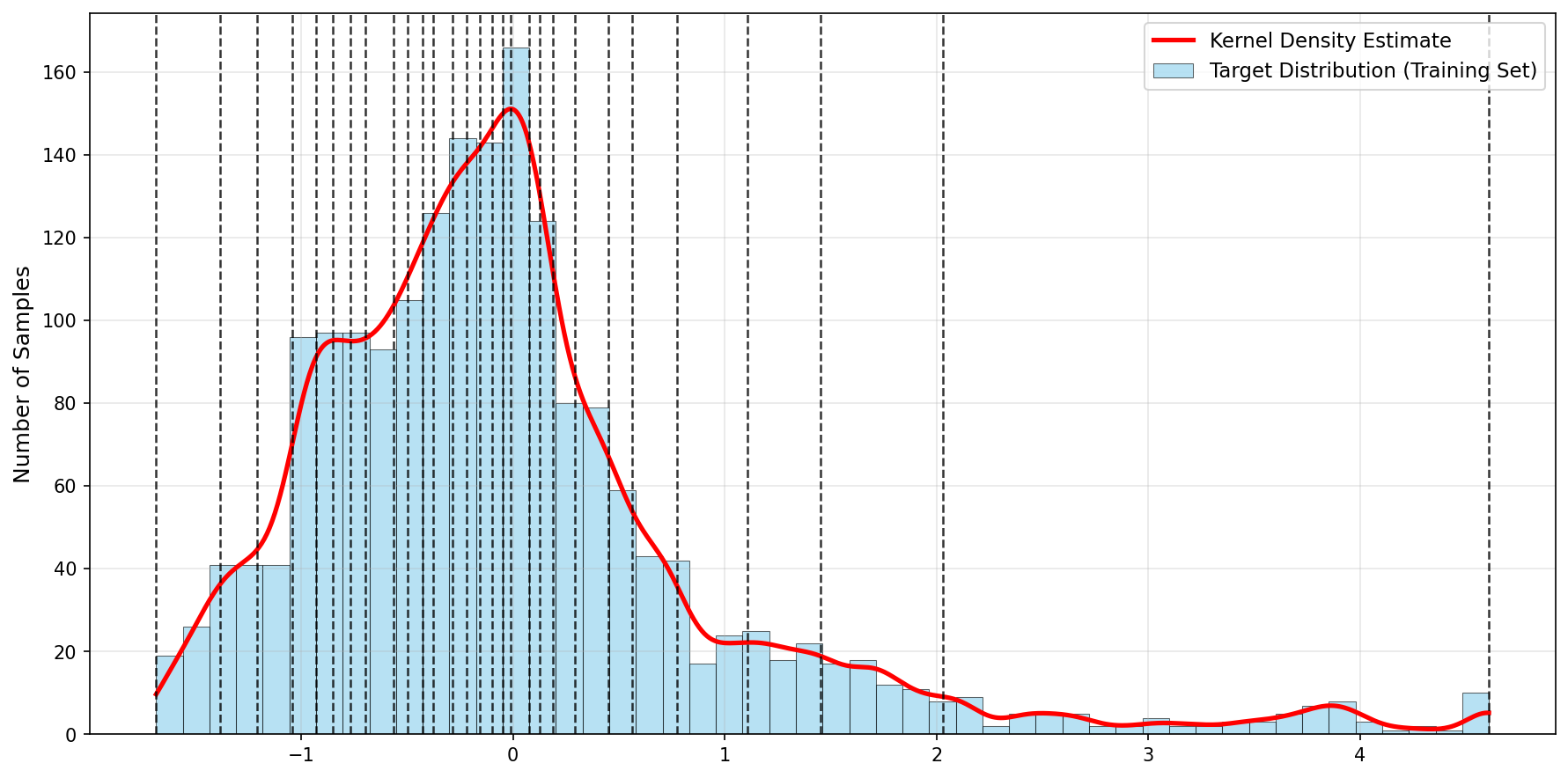}
			\label{fig:binned_debutanizer}
		}
		
		\subfloat[\centering available\_power]{
			\includegraphics[width=0.45\textwidth]{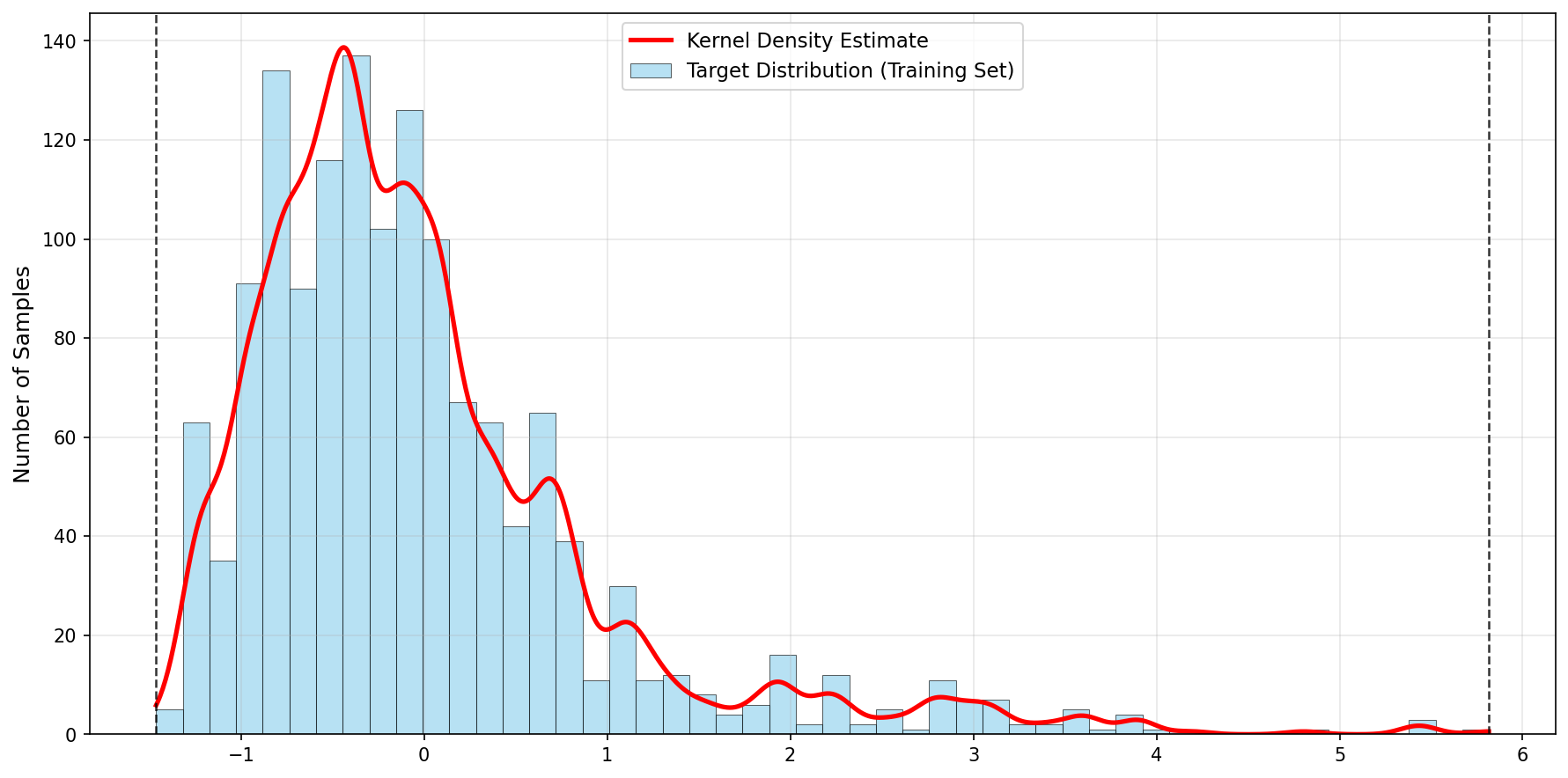}
			\label{fig:binned_available_power}
		}
		\hfill
		\subfloat[\centering maximal\_torque]{
			\includegraphics[width=0.45\textwidth]{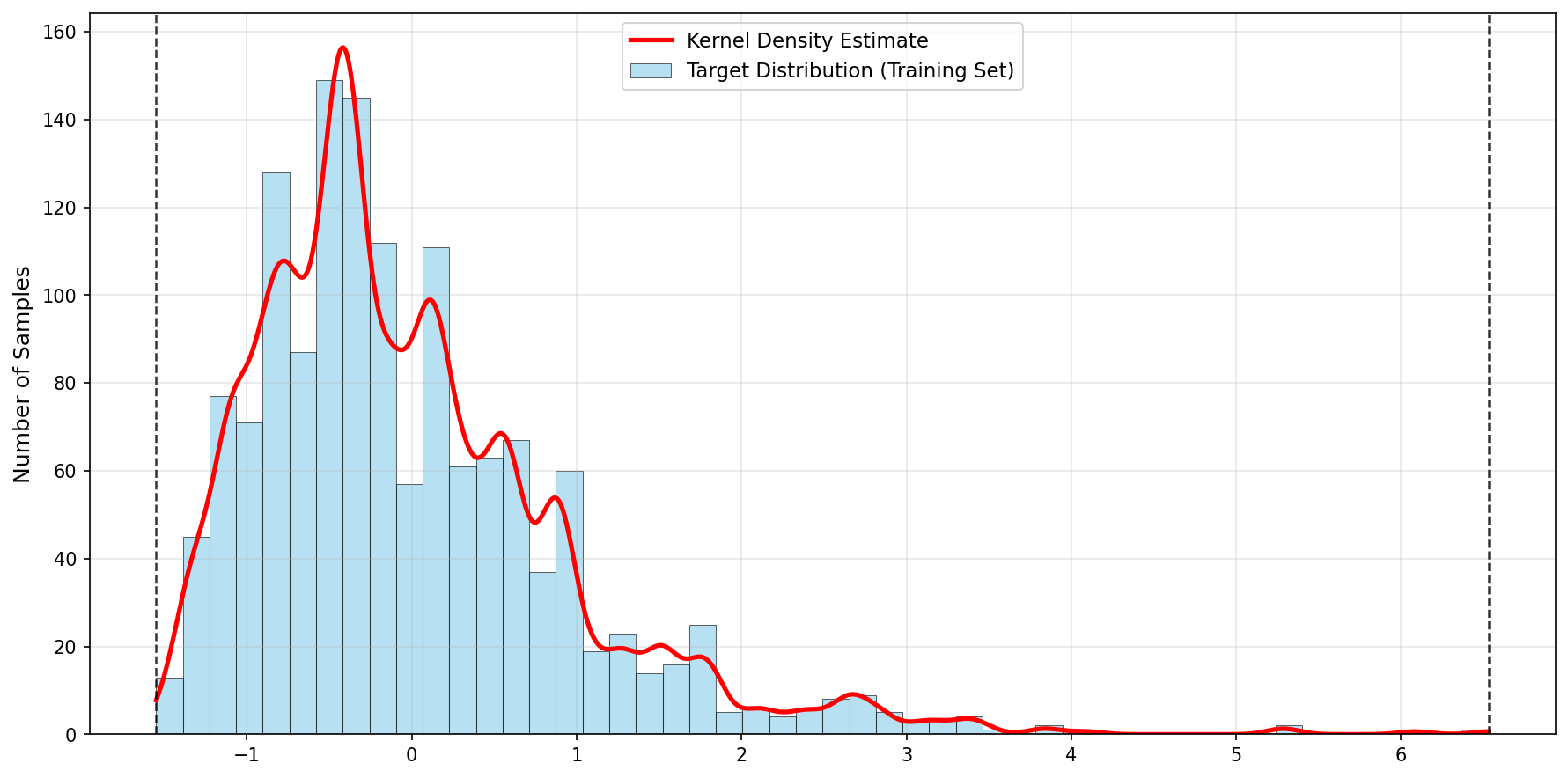}
			\label{fig:binned_maximal_torque}
		}
		
		\caption{Visualization of proposed adaptive bin partitioning method on applied datasets. The histogram and kernel density estimate are shown in standardized scale (z-scores, mean = 0, standard deviation = 1). Vertical dashed lines indicate the adaptive bin edges.}
		\label{fig:adaptive_bin_Partitioning_results}
	\end{figure}
	\FloatBarrier
	
	\begin{figure}[htbp]
		\ContinuedFloat
		\centering
		
		\subfloat[\centering fuel\_consumption\_country]{
			\includegraphics[width=0.45\textwidth]{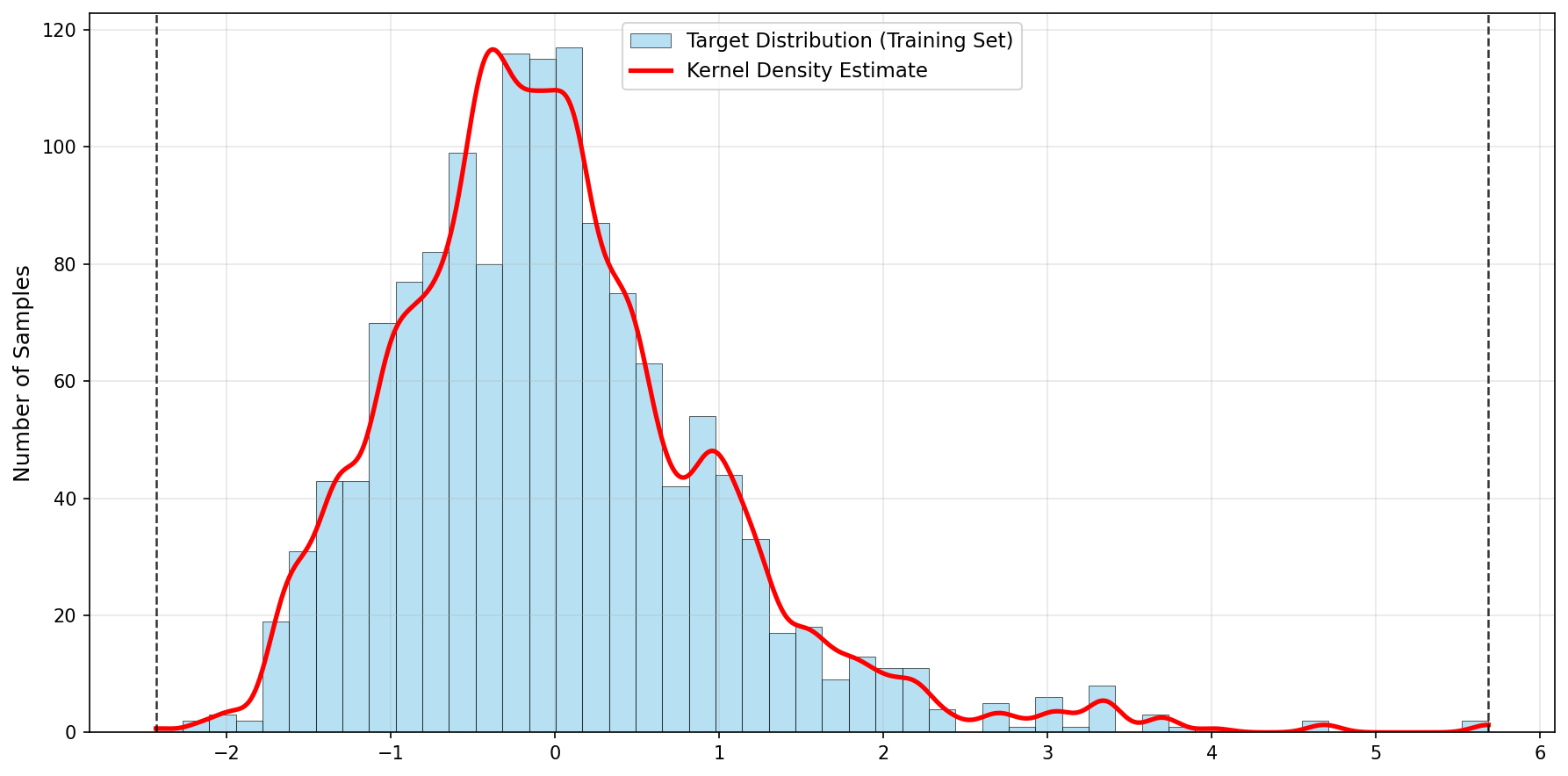}
			\label{fig:bins_fuel_consumption_country}
		}
		\hfill
		\subfloat[\centering acceleration]{
			\includegraphics[width=0.45\textwidth]{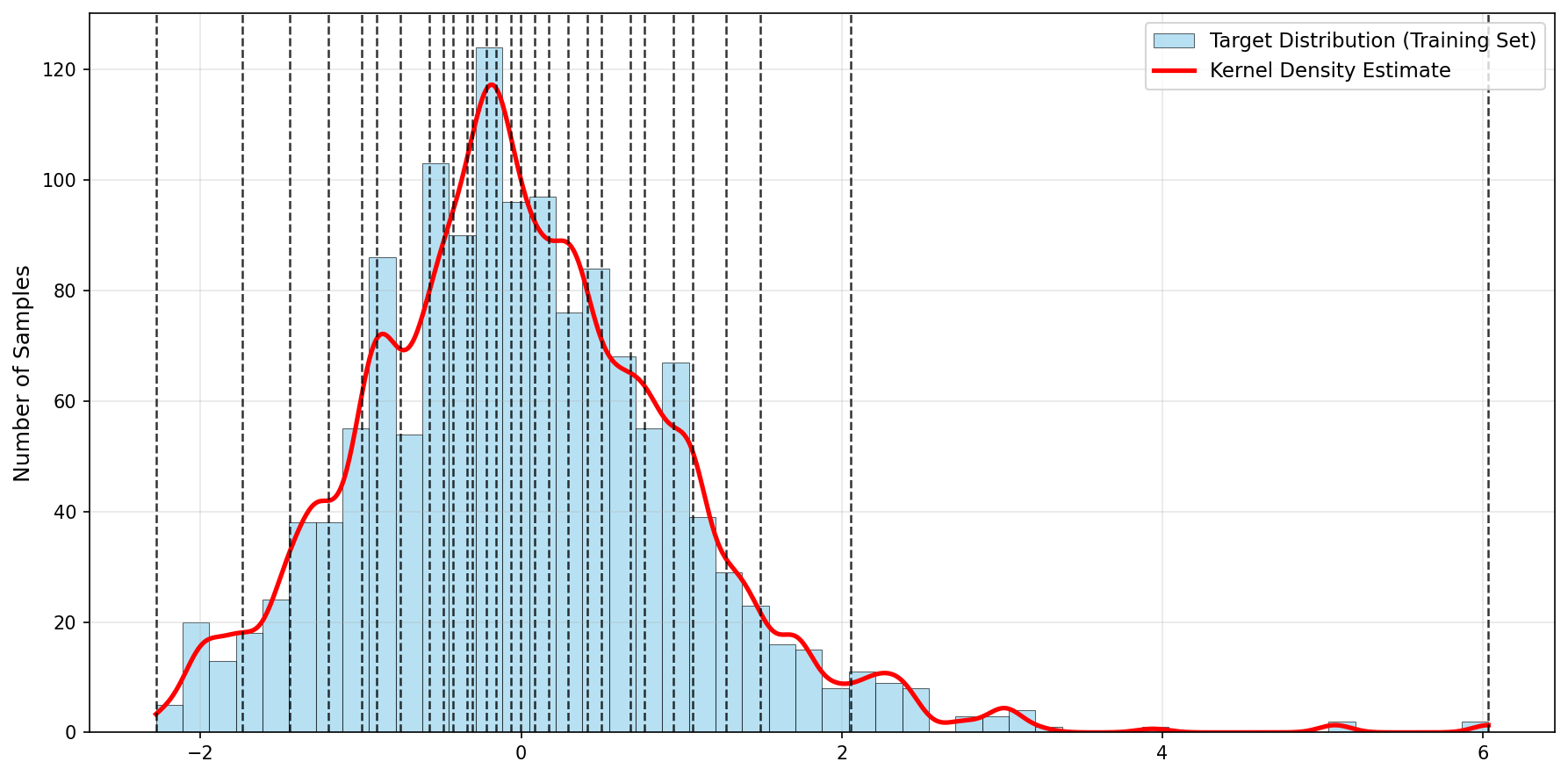}
			\label{fig:binned_acceleration}
		}
		
		\subfloat[\centering airfoild]{
			\includegraphics[width=0.45\textwidth]{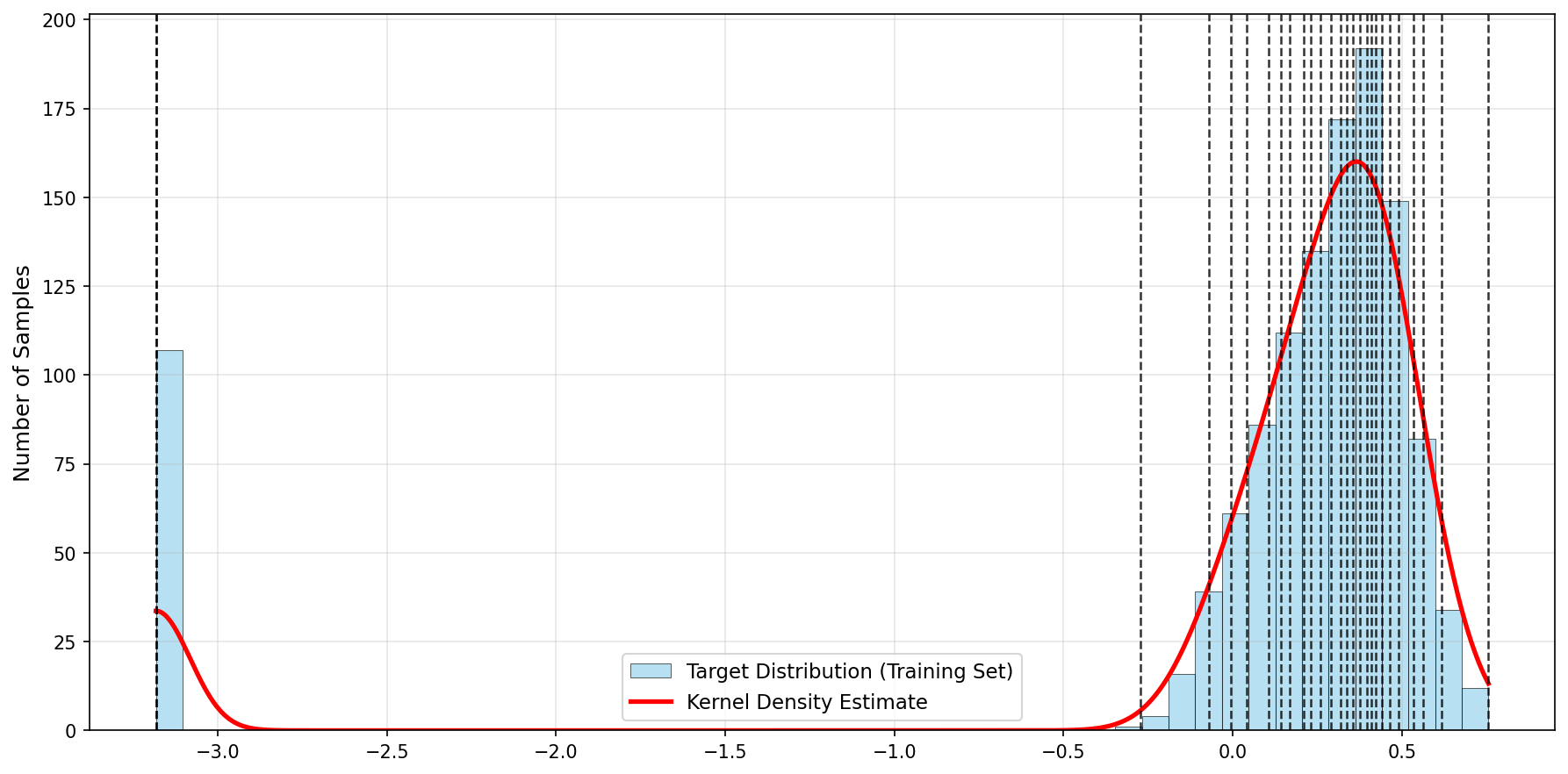}
			\label{fig:binned_airfoild}
		}
		\hfill
		\subfloat[\centering mortgage]{
			\includegraphics[width=0.45\textwidth]{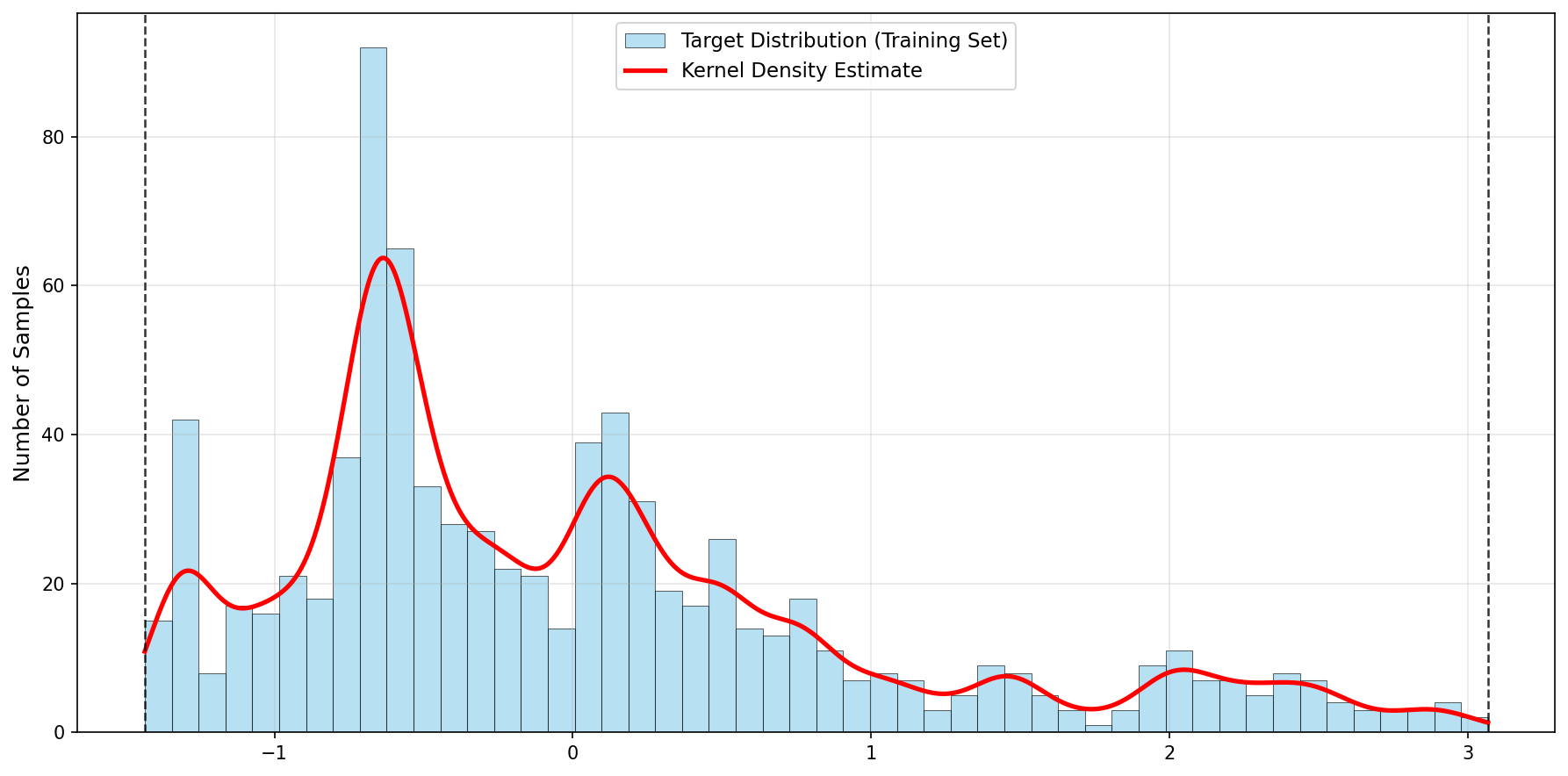}
			\label{fig:binned_mortgage}
		}
		
		\subfloat[\centering treasury]{
			\includegraphics[width=0.45\textwidth]{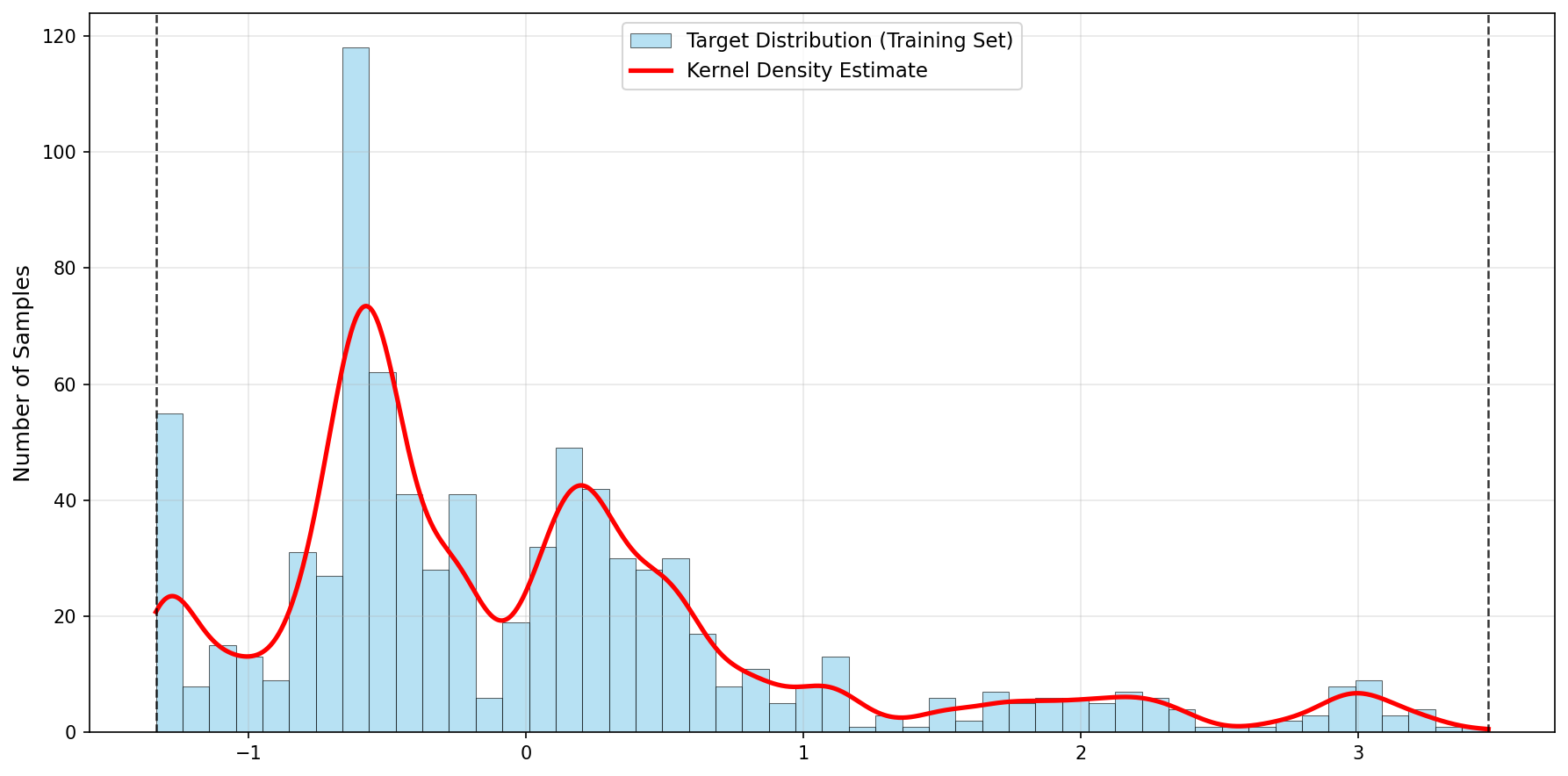}
			\label{fig:binned_treasury}
		}
		\hfill
		\subfloat[\centering concreteStrength]{
			\includegraphics[width=0.45\textwidth]{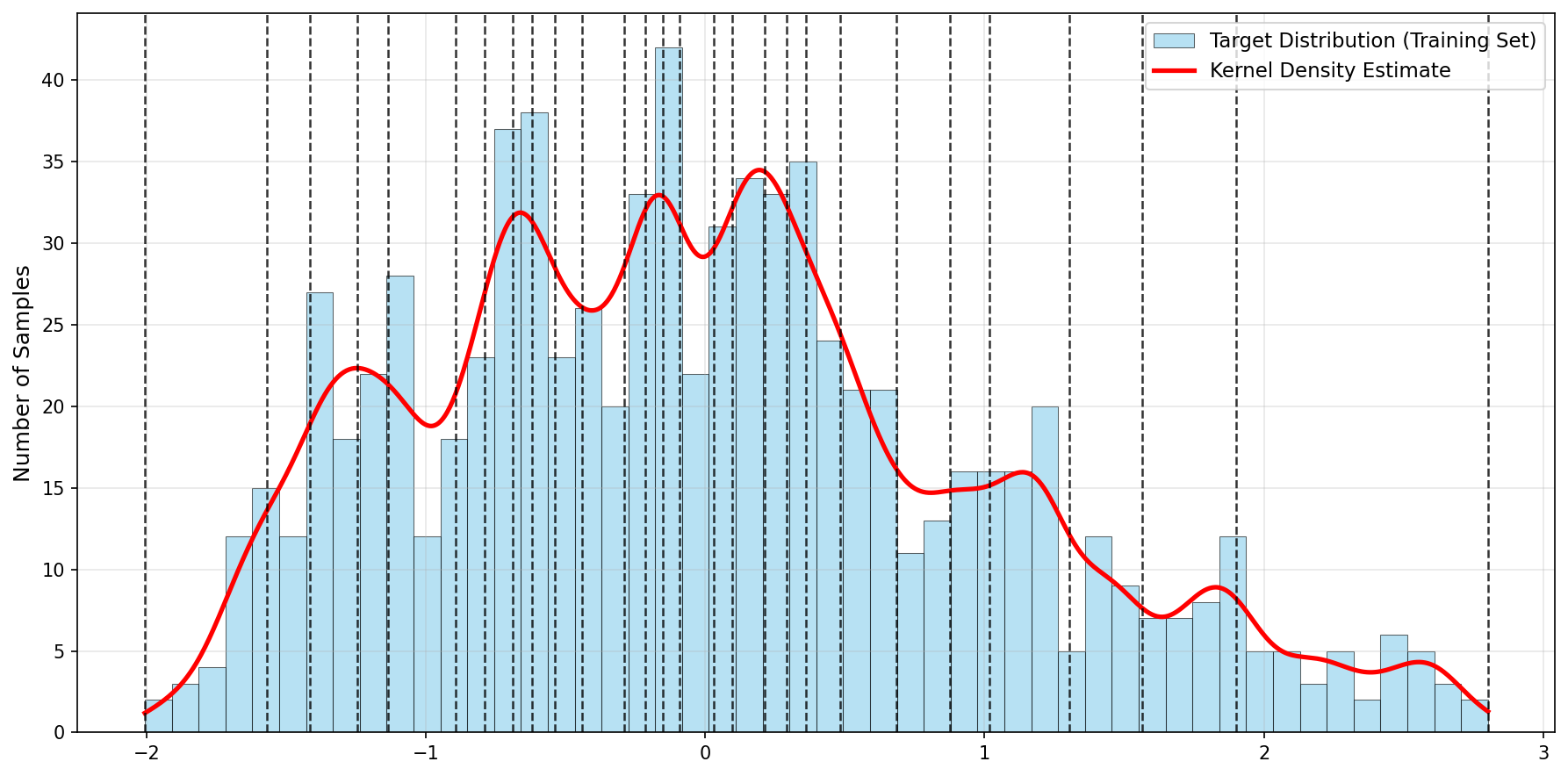}
			\label{fig:binned_concreteStrength}
		}
		
		\caption[]{(continued)}
	\end{figure}

	
	\begin{table}[!ht]
		\centering
		\tiny
		\caption{Adaptive bin partitioning statistics for all datasets. Bin sizes refer to the number of samples per bin, while bin widths indicate the target range covered by each bin}
		\label{tab:bin-partitioning-stats}
		\resizebox{\textwidth}{!}{%
			\begin{tabular}{lcccccccc}
				\hline
				\hline
				\textbf{Dataset} & \ \textbf{No. of samples} & \ \textbf{No. of bins} &\textbf{ Min bin size}  & \textbf{Max bin size} & \textbf{Avg. bin size}  & \textbf{Min bin width} & \textbf{Max bin width} & \textbf{Avg. bin width} \\
				\hline
				\hline
				california & 20,640 & 28 & 496 & 970 & 859.71 & 0.0 & 0.57 & 0.15  \\
				\hline
				compactive & 8,192 & 26 & 197 & 349 & 252.04 & 0.0 & 3.15 & 0.18  \\
				\hline
				cpu\_small & 8,192 & 27 & 197 & 341 & 242.07 & 0.0 & 3.15 & 0.17  \\
				\hline
				heat & 7,400 & 2 & 1,406 & 4,514 & 2960.0 & 2.09 & 6.01 & 4.05  \\
				\hline
				wine\_quality & 6,497 & 10 & 156 & 1,930 & 519.7 & 0.0 & 1.15 & 0.34  \\
				\hline
				abalone & 4,177 & 23 & 101 & 394 & 145.26 & 0.0 & 3.41 & 0.28  \\
				\hline
				space\_ga & 3,107 & 27 & 75 & 147 & 92.04 & 0.07 & 10.7 & 0.56  \\
				\hline
				debutanizer & 2,394 & 28 & 58 & 111 & 68.39 & 0.04 & 2.58 & 0.22  \\
				\hline
				available\_power & 1,802 & 1 & 1,802 & 1,802 & 1,802.0 & 7.28 & 7.28 & 7.28  \\
				\hline
				maximal\_torque & 1,802 & 1 & 1,802 & 1,802 & 1,802.0 & 8.08 & 8.08 & 8.08  \\
				\hline
				fuel\_consumption\_country & 1,764 & 1 & 1,764 & 1,764 & 1,764.0 & 8.12 & 8.12 & 8.12  \\
				\hline
				acceleration  & 1,732 & 28 & 42 & 79 & 49.46 & 0.03 & 3.98 & 0.29  \\
				\hline
				airfoild  & 1,503 & 26 & 37 & 70 & 46.23 & 0.0 & 0.19 & 0.04  \\
				\hline
				mortgage  & 1,049 & 1 & 1,049 & 1,049 & 1,049.0 & 4.5 & 4.5 & 4.5  \\
				\hline
				treasury  & 1,049 & 1 & 1,049 & 1,049 & 1,049.0 & 4.8 & 4.8 & 4.8  \\
				\hline
				concreteStrength  & 1,030 & 27 & 25 & 48 & 30.52 & 0.06 & 0.9 & 0.17  \\
				\hline
			\end{tabular}
		}
	\end{table}


	\subsubsection{Performance Evaluation of the Proposed Data-Level, Algorithm-Level, and Hybrid Approaches}
	This subsection directly addresses \textbf{RQ1}: ``\emph{Does the integration of data-level and algorithm-level balancing strategies within a unified hybrid framework improve predictive performance and generalization in large-scale imbalanced regression tasks compared to using either strategy in isolation?}''
	
	To address RQ1, we evaluate the proposed hybrid framework using several diverse regressors as the final prediction stage: \emph{Multi-Layer Perceptron (MLP)}, \emph{XGBoost (XGB)}, \emph{Linear Regression (LR)}, \emph{K-Nearest Neighbors (KNN)}, \emph{Support Vector Regression (SVR)}, and \emph{Ridge Regression}. As discussed before, our framework operates in five modular phases, forming a regressor-agnostic pipeline. Any base regressor can be seamlessly integrated at the end to leverage the full benefits of the hybrid imbalance correction. The selected regressors span linear and nonlinear models, parametric and instance-based learners, and shallow and deep architectures, ensuring a comprehensive and unbiased assessment of the framework’s effectiveness across varied modeling assumptions and complexity levels. All are applied with default hyperparameters (Table \ref{tab:regressors_default_hyperparams}) to isolate the hybrid imbalance correction’s contribution from model-specific tuning or inherent learner capacity. This deliberate simplicity ensures that performance gains reflect the effectiveness of proposed hybrid imbalance learning approach, not the inherent capacity of any individual learner.
	
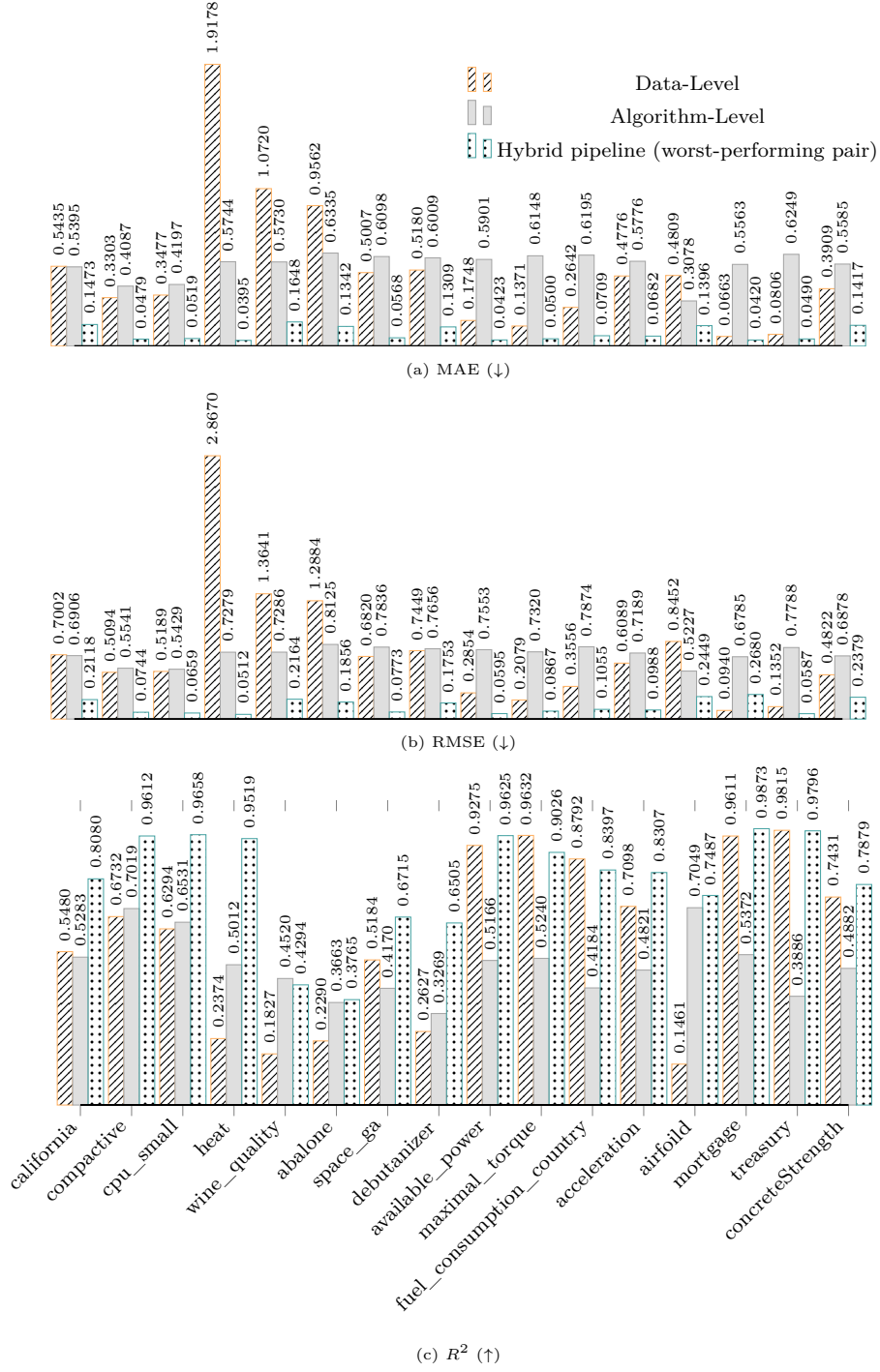
\begin{figure*}[htbp]
		\centering
		
		\subfloat[MAE ($\downarrow$)]{%
			\begin{tikzpicture}
				\begin{axis}[
					ybar=0pt,
					bar width=6pt,
					axis line style={draw=none},
					symbolic x coords={california, compactive, cpu\_small, heat,
						wine\_quality, abalone, space\_ga, debutanizer,
						available\_power, maximal\_torque, fuel\_consumption\_country,
						acceleration, airfoild, mortgage, treasury, concreteStrength},
					xtick=\empty,
					nodes near coords,
					nodes near coords align={vertical},
					every node near coord/.append style={%
						font=\tiny, rotate=90, anchor=west,
						/pgf/number format/.cd, fixed, fixed zerofill, precision=4},
					width=\linewidth, height=5.6cm,
					ymin=0, ymax=2.0,
					x=7mm,
					ytick=\empty,
					legend pos=north east,          
					legend style={font=\scriptsize, draw=none, fill=none},
					legend columns=1,               
					cycle list={%
						{fill=orange!20, draw=orange!60, pattern=north east lines},
						{fill=gray!25,  draw=gray!70},
						{fill=teal!20,  draw=teal!70, pattern=dots}
					},
					every axis plot/.append style={draw,fill}
					]
					\addplot coordinates {
						(california,0.5435) (compactive,0.3303) (cpu\_small,0.3477) (heat,1.9178)
						(wine\_quality,1.0720) (abalone,0.9562) (space\_ga,0.5007) (debutanizer,0.5180)
						(available\_power,0.1748) (maximal\_torque,0.1371) (fuel\_consumption\_country,0.2642)
						(acceleration,0.4776) (airfoild,0.4809) (mortgage,0.0663) (treasury,0.0806)
						(concreteStrength,0.3909)
					}; \addlegendentry{Data-Level}
					
					\addplot coordinates {
						(california,0.5395) (compactive,0.4087) (cpu\_small,0.4197) (heat,0.5744)
						(wine\_quality,0.5730) (abalone,0.6335) (space\_ga,0.6098) (debutanizer,0.6009)
						(available\_power,0.5901) (maximal\_torque,0.6148) (fuel\_consumption\_country,0.6195)
						(acceleration,0.5776) (airfoild,0.3078) (mortgage,0.5563) (treasury,0.6249)
						(concreteStrength,0.5585)
					}; \addlegendentry{Algorithm-Level}
					
					\addplot coordinates {
						(california,0.1473) (compactive,0.0479) (cpu\_small,0.0519) (heat,0.0395)
						(wine\_quality,0.1648) (abalone,0.1342) (space\_ga,0.0568) (debutanizer,0.1309)
						(available\_power,0.0423) (maximal\_torque,0.0500) (fuel\_consumption\_country,0.0709)
						(acceleration,0.0682) (airfoild,0.1396) (mortgage,0.0420) (treasury,0.0490)
						(concreteStrength,0.1417)
					}; \addlegendentry{Hybrid pipeline (worst-performing pair)}
					
					\draw[black,thick] (axis cs:california,0) -- (axis cs:concreteStrength,0);
				\end{axis}
			\end{tikzpicture}%
		}\hfill
		
		\subfloat[RMSE ($\downarrow$)]{%
			\begin{tikzpicture}
				\begin{axis}[
					ybar=0pt,
					bar width=6pt,
					axis line style={draw=none},
					symbolic x coords={california, compactive, cpu\_small, heat,
						wine\_quality, abalone, space\_ga, debutanizer,
						available\_power, maximal\_torque, fuel\_consumption\_country,
						acceleration, airfoild, mortgage, treasury, concreteStrength},
					xtick=\empty,
					nodes near coords,
					nodes near coords align={vertical},
					every node near coord/.append style={%
						font=\tiny, rotate=90, anchor=west,
						/pgf/number format/.cd, fixed, fixed zerofill, precision=4},
					width=\linewidth, height=5.6cm,
					ymin=0, ymax=3.2,
					x=7mm,
					ytick=\empty,
					cycle list={%
						{fill=orange!20, draw=orange!60, pattern=north east lines},
						{fill=gray!25,  draw=gray!70},
						{fill=teal!20,  draw=teal!70, pattern=dots}
					},
					every axis plot/.append style={draw,fill}
					]
					\addplot coordinates {
						(california,0.7002) (compactive,0.5094) (cpu\_small,0.5189) (heat,2.8670)
						(wine\_quality,1.3641) (abalone,1.2884) (space\_ga,0.6820) (debutanizer,0.7449)
						(available\_power,0.2854) (maximal\_torque,0.2079) (fuel\_consumption\_country,0.3556)
						(acceleration,0.6089) (airfoild,0.8452) (mortgage,0.0940) (treasury,0.1352)
						(concreteStrength,0.4822)
					};
					\addplot coordinates {
						(california,0.6906) (compactive,0.5541) (cpu\_small,0.5429) (heat,0.7279)
						(wine\_quality,0.7286) (abalone,0.8125) (space\_ga,0.7836) (debutanizer,0.7656)
						(available\_power,0.7553) (maximal\_torque,0.7320) (fuel\_consumption\_country,0.7874)
						(acceleration,0.7189) (airfoild,0.5227) (mortgage,0.6785) (treasury,0.7788)
						(concreteStrength,0.6878)
					};
					\addplot coordinates {
						(california,0.2118) (compactive,0.0744) (cpu\_small,0.0659) (heat,0.0512)
						(wine\_quality,0.2164) (abalone,0.1856) (space\_ga,0.0773) (debutanizer,0.1753)
						(available\_power,0.0595) (maximal\_torque,0.0867) (fuel\_consumption\_country,0.1055)
						(acceleration,0.0988) (airfoild,0.2449) (mortgage,0.2680) (treasury,0.0587)
						(concreteStrength,0.2379)
					};
					\draw[black,thick] (axis cs:california,0) -- (axis cs:concreteStrength,0);
				\end{axis}
			\end{tikzpicture}%
		}\hfill
		
		\subfloat[$R^2$ ($\uparrow$)]{%
			\begin{tikzpicture}
				\begin{axis}[
					ybar=0pt,
					bar width=6pt,
					axis line style={draw=none},
					symbolic x coords={california, compactive, cpu\_small, heat,
						wine\_quality, abalone, space\_ga, debutanizer,
						available\_power, maximal\_torque, fuel\_consumption\_country,
						acceleration, airfoild, mortgage, treasury, concreteStrength},
					xtick=data,
					x tick label style={rotate=45, anchor=east, font=\scriptsize},
					nodes near coords,
					nodes near coords align={vertical},
					every node near coord/.append style={%
						font=\tiny, rotate=90, anchor=west,
						/pgf/number format/.cd, fixed, fixed zerofill, precision=4},
					width=\linewidth, height=5.6cm,
					ymin=0, ymax=1.05,
					x=7mm,
					ytick=\empty,
					cycle list={%
						{fill=orange!20, draw=orange!60, pattern=north east lines},
						{fill=gray!25,  draw=gray!70},
						{fill=teal!20,  draw=teal!70, pattern=dots}
					},
					every axis plot/.append style={draw,fill}
					]
					\addplot coordinates {
						(california,0.5480) (compactive,0.6732) (cpu\_small,0.6294) (heat,0.2374)
						(wine\_quality,0.1827) (abalone,0.2290) (space\_ga,0.5184) (debutanizer,0.2627)
						(available\_power,0.9275) (maximal\_torque,0.9632) (fuel\_consumption\_country,0.8792)
						(acceleration,0.7098) (airfoild,0.1461) (mortgage,0.9611) (treasury,0.9815)
						(concreteStrength,0.7431)
					};
					\addplot coordinates {
						(california,0.5283) (compactive,0.7019) (cpu\_small,0.6531) (heat,0.5012)
						(wine\_quality,0.4520) (abalone,0.3663) (space\_ga,0.4170) (debutanizer,0.3269)
						(available\_power,0.5166) (maximal\_torque,0.5240) (fuel\_consumption\_country,0.4184)
						(acceleration,0.4821) (airfoild,0.7049) (mortgage,0.5372) (treasury,0.3886)
						(concreteStrength,0.4882)
					};
					\addplot coordinates {
						(california,0.8080) (compactive,0.9612) (cpu\_small,0.9658) (heat,0.9519)
						(wine\_quality,0.4294) (abalone,0.3765) (space\_ga,0.6715) (debutanizer,0.6505)
						(available\_power,0.9625) (maximal\_torque,0.9026) (fuel\_consumption\_country,0.8397)
						(acceleration,0.8307) (airfoild,0.7487) (mortgage,0.9873) (treasury,0.9796)
						(concreteStrength,0.7879)
					};
					\draw[black,thick] (axis cs:california,0) -- (axis cs:concreteStrength,0);
				\end{axis}
			\end{tikzpicture}%
		}
		
		\caption{
			Visual comparison of isolated data-level and algorithm-level balancing methods
			versus the hybrid pipeline (the worst-performing regressor pair is selected
			per dataset from Tables ~\ref{tab:MAE_DATA_ALG_Hybrid_allRegressors}--\ref{tab:R2_DATA_ALG_Hybrid_allRegressors} ). The hybrid approach consistently yields superior MAE ($\downarrow$), RMSE ($\downarrow$),
			and $R^2$ ($\uparrow$) across applied benchmark datasets.
		}
		\label{fig:comparing_the_isolated_data_alg_methods_with_Hybrid}  
	\end{figure*}

	Tables~\ref{tab:MAE_DATA_ALG_Hybrid_allRegressors}, \ref{tab:RMSE_DATA_ALG_Hybrid_allRegressors}, and \ref{tab:R2_DATA_ALG_Hybrid_allRegressors} report the MAE, RMSE, and $R^2$ performance, respectively, of the isolated data-level and algorithm-level methods alongside the proposed hybrid pipeline applied with several regressors. Across all datasets and metrics, the hybrid pipeline consistently outperforms both individual methods. This clearly demonstrates that integrating data- and algorithm-level balancing within a unified framework, substantially improves predictive accuracy and generalization in imbalanced regression tasks. Fig.~\ref{fig:comparing_the_isolated_data_alg_methods_with_Hybrid} provides a visual summary of Tables~\ref{tab:MAE_DATA_ALG_Hybrid_allRegressors}, \ref{tab:RMSE_DATA_ALG_Hybrid_allRegressors}, and \ref{tab:R2_DATA_ALG_Hybrid_allRegressors}. It depicts the robust superiority of the proposed hybrid pipeline across all evaluation metrics and datasets. To underscore its effectiveness, we deliberately selected the worst-performing  paired pipeline+regressor for each dataset, to prove that it still outperforms both the isolated data-level and algorithm-level methods.

	\begin{table}[htbp]
		\centering
		\caption{MAE of isolated data-level and algorithm-level imbalance learning strategies, as well as the proposed hybrid pipeline applied before different regressors. Lower values are better ($\downarrow$).}
		\label{tab:MAE_DATA_ALG_Hybrid_allRegressors}
		\resizebox{\textwidth}{!}{
			\begin{tabular}{l ||cc||cccccc}
				\hline
				\hline
				\textbf{Dataset} &
				\multicolumn{2}{c|}{\textbf{Isolated Methods}} &
				\multicolumn{6}{c|}{\textbf{Hybrid Approaches}} \\
				\cline{2-9}
				&
				\textbf{Data-Level} &
				\textbf{Algorithm-Level} &
				\textbf{Pipeline+MLP} &
				\textbf{Pipeline+XGB} &
				\textbf{Pipeline+LR} &
				\textbf{Pipeline+KNN} &
				\textbf{Pipeline+SVR} &
				\textbf{Pipeline+Ridge} \\
				\hline
				\hline
				california & 0.5435 & 0.5395 & 0.1429 & 0.1446 & 0.1410 & 0.1473 & 0.1397 & 0.1407 \\
				compactive & 0.3303 & 0.4087 & 0.0311 & 0.0415 & 0.0393 & 0.0433 & 0.0479 & 0.0393 \\
				cpu\_small & 0.3477 & 0.4197 & 0.0462 & 0.0348 & 0.0463 & 0.0422 & 0.0519 & 0.0446 \\
				heat & 1.9178 & 0.5744 & 0.0169 & 0.0118 & 0.0040 & 0.0117 & 0.0395 & 0.0044 \\
				wine\_quality & 1.0720 & 0.5730 & 0.1595 & 0.1604 & 0.1648 & 0.1640 & 0.1644 & 0.1640 \\
				abalone & 0.9562 & 0.6335 & 0.1227 & 0.1247 & 0.1257 & 0.1342 & 0.1217 & 0.1249 \\
				space\_ga & 0.5007 & 0.6098 & 0.0531 & 0.0567 & 0.0568 & 0.0543 & 0.0509 & 0.0556 \\
				debutanizer & 0.5180 & 0.6009 & 0.1155 & 0.1219 & 0.1309 & 0.0785 & 0.1223 & 0.1277 \\
				available\_power & 0.1748 & 0.5901 & 0.0336 & 0.0349 & 0.0305 & 0.0197 & 0.0423 & 0.0301 \\
				maximal\_torque & 0.1371 & 0.6148 & 0.0111 & 0.0304 & 0.0395 & 0.0500 & 0.0487 & 0.0229 \\
				fuel\_consumption\_country & 0.2642 & 0.6195 & 0.0581 & 0.0586 & 0.0653 & 0.0709 & 0.0568 & 0.0644 \\
				acceleration & 0.4776 & 0.5776 & 0.0589 & 0.0452 & 0.0678 & 0.0682 & 0.0563 & 0.0658 \\
				airfoild & 0.4809 & 0.3078 & 0.0801 & 0.1095 & 0.1396 & 0.0715 & 0.1305 & 0.1362 \\
				mortgage & 0.0663 & 0.5563 & 0.0254 & 0.0215 & 0.0110 & 0.0152 & 0.0420 & 0.0159 \\
				treasury & 0.0806 & 0.6249 & 0.0290 & 0.0241 & 0.0112 & 0.0119 & 0.0490 & 0.0190 \\
				concreteStrength & 0.3909 & 0.5585 & 0.1103 & 0.1257 & 0.1384 & 0.1417 & 0.1215 & 0.1336 \\
				\hline
				\hline
			\end{tabular}
		}
	\end{table}

	\begin{table}[htbp]
		\centering
		\caption{RMSE of isolated data-level and algorithm-level imbalance learning strategies, as well as the proposed hybrid pipeline applied before different regressors. Lower values are better ($\downarrow$).}
		\label{tab:RMSE_DATA_ALG_Hybrid_allRegressors}
		\resizebox{\textwidth}{!}{
			\begin{tabular}{l ||cc||cccccc}
				\hline
				\hline
				\textbf{Dataset} &
				\multicolumn{2}{c|}{\textbf{Isolated Methods}} &
				\multicolumn{6}{c|}{\textbf{Hybrid approach}} \\
				\cline{2-9}
				&
				\textbf{Data-Level} &
				\textbf{Algorithm-Level} &
				\textbf{Pipeline+MLP} &
				\textbf{Pipeline+XGB} &
				\textbf{Pipeline+LR} &
				\textbf{Pipeline+KNN} &
				\textbf{Pipeline+SVR} &
				\textbf{Pipeline+Ridge} \\
				\hline
				\hline
				california & 0.7002 & 0.6906 & 0.2073 & 0.2052 & 0.1978 & 0.2118 & 0.1974 & 0.1976 \\
				compactive & 0.5094 & 0.5541 & 0.0490 & 0.0717 & 0.0632 & 0.0744 & 0.0541 & 0.0629 \\
				cpu\_small & 0.5189 & 0.5429 & 0.0659 & 0.0477 & 0.0645 & 0.0594 & 0.0606 & 0.0635 \\
				heat & 2.8670 & 0.7279 & 0.0268 & 0.0176 & 0.0064 & 0.0172 & 0.0512 & 0.0074 \\
				wine\_quality & 1.3641 & 0.7286 & 0.2148 & 0.2058 & 0.2093 & 0.2164 & 0.2058 & 0.2084 \\
				abalone & 1.2884 & 0.8125 & 0.1652 & 0.1668 & 0.1646 & 0.1856 & 0.1676 & 0.1636 \\
				space\_ga & 0.6820 & 0.7836 & 0.0729 & 0.0745 & 0.0733 & 0.0773 & 0.0711 & 0.0718 \\
				debutanizer & 0.7449 & 0.7656 & 0.1618 & 0.1751 & 0.1753 & 0.1358 & 0.1697 & 0.1719 \\
				available\_power & 0.2854 & 0.7553 & 0.0512 & 0.0565 & 0.0467 & 0.0578 & 0.0595 & 0.0466 \\
				maximal\_torque & 0.2079 & 0.7320 & 0.0522 & 0.0562 & 0.0582 & 0.0867 & 0.0695 & 0.0397 \\
				fuel\_consumption\_country & 0.3556 & 0.7874 & 0.0859 & 0.0779 & 0.0871 & 0.1055 & 0.0792 & 0.0855 \\
				acceleration & 0.6089 & 0.7189 & 0.0874 & 0.0582 & 0.0880 & 0.0988 & 0.0742 & 0.0854 \\
				airfoild & 0.8452 & 0.5227 & 0.2205 & 0.1872 & 0.2292 & 0.1814 & 0.2449 & 0.2301 \\
				mortgage & 0.0940 & 0.6785 & 0.0340 & 0.0319 & 0.0165 & 0.0268 & 0.0497 & 0.0236 \\  
				treasury & 0.1352 & 0.7788 & 0.0458 & 0.0501 & 0.0314 & 0.0393 & 0.0587 & 0.0324 \\
				concreteStrength & 0.4822 & 0.6878 & 0.1329 & 0.1203 & 0.1738 & 0.1842 & 0.1549 & 0.2379 \\
				\hline
				\hline
			\end{tabular}
		}
	\end{table}
	
	\begin{table}[htbp]
		\centering
		\caption{$R^2$ of isolated data-level and algorithm-level imbalance learning strategies, compared with the proposed hybrid pipeline applied before six different regressors. Higher values are better ($\uparrow$).}
		\label{tab:R2_DATA_ALG_Hybrid_allRegressors}
		\resizebox{\textwidth}{!}{
			\begin{tabular}{l ||cc||cccccc}
				\hline
				\hline
				\textbf{Dataset} &
				\multicolumn{2}{c|}{\textbf{Isolated Methods}} &
				\multicolumn{6}{c|}{\textbf{Hybrid approach}} \\
				\cline{2-9}
				&
				\textbf{Data-Level} &
				\textbf{Algorithm-Level} &
				\textbf{Pipeline+MLP} &
				\textbf{Pipeline+XGB} &
				\textbf{Pipeline+LR} &
				\textbf{Pipeline+KNN} &
				\textbf{Pipeline+SVR} &
				\textbf{Pipeline+Ridge} \\
				\hline
				\hline
				california & 0.5480 & 0.5283 & 0.8168 & 0.8110 & 0.8298 & 0.8080 & 0.8312 & 0.8307 \\
				compactive & 0.6732 & 0.7019 & 0.9847 & 0.9657 & 0.9719 & 0.9612 & 0.9759 & 0.9723 \\  
				cpu\_small & 0.6294 & 0.6531 & 0.9709 & 0.9797 & 0.9658 & 0.9714 & 0.9675 & 0.9680 \\
				heat & 0.2374 & 0.5012 & 0.9892 & 0.9952 & 0.9994 & 0.9963 & 0.9519 & 0.9992 \\
				wine\_quality & 0.1827 & 0.4520 & 0.4375 & 0.4837 & 0.4659 & 0.4294 & 0.4705 & 0.4707 \\
				abalone & 0.2290 & 0.3663 & 0.5060 & 0.4962 & 0.5093 & 0.3765 & 0.4913 & 0.5152 \\
				space\_ga & 0.5184 & 0.4170 & 0.7123 & 0.6951 & 0.7052 & 0.6715 & 0.7164 & 0.7168 \\
				debutanizer & 0.2627 & 0.3269 & 0.7024 & 0.6514 & 0.6505 & 0.7689 & 0.6727 & 0.6643 \\
				available\_power & 0.9275 & 0.5166 & 0.9706 & 0.9642 & 0.9755 & 0.9625 & 0.9671 & 0.9756 \\
				maximal\_torque & 0.9632 & 0.5240 & 0.9646 & 0.9598 & 0.9816 & 0.9026 & 0.9343 & 0.9796 \\
				fuel\_consumption\_country & 0.8792 & 0.4184 & 0.8849 & 0.8997 & 0.8827 & 0.8397 & 0.8997 & 0.8870 \\
				acceleration & 0.7098 & 0.4821 & 0.8892 & 0.9186 & 0.8657 & 0.8307 & 0.8865 & 0.8737 \\
				airfoild & 0.1461 & 0.7049 & 0.7963 & 0.8413 & 0.7800 & 0.8622 & 0.7487 & 0.7783 \\
				mortgage & 0.9611 & 0.5372 & 0.9934 & 0.9948 & 0.9984 & 0.9960 & 0.9873 & 0.9972 \\
				treasury & 0.9815 & 0.3886 & 0.9878 & 0.9854 & 0.9936 & 0.9928 & 0.9796 & 0.9935 \\
				concreteStrength & 0.7431 & 0.4882 & 0.8751 & 0.8364 & 0.8112 & 0.7879 & 0.8316 & 0.8244 \\
				\hline
				\hline
			\end{tabular}
		}
	\end{table}

	\subsection{Discussion and Answers to Research Questions}
	\label{sec:discussion}
	This study introduces a unified hybrid framework for imbalanced regression, systematically integrating data-level and algorithm-level strategies to overcome their individual limitations and improve performance across the full target distribution. We structure this section around the remained RQs from the \hyperref[sec:Introduction]{Introduction}, analyzing how each component of the proposed pipeline enhances predictive accuracy in imbalanced regression scenarios.
	
	\textbf{RQ2:} ``\textit{How does the proposed hybrid pipeline perform in imbalanced regression compared to standard regression baselines without imbalance handling?}''
	
	To answer RQ2, we applied the same regressors from RQ1 (MLP, XGBoost, Linear Regression, KNN, SVR, and Ridge) both \textit{without} any imbalance correction (standard baselines) and \textit{integrated} into the proposed hybrid pipeline. Tables~\ref{tab:MAE_compare_with_Baseline_single_regressors}--\ref{tab:R2_compare_with_Baseline_single_regressors} report MAE, RMSE, and $R^2$ for each regressor in isolation and when integrated with the hybrid pipeline across all datasets. The last column shows the average percent Improvement of the hybrid pipeline over the corresponding standalone regressor. As shown, in every dataset and for every metric, the hybrid pipeline paired with a given regressor consistently outperforms the same regressor operating alone. This demonstrates that the proposed framework not only mitigates imbalance but also enhances the predictive capability of standard models. 
	
	The consistent superiority of the proposed pipeline stems from the synergistic contributions of its five phases. Adaptive bin partitioning explicitly reveals imbalance structure in the target space, while representation learning enhances latent separability of minority regions. Subsequent data-level balancing and algorithm-level cost reweighting jointly mitigate sample scarcity and prediction bias, and the final hybrid fusion integrates these benefits into a more expressive and balanced learning signal. As a result, models become capable of capturing informative variability across the full target range rather than overfitting majority-dense areas. This improvement is most pronounced for traditional linear regressors, such as Linear and Ridge regression, which inherently lack nonlinear flexibility to recover minority-region structure; the enriched and balanced feature-target distribution provided by the pipeline therefore yields substantial MAE/RMSE reductions and $R^2$ gains. Nevertheless, even nonlinear models (XGBoost, MLP) experience measurable generalization boosts, demonstrating that the hybrid approach complements (rather than replaces) their modeling capacity. Overall, the five-phase framework acts as a regressor-agnostic enhancement module that systematically strengthens regression performance under severe imbalance.

	\begin{table*}[htbp]
		\centering
		\caption{MAE ($\downarrow$) results for regressors without (gray columns) and with the proposed hybrid pipeline.}
		\label{tab:MAE_compare_with_Baseline_single_regressors}
		\resizebox{\textwidth}{!}{%
			\begin{tabular}{
					l
					>{\columncolor{gray!15}}c >{\columncolor{white}}c
					>{\columncolor{gray!15}}c >{\columncolor{white}}c
					>{\columncolor{gray!15}}c >{\columncolor{white}}c
					>{\columncolor{gray!15}}c >{\columncolor{white}}c
					>{\columncolor{gray!15}}c >{\columncolor{white}}c
					>{\columncolor{gray!15}}c >{\columncolor{white}}c
					c
				}
				\toprule
				\textbf{Dataset} &
				\multicolumn{2}{c}{\textbf{MLP}} &
				\multicolumn{2}{c}{\textbf{XGB}} &
				\multicolumn{2}{c}{\textbf{LR}} &
				\multicolumn{2}{c}{\textbf{KNN}} &
				\multicolumn{2}{c}{\textbf{SVR}} &
				\multicolumn{2}{c}{\textbf{Ridge}} &
				\textbf{Avg.\ \% of} \\
				\cmidrule(lr){2-3}\cmidrule(lr){4-5}\cmidrule(lr){6-7}
				\cmidrule(lr){8-9}\cmidrule(lr){10-11}\cmidrule(lr){12-13}
				& Single & Pipeline & Single & Pipeline & Single & Pipeline &
				Single & Pipeline & Single & Pipeline & Single & Pipeline &
				\textbf{Improvement} \\
				\midrule
				california        & 0.1504 & 0.1429 & 0.1833 & 0.1446 & 0.2115 & 0.1410 & 0.1759 & 0.1473 & 0.1696 & 0.1397 & 0.2119 & 0.1407 & \textbf{21.15\%} \\
				compactive        & 0.0367 & 0.0311 & 0.0541 & 0.0415 & 0.1255 & 0.0393 & 0.0687 & 0.0433 & 0.0500 & 0.0479 & 0.1262 & 0.0393 & \textbf{36.21\%} \\
				cpu\_small        & 0.0519 & 0.0462 & 0.0380 & 0.0348 & 0.1247 & 0.0463 & 0.0443 & 0.0422 & 0.0559 & 0.0519 & 0.1254 & 0.0446 & \textbf{26.43\%} \\
				heat              & 0.0182 & 0.0169 & 0.0583 & 0.0118 & 0.0734 & 0.0040 & 0.0120 & 0.0117 & 0.0492 & 0.0395 & 0.0734 & 0.0044 & \textbf{49.61\%} \\
				wine\_quality     & 0.1776 & 0.1595 & 0.1772 & 0.1604 & 0.1886 & 0.1648 & 0.1763 & 0.1640 & 0.1768 & 0.1644 & 0.1886 & 0.1640 & \textbf{9.89\%} \\
				abalone           & 0.1338 & 0.1227 & 0.1274 & 0.1247 & 0.1500 & 0.1257 & 0.1417 & 0.1342 & 0.1286 & 0.1217 & 0.1500 & 0.1249 & \textbf{9.00\%} \\
				space\_ga         & 0.0566 & 0.0531 & 0.0623 & 0.0567 & 0.0675 & 0.0568 & 0.0571 & 0.0543 & 0.0559 & 0.0509 & 0.0678 & 0.0556 & \textbf{10.48\%} \\
				debutanizer       & 0.1244 & 0.1155 & 0.1252 & 0.1219 & 0.1905 & 0.1309 & 0.0913 & 0.0785 & 0.1376 & 0.1223 & 0.1909 & 0.1277 & \textbf{16.55\%} \\
				available\_power  & 0.0403 & 0.0336 & 0.0395 & 0.0349 & 0.0501 & 0.0305 & 0.0287 & 0.0197 & 0.0489 & 0.0423 & 0.0501 & 0.0301 & \textbf{25.36\%} \\
				maximal\_torque   & 0.0121 & 0.0111 & 0.0325 & 0.0304 & 0.0430 & 0.0395 & 0.0579 & 0.0500 & 0.0597 & 0.0487 & 0.0428 & 0.0229 & \textbf{16.91\%} \\
				fuel\_consumption\_country & 0.0635 & 0.0581 & 0.0602 & 0.0586 & 0.0674 & 0.0653 & 0.0752 & 0.0709 & 0.0638 & 0.0568 & 0.0711 & 0.0644 & \textbf{6.73\%} \\
				acceleration      & 0.0641 & 0.0589 & 0.0530 & 0.0452 & 0.0836 & 0.0678 & 0.0755 & 0.0682 & 0.0657 & 0.0563 & 0.0771 & 0.0658 & \textbf{25.56\%} \\
				airfoild          & 0.1674 & 0.0801 & 0.1105 & 0.1095 & 0.2796 & 0.1396 & 0.0866 & 0.0715 & 0.1710 & 0.1305 & 0.2793 & 0.1362 & \textbf{32.58\%} \\
				mortgage          & 0.0289 & 0.0254 & 0.0484 & 0.0215 & 0.0127 & 0.0110 & 0.0164 & 0.0152 & 0.0435 & 0.0420 & 0.0186 & 0.0115 & \textbf{21.67\%} \\
				treasury          & 0.0363 & 0.0290 & 0.0509 & 0.0241 & 0.0190 & 0.0112 & 0.0162 & 0.0119 & 0.0503 & 0.0490 & 0.0201 & 0.0190 & \textbf{24.74\%} \\
				concreteStrength  & 0.1153 & 0.1103 & 0.1299 & 0.1257 & 0.1930 & 0.1384 & 0.1753 & 0.1417 & 0.1303 & 0.1215 & 0.1937 & 0.1336 & \textbf{15.47\%} \\
				\bottomrule
			\end{tabular}%
		}%
		\bigskip
		\parbox{\textwidth}{\raggedleft\itshape Average enhancement: \textbf{23.15\%}}
	\end{table*}

	\begin{table*}[htbp]
		\centering
		\caption{RMSE ($\downarrow$) results for regressors without (gray columns) and with the proposed hybrid pipeline.}
		\label{tab:RMSE_compare_with_Baseline_single_regressors}
		\resizebox{\textwidth}{!}{%
			\begin{tabular}{
					l
					>{\columncolor{gray!15}}c >{\columncolor{white}}c
					>{\columncolor{gray!15}}c >{\columncolor{white}}c
					>{\columncolor{gray!15}}c >{\columncolor{white}}c
					>{\columncolor{gray!15}}c >{\columncolor{white}}c
					>{\columncolor{gray!15}}c >{\columncolor{white}}c
					>{\columncolor{gray!15}}c >{\columncolor{white}}c
					c
				}
				\toprule
				\textbf{Dataset} &
				\multicolumn{2}{c}{\textbf{MLP}} &
				\multicolumn{2}{c}{\textbf{XGB}} &
				\multicolumn{2}{c}{\textbf{LR}} &
				\multicolumn{2}{c}{\textbf{KNN}} &
				\multicolumn{2}{c}{\textbf{SVR}} &
				\multicolumn{2}{c}{\textbf{Ridge}} &
				\textbf{Avg.\ \% of} \\
				\cmidrule(lr){2-3}\cmidrule(lr){4-5}\cmidrule(lr){6-7}
				\cmidrule(lr){8-9}\cmidrule(lr){10-11}\cmidrule(lr){12-13}
				& Single & Pipeline & Single & Pipeline & Single & Pipeline &
				Single & Pipeline & Single & Pipeline & Single & Pipeline &
				\textbf{Improvement} \\
				\midrule
				california        & 0.2243 & 0.2043 & 0.2517 & 0.2052 & 0.2879 & 0.1978 & 0.2631 & 0.2118 & 0.2497 & 0.1974 & 0.2883 & 0.1976 & \textbf{21.77\%} \\
				compactive        & 0.0512 & 0.0490 & 0.0732 & 0.0717 & 0.1904 & 0.0632 & 0.0948 & 0.0744 & 0.0640 & 0.0541 & 0.1908 & 0.0629 & \textbf{29.53\%} \\
				cpu\_small        & 0.0731 & 0.0659 & 0.0549 & 0.0477 & 0.1874 & 0.0645 & 0.0638 & 0.0594 & 0.0696 & 0.0606 & 0.1876 & 0.0629 & \textbf{29.14\%} \\
				heat              & 0.0296 & 0.0268 & 0.0856 & 0.0176 & 0.1056 & 0.0064 & 0.0187 & 0.0172 & 0.0573 & 0.0512 & 0.1056 & 0.0074 & \textbf{49.08\%} \\
				wine\_quality     & 0.2288 & 0.2148 & 0.2264 & 0.2058 & 0.2465 & 0.2093 & 0.2366 & 0.2164 & 0.2268 & 0.2085 & 0.2463 & 0.2084 & \textbf{10.38\%} \\
				abalone           & 0.1879 & 0.1652 & 0.1801 & 0.1668 & 0.2080 & 0.1646 & 0.1994 & 0.1856 & 0.1852 & 0.1676 & 0.2080 & 0.1636 & \textbf{13.02\%} \\
				space\_ga         & 0.0751 & 0.0729 & 0.0866 & 0.0745 & 0.0918 & 0.0733 & 0.0831 & 0.0773 & 0.0739 & 0.0711 & 0.0915 & 0.0718 & \textbf{11.56\%} \\
				debutanizer       & 0.1856 & 0.1618 & 0.1904 & 0.1751 & 0.2735 & 0.1753 & 0.1473 & 0.1358 & 0.2183 & 0.1697 & 0.2733 & 0.1719 & \textbf{20.66\%} \\
				available\_power  & 0.0591 & 0.0512 & 0.0676 & 0.0565 & 0.0765 & 0.0476 & 0.0606 & 0.0578 & 0.0638 & 0.0595 & 0.0766 & 0.0466 & \textbf{19.68\%} \\
				maximal\_torque   & 0.0598 & 0.0522 & 0.0608 & 0.0562 & 0.0637 & 0.0582 & 0.1075 & 0.0867 & 0.0744 & 0.0695 & 0.0637 & 0.0397 & \textbf{15.42\%} \\
				fuel\_consumption\_country & 0.0880 & 0.0859 & 0.0818 & 0.0779 & 0.0923 & 0.0871 & 0.1074 & 0.1055 & 0.0825 & 0.0792 & 0.0950 & 0.0855 & \textbf{4.76\%} \\
				acceleration      & 0.0843 & 0.0874 & 0.0698 & 0.0582 & 0.1393 & 0.0880 & 0.1038 & 0.0988 & 0.0849 & 0.0742 & 0.0998 & 0.0854 & \textbf{13.60\%} \\
				airfoild          & 0.2981 & 0.2205 & 0.1994 & 0.1872 & 0.4773 & 0.2292 & 0.2136 & 0.1814 & 0.4327 & 0.2449 & 0.4772 & 0.2301 & \textbf{32.40\%} \\
				mortgage          & 0.0394 & 0.0340 & 0.0662 & 0.0319 & 0.0188 & 0.0165 & 0.0284 & 0.0268 & 0.0516 & 0.0497 & 0.0264 & 0.0236 & \textbf{16.28\%} \\
				treasury          & 0.0559 & 0.0458 & 0.0767 & 0.0501 & 0.0343 & 0.0301 & 0.0408 & 0.0393 & 0.0618 & 0.0587 & 0.0355 & 0.0324 & \textbf{4.02\%} \\
				concreteStrength  & 0.1471 & 0.1329 & 0.1662 & 0.1203 & 0.2441 & 0.1738 & 0.2269 & 0.1842 & 0.1684 & 0.1549 & 0.2440 & 0.2379 & \textbf{15.90\%} \\
				\bottomrule
			\end{tabular}%
		}%
		\bigskip
		\parbox{\textwidth}{\raggedleft\itshape Average enhancement: \textbf{23.15\%}}
	\end{table*}

	\begin{table*}[htbp]
		\centering
		\caption{R\textsuperscript{2} ($\uparrow$) results for regressors without (gray columns) and with the proposed hybrid pipeline.}
		\label{tab:R2_compare_with_Baseline_single_regressors}
		\resizebox{\textwidth}{!}{%
			\begin{tabular}{
					l
					>{\columncolor{gray!15}}c >{\columncolor{white}}c
					>{\columncolor{gray!15}}c >{\columncolor{white}}c
					>{\columncolor{gray!15}}c >{\columncolor{white}}c
					>{\columncolor{gray!15}}c >{\columncolor{white}}c
					>{\columncolor{gray!15}}c >{\columncolor{white}}c
					>{\columncolor{gray!15}}c >{\columncolor{white}}c
					c
				}
				\toprule
				\textbf{Dataset} &
				\multicolumn{2}{c}{\textbf{MLP}} &
				\multicolumn{2}{c}{\textbf{XGB}} &
				\multicolumn{2}{c}{\textbf{LR}} &
				\multicolumn{2}{c}{\textbf{KNN}} &
				\multicolumn{2}{c}{\textbf{SVR}} &
				\multicolumn{2}{c}{\textbf{Ridge}} &
				\textbf{Avg.\ \% of} \\
				\cmidrule(lr){2-3}\cmidrule(lr){4-5}\cmidrule(lr){6-7}
				\cmidrule(lr){8-9}\cmidrule(lr){10-11}\cmidrule(lr){12-13}
				& Single & Pipeline & Single & Pipeline & Single & Pipeline &
				Single & Pipeline & Single & Pipeline & Single & Pipeline &
				\textbf{Improvement} \\
				\midrule
				california & 0.7802 & 0.8168 & 0.7233 & 0.8110 & 0.6380 & 0.8298 & 0.6976 & 0.8080 & 0.7276 & 0.8312 & 0.6370 & 0.8307 & \textbf{17.89\%} \\
				compactive & 0.9816 & 0.9847 & 0.9624 & 0.9657 & 0.7452 & 0.9719 & 0.9368 & 0.9612 & 0.9713 & 0.9759 & 0.7443 & 0.9723 & \textbf{10.80\%} \\
				cpu\_small & 0.9563 & 0.9709 & 0.9779 & 0.9797 & 0.7128 & 0.9658 & 0.9667 & 0.9714 & 0.9603 & 0.9675 & 0.7122 & 0.9680 & \textbf{12.39\%} \\
				heat & 0.9864 & 0.9892 & 0.8867 & 0.9952 & 0.9946 & 0.9963 & 0.9946 & 0.9963 & 0.9492 & 0.9519 & 0.8276 & 0.9992 & \textbf{5.65\%} \\
				wine\_quality & 0.3618 & 0.4375 & 0.3753 & 0.4837 & 0.2598 & 0.4659 & 0.3179 & 0.4294 & 0.3732 & 0.4705 & 0.2605 & 0.4707 & \textbf{45.16\%} \\
				abalone & 0.3609 & 0.5060 & 0.4129 & 0.4962 & 0.2169 & 0.5093 & 0.2805 & 0.3765 & 0.3788 & 0.4913 & 0.2169 & 0.5152 & \textbf{66.11\%} \\
				space\_ga & 0.6903 & 0.7123 & 0.5885 & 0.6951 & 0.5367 & 0.7052 & 0.6203 & 0.6715 & 0.7000 & 0.7164 & 0.5398 & 0.7168 & \textbf{16.01\%} \\
				debutanizer & 0.6083 & 0.7024 & 0.5878 & 0.6514 & 0.1497 & 0.6505 & 0.7533 & 0.7689 & 0.4582 & 0.6727 & 0.1509 & 0.6643 & \textbf{124.99\%} \\
				available\_power & 0.9608 & 0.9706 & 0.9487 & 0.9642 & 0.9343 & 0.9755 & 0.9588 & 0.9625 & 0.9543 & 0.9671 & 0.9342 & 0.9756 & \textbf{2.20\%} \\
				maximal\_torque & 0.8223 & 0.9646 & 0.9520 & 0.9598 & 0.9474 & 0.9816 & 0.8502 & 0.9026 & 0.9281 & 0.9343 & 0.9474 & 0.9796 & \textbf{5.33\%} \\
				fuel\_consumption\_country & 0.8804 & 0.8849 & 0.8965 & 0.8997 & 0.8684 & 0.8827 & 0.8218 & 0.8297 & 0.8948 & 0.8997 & 0.8606 & 0.8870 & \textbf{1.18\%} \\
				acceleration & 0.8769 & 0.8892 & 0.9157 & 0.9186 & 0.1021 & 0.8657 & 0.8134 & 0.8307 & 0.8753 & 0.8865 & 0.8274 & 0.8737 & \textbf{126.44\%} \\
				airfoild & 0.6277 & 0.7963 & 0.8335 & 0.8413 & 0.0465 & 0.7800 & 0.8090 & 0.8622 & 0.2456 & 0.7487 & 0.0459 & 0.7783 & \textbf{568.71\%} \\
				mortgage & 0.9921 & 0.9934 & 0.9776 & 0.9948 & 0.9982 & 0.9984 & 0.9959 & 0.9960 & 0.9864 & 0.9873 & 0.9965 & 0.9972 & \textbf{0.35\%} \\
				treasury & 0.9818 & 0.9878 & 0.9658 & 0.9854 & 0.9929 & 0.9936 & 0.9903 & 0.9928 & 0.9778 & 0.9796 & 0.9927 & 0.9935 & \textbf{0.54\%} \\
				concreteStrength & 0.8647 & 0.8751 & 0.8272 & 0.8364 & 0.6276 & 0.8112 & 0.6782 & 0.7879 & 0.8228 & 0.8316 & 0.6277 & 0.8244 & \textbf{13.36\%} \\
				\bottomrule
			\end{tabular}%
		}%
		\bigskip
		\parbox{\textwidth}{\raggedleft\itshape Average enhancement: \textbf{23.64\%}}
	\end{table*}

	
	\textbf{RQ3:} ``\textit{How does the degree and nature of target variable skewness across datasets influence the effectiveness and robustness of the proposed hybrid pipeline?}''
	
	In regression problems, a (positive or negative) skewed target distribution indicates that the majority of samples are concentrated in a limited region of the range, while a minority of samples occupy the tails. Therefore, higher absolute skewness directly reflects a stronger imbalance between the dense majority region and the sparse rare-value regions of the target space. The further the tail extends, the more severe the under-representation of extreme values becomes. Table \ref{tab:skewness_report} demonstrates the skewness of the target variables in applied benchmark datasets. Skewness computed via \emph{Pearson's second coefficient} of skewness. 
	
	To answer RQ3, we computed \textit{Spearman’s rank correlation} ($\rho$) between absolute skewness $|\gamma|$ and average percent improvement across the three evaluation metrics. Fig. ~\ref{fig:skewness_3panel} visualizes this relationship. Spearman’s rank correlation analysis reveals a significant positive relationship between absolute target skewness and performance improvement: $\rho = 0.618$ ($p = 0.011$) for MAE, $\rho = 0.579$ ($p = 0.019$) for RMSE, and $\rho = 0.546$ ($p = 0.029$) for $R^2$. Higher absolute skewness consistently predicts greater performance gains, with MAE showing the strongest and most reliable correlation ($\rho = 0.62$), indicating that greater skewness drives larger reductions in average prediction error; RMSE follows closely ($\rho = 0.58$), demonstrating robust improvement in controlling large, outlier-sensitive errors; and $R^2$ exhibits a clear upward trend ($\rho = 0.55$), where extreme imbalance (such as in airfoild with $|\gamma| \approx 2.75$ yielding 568\% improvement) enhances explained variance, collectively confirming that the hybrid pipeline’s effectiveness scales with the severity of target skewness across all metrics.

	According to analysis, the positive monotonic correlation confirms that skewness is a reliable proxy for target imbalance: the more asymmetric the distribution, the more the hybrid framework outperforms standalone regressors. This robustness stems from its adaptive binning and cost-sensitive mechanisms, which prioritize underrepresented regions. Therefore, the proposed hybrid pipeline is most effective under high target skewness, the greater the imbalance, the stronger its advantage, demonstrating superior robustness across diverse real-world regression scenarios.

	\begin{table}[htbp]
		\centering
		\caption{Skewness ($\gamma$) of target variable and performance improvement of the hybrid pipeline across 16 benchmark datasets.}
		\label{tab:skewness_report}
		\resizebox{0.6\textwidth}{!}{
			\begin{tabular}{l c c c c}
				\toprule
				\textbf{Dataset} & \textbf{Skewness $(\gamma)$} & \multicolumn{3}{c}{\textbf{Avg. \% of Enhancement}} \\
				\cmidrule(lr){3-5}
				&  & \textbf{MAE $\downarrow$} & \textbf{RMSE $\downarrow$} & \textbf{$R^2$ $\uparrow$} \\
				\midrule
				california & 0.9777 & 21.15 & 21.77 & 17.89 \\
				compactive & -3.4161 & 36.21 & 29.53 & 10.80 \\
				cpu\_small & -3.4161 & 26.43 & 29.14 & 12.39 \\
				heat & 1.6340 & 49.61 & 49.08 & 5.65 \\
				wine\_quality & 0.2 & 9.89 & 10.38 & 45.16 \\
				abalone & 1.1137 & 9.00 & 13.02 & 66.11 \\
				space\_ga & $\approx$1.0 & 10.48 & 11.56 & 16.01 \\
				debutanizer & 1.7088 & 16.55 & 20.66 & 124.99 \\
				available\_power & 1.9036 & 25.36 & 19.68 & 2.20 \\
				maximal\_torque & 1.6338 & 16.91 & 15.42 & 5.33 \\
				fuel\_consumption\_country & 1.1378 & 6.73 & 4.76 & 1.18 \\
				acceleration & 0.7694 & 25.56 & 13.60 & 126.44 \\
				airfoild & -2.7545 & 32.58 & 32.40 & 568.71 \\
				mortgage & 1.0282 & 21.67 & 16.28 & 0.35 \\
				treasury & $\approx$0.3 & 24.74 & 4.02 & 0.54 \\
				concreteStrength & $\approx$1.0 & 15.47 & 15.90 & 13.36 \\
				\bottomrule
			\end{tabular}
		}
	\end{table}

	\begin{figure*}[htbp]
		\centering
		\begin{tikzpicture}
			
			\begin{axis}[
				name=plot1,
				width=\textwidth,
				height=5.5cm,
				title={\small\textbf{(a) MAE improvement vs. skewness}},
				xlabel={Absolute Skewness $|\gamma|$},
				ylabel={Avg. \% of Improvement in MAE},
				xmin=0, xmax=3.6,
				ymin=0, ymax=60,
				xtick={0,1,2,3},
				ytick={0,10,20,30,40,50,60},
				grid=major,
				legend style={font=\scriptsize, draw=none, at={(0.01,0.99)}, anchor=north west},
				label style={font=\scriptsize},
				tick label style={font=\scriptsize},
				title style={font=\small}
				]
				\addplot[
				only marks, 
				mark=*, 
				mark size=2.5pt,
				color=blue!70!black
				] coordinates {
					(0.9777,21.15) (3.4161,36.21) (3.4161,26.43) (1.6340,49.61) (0.2000,9.89)
					(1.1137,9.00) (1.0000,10.48) (1.7088,16.55) (1.9036,25.36) (1.6338,16.91)
					(1.1378,6.73) (0.7694,25.56) (2.7545,32.58) (1.0282,21.67) (0.3000,24.74)
					(1.0000,15.47)
				};
				\addlegendentry{MAE}
				
				\addplot[
				thick, 
				dashed, 
				color=blue!70!black
				] coordinates {(0,12) (3.5,45)};
				\addlegendentry{Trend ($\rho=0.62$)}
				
			\end{axis}
			
			\begin{axis}[
				anchor=north west,
				at={(plot1.south west)},
				yshift=-2.5cm,
				name=plot2,
				width=\textwidth,
				height=5.5cm,
				title={\small\textbf{(b) RMSE improvement vs. skewness}},
				xlabel={Absolute Skewness $|\gamma|$},
				ylabel={Avg. \% of Improvement in RMSE},
				xmin=0, xmax=3.6,
				ymin=0, ymax=60,
				xtick={0,1,2,3},
				ytick={0,10,20,30,40,50,60},
				grid=major,
				legend style={font=\scriptsize, draw=none, at={(0.01,0.99)}, anchor=north west},
				label style={font=\scriptsize},
				tick label style={font=\scriptsize},
				title style={font=\small}
				]
				\addplot[
				only marks, 
				mark=*, 
				mark size=2.5pt,
				color=green!70!black
				] coordinates {
					(0.9777,21.77) (3.4161,29.53) (3.4161,29.14) (1.6340,49.08) (0.2000,10.38)
					(1.1137,13.02) (1.0000,11.56) (1.7088,20.66) (1.9036,19.68) (1.6338,15.42)
					(1.1378,4.76) (0.7694,13.60) (2.7545,32.40) (1.0282,16.28) (0.3000,4.02)
					(1.0000,15.90)
				};
				\addlegendentry{RMSE}
				
				\addplot[
				thick, 
				dashed, 
				color=green!70!black
				] coordinates {(0,10) (3.5,42)};
				\addlegendentry{Trend ($\rho=0.58$)}
				
			\end{axis}
			
			\begin{axis}[
				anchor=north west,
				at={(plot2.south west)},
				yshift=-2.5cm,
				name=plot3,
				width=\textwidth,
				height=5.5cm,
				title={\small\textbf{(c) $R^2$ improvement vs. skewness}},
				xlabel={Absolute Skewness $|\gamma|$},
				ylabel={Avg. \% of Improvement in $R^2$},
				xmin=0, xmax=3.6,
				ymin=0, ymax=600,
				xtick={0,1,2,3},
				ytick={0,100,200,300,400,500,600},
				grid=major,
				legend style={font=\scriptsize, draw=none, at={(0.01,0.99)}, anchor=north west},
				label style={font=\scriptsize},
				tick label style={font=\scriptsize},
				title style={font=\small}
				]
				\addplot[
				only marks, 
				mark=*, 
				mark size=2.5pt,
				color=orange!80!black
				] coordinates {
					(0.9777,17.89) (3.4161,10.80) (3.4161,12.39) (1.6340,5.65) (0.2000,45.16)
					(1.1137,66.11) (1.0000,16.01) (1.7088,124.99) (1.9036,2.20) (1.6338,5.33)
					(1.1378,1.18) (0.7694,126.44) (2.7545,568.71) (1.0282,0.35) (0.3000,0.54)
					(1.0000,13.36)
				};
				\addlegendentry{$R^2$}
				
				\addplot[
				thick, 
				dashed, 
				color=orange!80!black
				] coordinates {(0,20) (3.5,300)};
				\addlegendentry{Trend ($\rho=0.55$)}
				
			\end{axis}
			
		\end{tikzpicture}
		\caption{Impact of target variable skewness on hybrid pipeline performance across applied datasets. The points and trend lines show consistent performance gains with increasing absolute skewness. Outliers in $R^2$ (like airfoild, acceleration) highlight exceptional effectiveness under extreme imbalance. Spearman’s rank correlations: between $|\gamma|$ and \% improvement: MAE ($\rho = 0.618$, $p = 0.011$), RMSE ($\rho = 0.579$, $p = 0.019$), $R^2$ ($\rho = 0.546$, $p = 0.029$) all $p < 0.05$. The dashed line is a hand-drawn illustrative guide used solely to depict the overall trend.}
		\label{fig:skewness_3panel}
	\end{figure*}

	\textbf{RQ4:} ``\textit{What are the limitations of the proposed hybrid framework?}''
	
	During the evaluation of the proposed hybrid imbalanced learning pipeline on widely recognized benchmark datasets in imbalanced regression, a key limitation emerged that shapes its practical applicability: the framework exhibits a \textit{dependency on sufficient data volume} to support its five-phase architecture. The adaptive bin partitioning (Phase 0) requires reliable local $R^2$ estimation, the CVAE (Phase I) demands stable conditional latent learning, and both data-level oversampling (Phase II) and LDWL (Phase III) rely on accurate density estimation in minority regions. Our experiments show that the proposed hybrid pipeline underperforms standalone regressors in small datasets (approximately with size below 1,000 samples), whereas it consistently outperforms them in large datasets.
	Table \ref{tab:small_datasets} lists the small datasets on which the framework was evaluated. Tables ~\ref{tab:mae_smal_datasets_single_vs_pipeline}--\ref{tab:r2_small_datasets_single_vs_pipeline} demonstrates MAE, RMSE, and $R^2$ for each regressor in isolation and when integrated with the hybrid pipeline across the small datasets. As can be seen in some cases (underlined cells) the proposed hybrid pipeline fails to improve the performance compared to standalone regressors. These degradations stem from overfitting in adaptive binning and CVAE training due to sparse minority regions, unstable latent representations, and amplified noise from oversampling, issues that vanish in larger datasets where sufficient samples stabilize all phases. This pattern strongly confirms the framework’s reliance on data volume for robust operation. Overall, these patterns strongly support a conclusion that the proposed framework is most effective when enough observations exist to stabilize latent representation learning and minority density estimation. This limitation does not undermine the core contribution (a scalable, unified solution for large-scale imbalanced regression) but highlight opportunities for future work.

	\begin{table}[h!]
		\centering
		\caption{Small datasets used in the evaluation.}
		\label{tab:small_datasets}
		\resizebox{0.5\textwidth}{!}{
			\begin{tabular}{lcc}
				\hline
				\textbf{Dataset} & \textbf{No. of Samples} & \textbf{No. of Features} \\
				\hline
				boston & 506 & 13 \\
				analcatdata\_apnea3 & 450 & 11 \\
				kdd\_coil\_1 & 316 & 18 \\
				triazines & 186 & 60 \\
				cocomo\_numeric & 60 & 56 \\
				\hline
			\end{tabular}
		}
	\end{table}

	\begin{table}[htbp]
		\centering
		\caption{MAE ($\downarrow$) results for regressors without (gray columns) and with the proposed hybrid pipeline across small benchmark datasets. Underlined pipeline values indicate cases where the proposed hybrid pipeline fails to improve performance compared to the standalone regressor.}
		
		\label{tab:mae_smal_datasets_single_vs_pipeline}
		\resizebox{\textwidth}{!}{
			\begin{tabular}{l cc cc cc cc cc cc cc}
				\toprule
				\multirow{2}{*}{\textbf{Dataset}} &
				\multicolumn{2}{c}{\textbf{MLP}} &
				\multicolumn{2}{c}{\textbf{XGB}} &
				\multicolumn{2}{c}{\textbf{LR}} &
				\multicolumn{2}{c}{\textbf{KNN}} &
				\multicolumn{2}{c}{\textbf{SVR}} &
				\multicolumn{2}{c}{\textbf{Ridge}} \\
				\cmidrule(lr){2-3}
				\cmidrule(lr){4-5}
				\cmidrule(lr){6-7}
				\cmidrule(lr){8-9}
				\cmidrule(lr){10-11}
				\cmidrule(lr){12-13}
				& \cellcolor{lightgray}\textbf{Single} & \textbf{Pipeline}
				& \cellcolor{lightgray}\textbf{Single} & \textbf{Pipeline}
				& \cellcolor{lightgray}\textbf{Single} & \textbf{Pipeline}
				& \cellcolor{lightgray}\textbf{Single} & \textbf{Pipeline}
				& \cellcolor{lightgray}\textbf{Single} & \textbf{Pipeline}
				& \cellcolor{lightgray}\textbf{Single} & \textbf{Pipeline} \\
				\midrule
				boston &
				\cellcolor{lightgray}0.1396 & \underline{0.1477} &
				\cellcolor{lightgray}0.1156 & \underline{0.1375} &
				\cellcolor{lightgray}0.1557 & 0.1555 &
				\cellcolor{lightgray}0.1655 & 0.1263 &
				\cellcolor{lightgray}0.1236 & \underline{0.1461} &
				\cellcolor{lightgray}0.1543 & 0.1499 \\
				analcatdata\_apnea3 &
				\cellcolor{lightgray}0.0770 & 0.0506 &
				\cellcolor{lightgray}0.0372 & 0.0363 &
				\cellcolor{lightgray}0.1438 & 0.0662 &
				\cellcolor{lightgray}0.0393 & 0.0247 &
				\cellcolor{lightgray}0.0662 & \underline{0.1050} &
				\cellcolor{lightgray}0.1435 & 0.0468 \\
				kdd\_coil\_1 &
				\cellcolor{lightgray}0.2134 & \underline{0.2423} &
				\cellcolor{lightgray}0.1872 & \underline{0.2369} &
				\cellcolor{lightgray}0.2508 & 0.2433 &
				\cellcolor{lightgray}0.2850 & 0.2099 &
				\cellcolor{lightgray}0.2114 & \underline{0.2214} &
				\cellcolor{lightgray}0.2492 & 0.2291 \\
				triazines &
				\cellcolor{lightgray}0.2927 & \underline{0.3188} &
				\cellcolor{lightgray}0.2271 & \underline{0.2765} &
				\cellcolor{lightgray}0.2876 & \underline{0.2976} &
				\cellcolor{lightgray}0.2691 & 0.2658 &
				\cellcolor{lightgray}0.2440 & \underline{0.2557} &
				\cellcolor{lightgray}0.2795 & \underline{0.2965} \\
				cocomo\_numeric &
				\cellcolor{lightgray}0.5399 & \underline{0.5468} &
				\cellcolor{lightgray}0.3737 & \underline{0.4668} &
				\cellcolor{lightgray}0.2481 & \underline{0.4225} &
				\cellcolor{lightgray}0.2176 & \underline{0.4701} &
				\cellcolor{lightgray}0.4621 & 0.1210 &
				\cellcolor{lightgray}0.3533 & \underline{0.4440} \\
				\bottomrule
			\end{tabular}
		}
	\end{table}

	\begin{table}[htbp]
		\centering
		\caption{RMSE ($\downarrow$) results for regressors without (gray columns) and with the proposed hybrid pipeline across small benchmark datasets. Underlined pipeline values indicate cases where the proposed hybrid pipeline fails to improve performance compared to the standalone regressor.}
		\label{tab:rmse_smal_datasets_single_vs_pipeline}
		\resizebox{\textwidth}{!}{
			\begin{tabular}{l cc cc cc cc cc cc cc}
				\toprule
				\multirow{2}{*}{\textbf{Dataset}} &
				\multicolumn{2}{c}{\textbf{MLP}} &
				\multicolumn{2}{c}{\textbf{XGB}} &
				\multicolumn{2}{c}{\textbf{LR}} &
				\multicolumn{2}{c}{\textbf{KNN}} &
				\multicolumn{2}{c}{\textbf{SVR}} &
				\multicolumn{2}{c}{\textbf{Ridge}} \\
				\cmidrule(lr){2-3}
				\cmidrule(lr){4-5}
				\cmidrule(lr){6-7}
				\cmidrule(lr){8-9}
				\cmidrule(lr){10-11}
				\cmidrule(lr){12-13}
				& \cellcolor{lightgray}\textbf{Single} & \textbf{Pipeline}
				& \cellcolor{lightgray}\textbf{Single} & \textbf{Pipeline}
				& \cellcolor{lightgray}\textbf{Single} & \textbf{Pipeline}
				& \cellcolor{lightgray}\textbf{Single} & \textbf{Pipeline}
				& \cellcolor{lightgray}\textbf{Single} & \textbf{Pipeline}
				& \cellcolor{lightgray}\textbf{Single} & \textbf{Pipeline} \\
				\midrule
				boston &
				\cellcolor{lightgray}0.2152 & \underline{0.2418} &
				\cellcolor{lightgray}0.1725 & \underline{0.2203} &
				\cellcolor{lightgray}0.2297 & 0.2161 &
				\cellcolor{lightgray}0.2687 & 0.2082 &
				\cellcolor{lightgray}0.2047 & \underline{0.2173} &
				\cellcolor{lightgray}0.2292 & 0.2097 \\
				analcatdata\_apnea3 &
				\cellcolor{lightgray}0.1075 & 0.0758 &
				\cellcolor{lightgray}0.1046 & 0.0888 &
				\cellcolor{lightgray}0.2558 & 0.1001 &
				\cellcolor{lightgray}0.0309 & 0.0247 &
				\cellcolor{lightgray}0.0662 & \underline{0.1050} &
				\cellcolor{lightgray}0.1435 & 0.0468 \\
				kdd\_coil\_1 &
				\cellcolor{lightgray}0.3194 & \underline{0.3509} &
				\cellcolor{lightgray}0.2595 & \underline{0.3072} &
				\cellcolor{lightgray}0.3149 & 0.3144 &
				\cellcolor{lightgray}0.3683 & 0.2773 &
				\cellcolor{lightgray}0.2949 & \underline{0.2691} &
				\cellcolor{lightgray}0.3120 & 0.3049 \\
				triazines &
				\cellcolor{lightgray}0.4000 & \underline{0.4078} &
				\cellcolor{lightgray}0.3217 & \underline{0.3682} &
				\cellcolor{lightgray}0.4045 & 0.3851 &
				\cellcolor{lightgray}0.3545 & \underline{0.3718} &
				\cellcolor{lightgray}0.3307 & \underline{0.3718} &
				\cellcolor{lightgray}0.3760 & \underline{0.3802} \\
				cocomo\_numeric &
				\cellcolor{lightgray}0.6565 & \underline{0.7334} &
				\cellcolor{lightgray}0.6418 & \underline{0.7090} &
				\cellcolor{lightgray}0.3622 & \underline{0.7449} &
				\cellcolor{lightgray}0.3369 & \underline{0.7657} &
				\cellcolor{lightgray}0.7589 & 0.7051 &
				\cellcolor{lightgray}0.4913 & \underline{0.7424} \\
				\bottomrule
			\end{tabular}
		}
	\end{table}

	\begin{table}[htbp]
		\centering
		\caption{$R^2$ ($\uparrow$) results for regressors without (gray columns) and with the proposed hybrid pipeline across small benchmark datasets. Underlined pipeline values indicate cases where the proposed hybrid pipeline fails to improve performance compared to the standalone regressor.}
		\label{tab:r2_small_datasets_single_vs_pipeline}
		\resizebox{\textwidth}{!}{
			\begin{tabular}{l cc cc cc cc cc cc cc}
				\toprule
				\multirow{2}{*}{\textbf{Dataset}} &
				\multicolumn{2}{c}{\textbf{MLP}} &
				\multicolumn{2}{c}{\textbf{XGB}} &
				\multicolumn{2}{c}{\textbf{LR}} &
				\multicolumn{2}{c}{\textbf{KNN}} &
				\multicolumn{2}{c}{\textbf{SVR}} &
				\multicolumn{2}{c}{\textbf{Ridge}} \\
				\cmidrule(lr){2-3}
				\cmidrule(lr){4-5}
				\cmidrule(lr){6-7}
				\cmidrule(lr){8-9}
				\cmidrule(lr){10-11}
				\cmidrule(lr){12-13}
				& \cellcolor{lightgray}\textbf{Single} & \textbf{Pipeline}
				& \cellcolor{lightgray}\textbf{Single} & \textbf{Pipeline}
				& \cellcolor{lightgray}\textbf{Single} & \textbf{Pipeline}
				& \cellcolor{lightgray}\textbf{Single} & \textbf{Pipeline}
				& \cellcolor{lightgray}\textbf{Single} & \textbf{Pipeline}
				& \cellcolor{lightgray}\textbf{Single} & \textbf{Pipeline} \\
				\midrule
				boston &
				\cellcolor{lightgray}0.6804 & 0.7011 &
				\cellcolor{lightgray}0.7945 & \underline{0.6787} &
				\cellcolor{lightgray}0.6357 & 0.7196 &
				\cellcolor{lightgray}0.5016 & 0.6918 &
				\cellcolor{lightgray}0.7107 & \underline{0.6790} &
				\cellcolor{lightgray}0.6374 & 0.6896 \\
				analcatdata\_apnea3 &
				\cellcolor{lightgray}0.8905 & 0.9456 &
				\cellcolor{lightgray}0.8964 & 0.9254 &
				\cellcolor{lightgray}0.3800 & 0.9051 &
				\cellcolor{lightgray}0.9258 & 0.9561 &
				\cellcolor{lightgray}0.8874 & \underline{0.8665} &
				\cellcolor{lightgray}0.3794 & 0.9367 \\
				kdd\_coil\_1 &
				\cellcolor{lightgray}0.0500 & 0.1468 &
				\cellcolor{lightgray}0.3728 & \underline{0.1209} &
				\cellcolor{lightgray}0.0763 & 0.0792 &
				\cellcolor{lightgray}0.2631 & 0.2841 &
				\cellcolor{lightgray}0.1902 & \underline{0.3254} &
				\cellcolor{lightgray}0.0935 & 0.1343 \\
				triazines &
				\cellcolor{lightgray}0.0347 & \underline{0.0032} &
				\cellcolor{lightgray}0.3754 & \underline{0.1824} &
				\cellcolor{lightgray}0.0130 & 0.1053 &
				\cellcolor{lightgray}0.2418 & \underline{0.1658} &
				\cellcolor{lightgray}0.3402 & \underline{0.3338} &
				\cellcolor{lightgray}0.1472 & \underline{0.1278} \\
				cocomo\_numeric &
				\cellcolor{lightgray}0.4003 & \underline{0.2515} &
				\cellcolor{lightgray}0.7511 & \underline{0.3004} &
				\cellcolor{lightgray}0.8175 & \underline{0.2278} &
				\cellcolor{lightgray}0.6480 & \underline{0.1841} &
				\cellcolor{lightgray}0.1986 & 0.3081 &
				\cellcolor{lightgray}0.6642 & \underline{0.2330} \\
				\bottomrule
			\end{tabular}
		}
	\end{table}

	\subsection{Ablation Study}
	\label{sec:ablation}
	\textbf{RQ5:} ``\textit{How does each phase of the proposed framework (adaptive bin partitioning, representation learning, data-level balancing, algorithm-level balancing, and fusion) contribute to overall performance in large-scale imbalanced regression?}''
	
	To investigate this question, we perform a phase-wise ablation study to isolate and quantify the importance of each component within the proposed hybrid framework. Specifically, for each ablation, one phase is \emph{substituted} with a standard baseline (quantile binning, SMOGN, MSE, or concatenation) or \emph{removed entirely} (representation learning, Phase I) when no meaningful alternative exists, while all other phases remain unchanged. In particular: (i) the adaptive bin partitioning is replaced with a standard quantile-based ($q=10$) binning strategy; (ii) the CVAE-based latent representation learning is removed entirely; (iii) the proposed data-level oversampling is replaced with SMOGN, a commonly used regression-oriented oversampling technique; (iv) the LDWL cost-sensitive loss is substituted with a standard regression loss (Mean Squared Error-MSE); and (v) the proposed feature fusion is replaced with a simple feature concatenation scheme. The resulting variants are then evaluated across all benchmark datasets and regressors. 
	
	For each ablation condition, we report the average \emph{relative performance change} (in percent) with respect to the full proposed framework, computed across three evaluation metrics (MAE, RMSE, and $R^2$) over all benchmark datasets and regressors. This relative change is defined as:
	\begin{equation}
		\text{Relative performance change (\%)} = \frac{\text{Metric}_{\text{ablation}} - \text{Metric}_{\text{full}}}{\left| \text{Metric}_{\text{full}} \right|} \times 100,
	\end{equation}

	Table \ref{tab:ablation-main} summarizes the average relative performance change (\%) observed when each phase of the framework is removed or replaced with its baseline counterpart. A positive $\Delta$ ($\Delta$ = ablated value $-$ full-framework value) indicates degraded performance for MAE and RMSE, while a negative $\Delta$ indicates degradation for $R^2$. These values reflect the contribution of each component and are averaged across all benchmark datasets and regressors. Detailed per-dataset and per-metric ablation results are provided in \ref{appen:ablation-study}.

	\begin{table}[htbp]
		\centering
		\caption{Ablation Study. Average relative change (\%) in performance when each phase is removed, compared to the full framework. Positive $\Delta$ indicates degradation for MAE/RMSE, negative for R$^2$. Full details in ~\ref{appen:ablation-study}.}
		\label{tab:ablation-main}
		\resizebox{\textwidth}{!}{
			
			\begin{tabular}{lccc}
				\toprule
				\textbf{Phase Removed} & \textbf{Avg. $\Delta$ MAE (\%)} & \textbf{Avg. $\Delta$ RMSE (\%)} & \textbf{Avg. $\Delta$ R$^2$ (\%)} \\
				\midrule
				Without Adaptive Bin Partitioning (Phase 0) & 130.61 & 104.66 & -4.92 \\
				Without Representation Learning (Phase 1)   & 114.57 & 100.76 & -8.02 \\
				Without Data-Level Method (Phase 2)        & 37.33  & 10.33  & -2.96 \\
				Without Algorithm-Level Method (Phase 3)   & 16.03  & 19.83  & -3.11 \\
				Without Feature Fusion (Phase 4)            & 24.24 & 18.76 & -6.62 \\
				
				\bottomrule
			\end{tabular}
		}
	\end{table}

	\section{Conclusion}
	\label{sec:conclusion}
	This study addresses the persistent challenge of imbalanced regression, where non-uniform target distributions lead to biased models that underperform on rare yet critical values. We proposed a unified hybrid framework that synergistically integrates data-level and algorithm-level strategies through a five-phase pipeline: adaptive target-space bin partitioning to discretize continuous targets; target-conditioned representation learning via CVAE for enhanced latent modeling; multistage data-level balancing with feature clustering and minority oversampling; algorithm-level balancing using a novel Latent-Density Weighted Loss (LDWL); and attention-based gated fusion for robust feature integration.
	
	To demonstrate that the hybrid integration of data-level and algorithm-level balancing outperforms isolated strategies, we evaluated the proposed framework upstream of several benchmark regressors with and without the pipeline across well-known imbalanced regression datasets. The results reveal consistent improvements in MAE, RMSE, and $R^2$ when our framework is applied, confirming the superiority of the combined approach. Ablation studies further validate each phase’s essential contribution, justifying the effectiveness of the proposed phases in imbalanced learning. These findings conclusively affirm that the hybrid paradigm significantly outperforms standalone data- or algorithm-level methods. Future work should focus on adaptive parameter tuning across framework components to optimize performance for dataset-specific characteristics.
	
	\appendix
	
	\section{Target Variable Distributions Across Datasets}
	\label{appen:datasets-target-lable-distrib}
\begin{figure}[ht]
		\centering
		\includegraphics[width=1.0\textwidth]{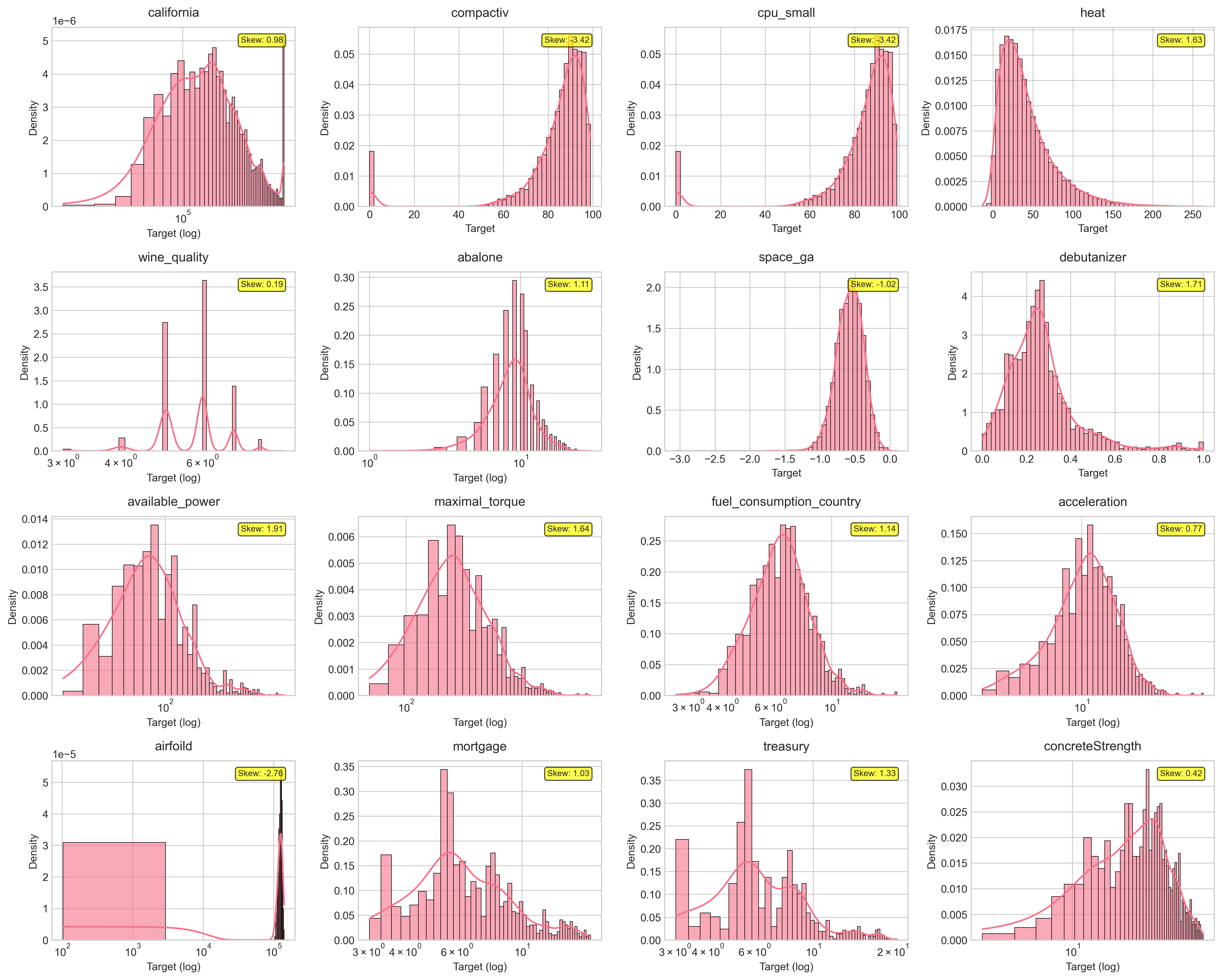} 
		\caption{Distributions of the target variable across all applied datasets. Histogram bars represent empirical density, while the overlaid smooth curve is a Kernel Density Estimate (KDE) illustrating the underlying probability distribution and highlighting skewness or multimodality.  In datasets with severe positive skew or scale variation, a logarithmic x-axis is applied to better reveal tail behavior and support visual assessment of regression imbalance.}
		\label{fig:target_distributions_appendix}
	\end{figure}

	Fig.~\ref{fig:target_distributions_appendix} presents the distribution of the target variable for each dataset, clearly revealing pronounced skewness and heavy tails, hallmarks of imbalance in regression tasks.

	\section{Detailed Ablation Study Results}
	\label{appen:ablation-study}
	Tables \ref{tab:detailed_Ablation_Ph0} to \ref{tab:detailed_Ablation_Ph4} present a detailed ablation study for each phase of the proposed hybrid imbalanced learning framework across three evaluation metrics: MAE, RMSE, and $R^2$. For every dataset and regressor, removing any phase consistently degrades performance. Specifically, MAE and RMSE increase, while $R^2$ decreases, confirming the contribution of each component. Although the magnitude of improvement varies across datasets and models, the full framework uniformly outperforms all ablated configurations, underscoring the effectiveness and complementarity of the proposed phases.

	\begin{table}[p]
		\centering
		\caption{Ablation study \emph{without Phase~0 (Adaptive Bin Partitioning)}: for each dataset and regressor we report \emph{Full framework} vs.\ \emph{w/o Phase~0}. “Avg.\ Percent Improvement” is the average over regressors of $((\text{w/o}-\text{Full})/\text{Full})\times100$ (positive means worse without the phase for MAE and RMSE; negative means worse without the phase for $R^2$).}
		\label{tab:detailed_Ablation_Ph0}
		\small
		\setlength{\tabcolsep}{3pt}
		\renewcommand{\arraystretch}{0.92}
		\resizebox{\textwidth}{!}{
			\begin{tabular}{
					l
					l|
					>{\columncolor{lightgray}}c c|
					>{\columncolor{lightgray}}c c|
					>{\columncolor{lightgray}}c c|
					>{\columncolor{lightgray}}c c|
					>{\columncolor{lightgray}}c c|
					>{\columncolor{lightgray}}c c|c}
				\toprule
				& \textbf{Dataset}
				& \multicolumn{2}{c|}{\textbf{MLP}}
				& \multicolumn{2}{c|}{\textbf{XGB}}
				& \multicolumn{2}{c|}{\textbf{LR}}
				& \multicolumn{2}{c|}{\textbf{KNN}}
				& \multicolumn{2}{c|}{\textbf{SVR}}
				& \multicolumn{2}{c|}{\textbf{Ridge}}
				& \\
				\cmidrule(lr){3-4}\cmidrule(lr){5-6}\cmidrule(lr){7-8}\cmidrule(lr){9-10}\cmidrule(lr){11-12}\cmidrule(lr){13-14}
				& &
				\textbf{Full} & \textbf{w/o Ph.0} &
				\textbf{Full} & \textbf{w/o Ph.0} &
				\textbf{Full} & \textbf{w/o Ph.0} &
				\textbf{Full} & \textbf{w/o Ph.0} &
				\textbf{Full} & \textbf{w/o Ph.0} &
				\textbf{Full} & \textbf{w/o Ph.0} &
				\textbf{Avg.\ \% Imp.} \\
				\midrule
				\multirow{16}{*}{\rotatebox{90}{\textbf{MAE}}}
				& california                 & 0.1429 & 0.1463 & 0.1446 & 0.1497 & 0.1410 & 0.1524 & 0.1473 & 0.1498 & 0.1397 & 0.1512 & 0.1407 & 0.1467 & 4.70 \\
				& compactive                 & 0.0311 & 0.0405 & 0.0415 & 0.0499 & 0.0393 & 0.0472 & 0.0433 & 0.0502 & 0.0479 & 0.0611 & 0.0393 & 0.0470 & 22.28 \\
				& cpu\_small                 & 0.0462 & 0.0481 & 0.0348 & 0.0499 & 0.0463 & 0.0472 & 0.0422 & 0.0502 & 0.0519 & 0.0611 & 0.0446 & 0.0470 & 15.25 \\
				& heat                       & 0.0169 & 0.0179 & 0.0118 & 0.1602 & 0.0040 & 0.1635 & 0.0117 & 0.1668 & 0.0395 & 0.1651 & 0.0044 & 0.1626 & 1748.34 \\
				& wine\_quality              & 0.1595 & 0.1665 & 0.1604 & 0.1645 & 0.1648 & 0.1665 & 0.1640 & 0.1668 & 0.1644 & 0.1652 & 0.1640 & 0.1658 & 1.88 \\
				& abalone                    & 0.1227 & 0.1270 & 0.1247 & 0.1252 & 0.1257 & 0.1559 & 0.1342 & 0.1579 & 0.1217 & 0.1274 & 0.1249 & 0.1322 & 9.35 \\
				& space\_ga                  & 0.0531 & 0.0556 & 0.0567 & 0.1279 & 0.0568 & 0.1280 & 0.0543 & 0.1040 & 0.0509 & 0.1252 & 0.0556 & 0.1254 & 103.11 \\
				& debutanizer                & 0.1155 & 0.1166 & 0.1219 & 0.1279 & 0.1309 & 0.1379 & 0.0785 & 0.1041 & 0.1223 & 0.1253 & 0.1277 & 0.1284 & 7.81 \\
				& available\_power           & 0.0336 & 0.0364 & 0.0349 & 0.0356 & 0.0305 & 0.0314 & 0.0197 & 0.0350 & 0.0423 & 0.0608 & 0.0301 & 0.0317 & 23.33 \\
				& maximal\_torque            & 0.0111 & 0.0347 & 0.0304 & 0.0330 & 0.0395 & 0.0401 & 0.0500 & 0.0524 & 0.0487 & 0.0800 & 0.0229 & 0.0245 & 49.79 \\
				& fuel\_consumption\_country & 0.0581 & 0.0679 & 0.0586 & 0.0643 & 0.0653 & 0.0665 & 0.0709 & 0.0737 & 0.0568 & 0.0827 & 0.0644 & 0.0652 & 13.20 \\
				& acceleration               & 0.0589 & 0.0683 & 0.0452 & 0.0693 & 0.0678 & 0.0683 & 0.0682 & 0.0745 & 0.0563 & 0.0770 & 0.0658 & 0.0663 & 19.46 \\
				& airfoild                   & 0.0801 & 0.0855 & 0.1095 & 0.1169 & 0.1396 & 0.1430 & 0.0715 & 0.0792 & 0.1305 & 0.1348 & 0.1362 & 0.1374 & 5.15 \\
				& mortgage                   & 0.0254 & 0.0259 & 0.0215 & 0.0261 & 0.0110 & 0.0125 & 0.0152 & 0.0230 & 0.0420 & 0.0631 & 0.0115 & 0.0133 & 25.70 \\
				& treasury                   & 0.0290 & 0.0293 & 0.0241 & 0.0255 & 0.0112 & 0.0202 & 0.0119 & 0.0193 & 0.0490 & 0.0688 & 0.0190 & 0.0201 & 32.60 \\
				& concreteStrength           & 0.1103 & 0.1337 & 0.1257 & 0.1335 & 0.1384 & 0.1395 & 0.1417 & 0.1455 & 0.1215 & 0.1371 & 0.1336 & 0.1379 & 7.83 \\
				& \multicolumn{14}{r}{\textbf{Avg.\ 130.61}} \\
				\midrule\midrule
				& \textbf{Dataset}
				& \multicolumn{2}{c|}{\textbf{MLP}}
				& \multicolumn{2}{c|}{\textbf{XGB}}
				& \multicolumn{2}{c|}{\textbf{LR}}
				& \multicolumn{2}{c|}{\textbf{KNN}}
				& \multicolumn{2}{c|}{\textbf{SVR}}
				& \multicolumn{2}{c|}{\textbf{Ridge}}
				& \\
				\cmidrule(lr){3-4}\cmidrule(lr){5-6}\cmidrule(lr){7-8}\cmidrule(lr){9-10}\cmidrule(lr){11-12}\cmidrule(lr){13-14}
				& &
				\textbf{Full} & \textbf{w/o Ph.0} &
				\textbf{Full} & \textbf{w/o Ph.0} &
				\textbf{Full} & \textbf{w/o Ph.0} &
				\textbf{Full} & \textbf{w/o Ph.0} &
				\textbf{Full} & \textbf{w/o Ph.0} &
				\textbf{Full} & \textbf{w/o Ph.0} &
				\textbf{Avg.\ \% Imp.} \\
				\midrule
				\multirow{16}{*}{\rotatebox{90}{\textbf{RMSE}}}
				& california                 & 0.2043 & 0.2162 & 0.2052 & 0.2092 & 0.1978 & 0.2075 & 0.2118 & 0.2164 & 0.1974 & 0.2109 & 0.1976 & 0.2009 & 3.89 \\
				& compactive                 & 0.0490 & 0.0647 & 0.0717 & 0.0771 & 0.0632 & 0.0668 & 0.0744 & 0.0762 & 0.0541 & 0.0852 & 0.0629 & 0.0665 & 18.48 \\
				& cpu\_small                 & 0.0659 & 0.0687 & 0.0477 & 0.0764 & 0.0645 & 0.0668 & 0.0594 & 0.0722 & 0.0606 & 0.0852 & 0.0629 & 0.0665 & 22.64 \\
				& heat                       & 0.0268 & 0.0312 & 0.0176 & 0.2036 & 0.0064 & 0.2070 & 0.0172 & 0.2186 & 0.0512 & 0.2091 & 0.0074 & 0.2060 & 1395.12 \\
				& wine\_quality              & 0.2148 & 0.2172 & 0.2058 & 0.2089 & 0.2093 & 0.2070 & 0.2164 & 0.2187 & 0.2085 & 0.2099 & 0.2084 & 0.2082 & 0.53 \\
				& abalone                    & 0.1652 & 0.1703 & 0.1668 & 0.1734 & 0.1646 & 0.1663 & 0.1856 & 0.1861 & 0.1676 & 0.1691 & 0.1636 & 0.1697 & 2.16 \\
				& space\_ga                  & 0.0729 & 0.0730 & 0.0745 & 0.1797 & 0.0733 & 0.1695 & 0.0773 & 0.1520 & 0.0711 & 0.1707 & 0.0718 & 0.1660 & 106.75 \\
				& debutanizer                & 0.1618 & 0.1650 & 0.1751 & 0.1797 & 0.1753 & 0.1795 & 0.1358 & 0.1522 & 0.1697 & 0.1707 & 0.1719 & 0.1760 & 3.68 \\
				& available\_power           & 0.0512 & 0.0515 & 0.0565 & 0.0571 & 0.0476 & 0.0484 & 0.0578 & 0.0612 & 0.0595 & 0.0798 & 0.0466 & 0.0470 & 7.36 \\
				& maximal\_torque            & 0.0522 & 0.0622 & 0.0562 & 0.0631 & 0.0582 & 0.0612 & 0.0867 & 0.0943 & 0.0695 & 0.1018 & 0.0397 & 0.0452 & 17.61 \\
				& fuel\_consumption\_country & 0.0859 & 0.0938 & 0.0779 & 0.0945 & 0.0871 & 0.0881 & 0.1055 & 0.1163 & 0.0792 & 0.1155 & 0.0855 & 0.0870 & 14.91 \\
				& acceleration               & 0.0874 & 0.0891 & 0.0582 & 0.0898 & 0.0880 & 0.0897 & 0.0988 & 0.0991 & 0.0742 & 0.0948 & 0.0854 & 0.0859 & 14.47 \\
				& airfoild                   & 0.2205 & 0.2230 & 0.1872 & 0.2173 & 0.2292 & 0.2292 & 0.1814 & 0.1867 & 0.2449 & 0.2457 & 0.2301 & 0.2322 & 3.56 \\
				& mortgage                   & 0.0340 & 0.0352 & 0.0319 & 0.0615 & 0.0165 & 0.0192 & 0.0268 & 0.0376 & 0.0497 & 0.0727 & 0.0236 & 0.0297 & 37.52 \\
				& treasury                   & 0.0458 & 0.0460 & 0.0501 & 0.0516 & 0.0501 & 0.0520 & 0.0393 & 0.0400 & 0.0587 & 0.0833 & 0.0324 & 0.0361 & 10.39 \\
				& concreteStrength           & 0.1329 & 0.1792 & 0.1203 & 0.1715 & 0.1738 & 0.1747 & 0.1842 & 0.1866 & 0.1549 & 0.1755 & 0.2379 & 0.2381 & 15.43 \\
				& \multicolumn{14}{r}{\textbf{Avg.\ 104.66}} \\
				\midrule\midrule
				& \textbf{Dataset}
				& \multicolumn{2}{c|}{\textbf{MLP}}
				& \multicolumn{2}{c|}{\textbf{XGB}}
				& \multicolumn{2}{c|}{\textbf{LR}}
				& \multicolumn{2}{c|}{\textbf{KNN}}
				& \multicolumn{2}{c|}{\textbf{SVR}}
				& \multicolumn{2}{c|}{\textbf{Ridge}}
				& \\
				\cmidrule(lr){3-4}\cmidrule(lr){5-6}\cmidrule(lr){7-8}\cmidrule(lr){9-10}\cmidrule(lr){11-12}\cmidrule(lr){13-14}
				& &
				\textbf{Full} & \textbf{w/o Ph.0} &
				\textbf{Full} & \textbf{w/o Ph.0} &
				\textbf{Full} & \textbf{w/o Ph.0} &
				\textbf{Full} & \textbf{w/o Ph.0} &
				\textbf{Full} & \textbf{w/o Ph.0} &
				\textbf{Full} & \textbf{w/o Ph.0} &
				\textbf{Avg.\ \% Imp.} \\
				\midrule
				\multirow{16}{*}{\rotatebox{90}{\textbf{$R^{2}$}}}
				& california                 & 0.8168 & 0.7843 & 0.8110 & 0.8011 & 0.8298 & 0.8104 & 0.8080 & 0.7924 & 0.8312 & 0.8059 & 0.8307 & 0.8236 & -2.23 \\
				& compactive                 & 0.9847 & 0.9706 & 0.9657 & 0.9526 & 0.9719 & 0.9635 & 0.9612 & 0.9544 & 0.9759 & 0.9407 & 0.9723 & 0.9638 & -1.47 \\
				& cpu\_small                 & 0.9709 & 0.9624 & 0.9797 & 0.9526 & 0.9658 & 0.9615 & 0.9714 & 0.9574 & 0.9675 & 0.9407 & 0.9680 & 0.9608 & -1.51 \\
				& heat                       & 0.9892 & 0.9802 & 0.9952 & 0.4948 & 0.9963 & 0.4777 & 0.9963 & 0.4172 & 0.9519 & 0.4672 & 0.9992 & 0.4826 & -53.00 \\
				& wine\_quality              & 0.4375 & 0.4460 & 0.4837 & 0.4949 & 0.4659 & 0.4777 & 0.4294 & 0.4172 & 0.4705 & 0.4672 & 0.4707 & 0.4826 & 2.41 \\
				& abalone                    & 0.5060 & 0.4751 & 0.4962 & 0.4897 & 0.5093 & 0.4921 & 0.3765 & 0.3723 & 0.4913 & 0.4867 & 0.5152 & 0.5102 & -2.30 \\
				& space\_ga                  & 0.7123 & 0.7078 & 0.6951 & 0.6328 & 0.7052 & 0.6731 & 0.6715 & 0.6614 & 0.7164 & 0.6686 & 0.7168 & 0.6865 & -4.42 \\
				& debutanizer                & 0.7024 & 0.6954 & 0.6514 & 0.6328 & 0.6505 & 0.6431 & 0.7689 & 0.7372 & 0.6727 & 0.6686 & 0.6643 & 0.6565 & -1.82 \\
				& available\_power           & 0.9706 & 0.9701 & 0.9642 & 0.9606 & 0.9755 & 0.9747 & 0.9625 & 0.9579 & 0.9671 & 0.9284 & 0.9756 & 0.9743 & -0.85 \\
				& maximal\_torque            & 0.9646 & 0.9497 & 0.9598 & 0.9483 & 0.9816 & 0.9748 & 0.9026 & 0.8844 & 0.9343 & 0.8655 & 0.9796 & 0.9735 & -2.24 \\
				& fuel\_consumption\_country & 0.8849 & 0.8639 & 0.8997 & 0.8620 & 0.8827 & 0.8800 & 0.8297 & 0.7911 & 0.8997 & 0.7937 & 0.8870 & 0.8829 & -3.96 \\
				& acceleration               & 0.8892 & 0.8623 & 0.9186 & 0.8603 & 0.8657 & 0.8648 & 0.8307 & 0.8277 & 0.8865 & 0.8442 & 0.8737 & 0.8740 & -2.43 \\
				& airfoild                   & 0.7963 & 0.7872 & 0.8413 & 0.8022 & 0.7800 & 0.7799 & 0.8622 & 0.8537 & 0.7487 & 0.7432 & 0.7783 & 0.7778 & -1.26 \\
				& mortgage                   & 0.9934 & 0.9913 & 0.9948 & 0.9807 & 0.9984 & 0.9981 & 0.9960 & 0.9927 & 0.9873 & 0.9731 & 0.9972 & 0.9970 & -0.57 \\
				& treasury                   & 0.9878 & 0.9872 & 0.9854 & 0.9842 & 0.9936 & 0.9920 & 0.9928 & 0.9907 & 0.9796 & 0.9596 & 0.9935 & 0.9924 & -0.45 \\
				& concreteStrength           & 0.8751 & 0.7992 & 0.8364 & 0.8160 & 0.8112 & 0.8077 & 0.7879 & 0.7824 & 0.8316 & 0.8075 & 0.8244 & 0.8233 & -2.55 \\
				& \multicolumn{14}{r}{\textbf{Avg.\ -4.92}} \\
				\bottomrule
			\end{tabular}
		}
	\end{table}

	\begin{table}[p]
		\centering
		\caption{Ablation study \emph{without Phase~1 (Representation Learning)}: for each dataset and regressor we report \emph{Full framework} vs.\ \emph{w/o Phase~1}. “Avg.\ Percent Improvement” is the average over regressors of $((\text{w/o}-\text{Full})/\text{Full})\times100$ (positive means worse without the phase for MAE and RMSE; negative means worse without the phase for $R^2$).}
		\label{tab:detailed_Ablation_Ph1}
		\small
		\setlength{\tabcolsep}{3pt}
		\renewcommand{\arraystretch}{0.92}
		\resizebox{\textwidth}{!}{
			\begin{tabular}{
					l
					l|
					>{\columncolor{lightgray}}c c|
					>{\columncolor{lightgray}}c c|
					>{\columncolor{lightgray}}c c|
					>{\columncolor{lightgray}}c c|
					>{\columncolor{lightgray}}c c|
					>{\columncolor{lightgray}}c c|c}
				\toprule
				& \textbf{Dataset}
				& \multicolumn{2}{c|}{\textbf{MLP}}
				& \multicolumn{2}{c|}{\textbf{XGB}}
				& \multicolumn{2}{c|}{\textbf{LR}}
				& \multicolumn{2}{c|}{\textbf{KNN}}
				& \multicolumn{2}{c|}{\textbf{SVR}}
				& \multicolumn{2}{c|}{\textbf{Ridge}}
				& \\
				\cmidrule(lr){3-4}\cmidrule(lr){5-6}\cmidrule(lr){7-8}\cmidrule(lr){9-10}\cmidrule(lr){11-12}\cmidrule(lr){13-14}
				& &
				\textbf{Full} & \textbf{w/o Ph.1} &
				\textbf{Full} & \textbf{w/o Ph.1} &
				\textbf{Full} & \textbf{w/o Ph.1} &
				\textbf{Full} & \textbf{w/o Ph.1} &
				\textbf{Full} & \textbf{w/o Ph.1} &
				\textbf{Full} & \textbf{w/o Ph.1} &
				\textbf{Avg.\ \% Imp.} \\
				\midrule
				\multirow{16}{*}{\rotatebox{90}{\textbf{MAE}}}
				& california & 0.1429 & 0.1388 & 0.1446 & 0.1453 & 0.1410 & 0.1425 & 0.1473 & 0.1493 & 0.1397 & 0.1401 & 0.1407 & 0.1429 & 0.31 \\
				& compactive & 0.0311 & 0.0370 & 0.0415 & 0.0435 & 0.0393 & 0.0395 & 0.0433 & 0.0439 & 0.0479 & 0.0573 & 0.0393 & 0.0399 & 7.81 \\
				& cpu\_small & 0.0462 & 0.0468 & 0.0348 & 0.0432 & 0.0463 & 0.0471 & 0.0422 & 0.0478 & 0.0519 & 0.0582 & 0.0446 & 0.0449 & 8.87 \\
				& heat & 0.0169 & 0.0171 & 0.0118 & 0.1343 & 0.0040 & 0.1371 & 0.0117 & 0.1426 & 0.0395 & 0.1445 & 0.0044 & 0.1359 & 1456.68 \\
				& wine\_quality & 0.1595 & 0.1650 & 0.1604 & 0.1676 & 0.1648 & 0.1769 & 0.1640 & 0.1787 & 0.1644 & 0.1743 & 0.1640 & 0.1760 & 6.26 \\
				& abalone & 0.1227 & 0.1381 & 0.1247 & 0.1343 & 0.1257 & 0.1371 & 0.1342 & 0.1426 & 0.1217 & 0.1445 & 0.1249 & 0.1359 & 10.52 \\
				& space\_ga & 0.0531 & 0.0545 & 0.0567 & 0.0962 & 0.0568 & 0.1073 & 0.0543 & 0.0939 & 0.0509 & 0.1034 & 0.0556 & 0.1061 & 71.35 \\
				& debutanizer & 0.1155 & 0.1198 & 0.1219 & 0.1462 & 0.1309 & 0.1372 & 0.0785 & 0.0939 & 0.1223 & 0.1243 & 0.1277 & 0.1360 & 9.37 \\
				& available\_power & 0.0336 & 0.0345 & 0.0349 & 0.0354 & 0.0305 & 0.0332 & 0.0197 & 0.0274 & 0.0423 & 0.0522 & 0.0301 & 0.0321 & 13.68 \\
				& maximal\_torque & 0.0111 & 0.0290 & 0.0304 & 0.0329 & 0.0395 & 0.0415 & 0.0500 & 0.0509 & 0.0487 & 0.0675 & 0.0229 & 0.0255 & 37.72 \\
				& fuel\_consumption\_country & 0.0581 & 0.0601 & 0.0586 & 0.0595 & 0.0653 & 0.0672 & 0.0709 & 0.0748 & 0.0568 & 0.0658 & 0.0644 & 0.0667 & 5.47 \\
				& acceleration & 0.0589 & 0.0642 & 0.0452 & 0.0519 & 0.0678 & 0.0690 & 0.0682 & 0.0705 & 0.0563 & 0.0688 & 0.0658 & 0.0662 & 8.63 \\
				& airfoild & 0.0801 & 0.0890 & 0.1095 & 0.1121 & 0.1396 & 0.1403 & 0.0715 & 0.0720 & 0.1305 & 0.1317 & 0.1362 & 0.1384 & 2.87 \\
				& mortgage & 0.0254 & 0.0258 & 0.0215 & 0.0221 & 0.0110 & 0.0116 & 0.0152 & 0.0170 & 0.0420 & 0.0602 & 0.0115 & 0.0116 & 10.98 \\
				& treasury & 0.0290 & 0.0304 & 0.0241 & 0.2590 & 0.0112 & 0.0149 & 0.0119 & 0.0169 & 0.0490 & 0.0636 & 0.0190 & 0.0192 & 180.90 \\
				& concreteStrength & 0.1103 & 0.1145 & 0.1257 & 0.1309 & 0.1384 & 0.1395 & 0.1417 & 0.1419 & 0.1215 & 0.1219 & 0.1336 & 0.1347 & 1.67 \\
				& \multicolumn{14}{r}{\textbf{Avg.\ 114.57}} \\
				\midrule\midrule
				& \textbf{Dataset}
				& \multicolumn{2}{c|}{\textbf{MLP}}
				& \multicolumn{2}{c|}{\textbf{XGB}}
				& \multicolumn{2}{c|}{\textbf{LR}}
				& \multicolumn{2}{c|}{\textbf{KNN}}
				& \multicolumn{2}{c|}{\textbf{SVR}}
				& \multicolumn{2}{c|}{\textbf{Ridge}}
				& \\
				\cmidrule(lr){3-4}\cmidrule(lr){5-6}\cmidrule(lr){7-8}\cmidrule(lr){9-10}\cmidrule(lr){11-12}\cmidrule(lr){13-14}
				& &
				\textbf{Full} & \textbf{w/o Ph.1} &
				\textbf{Full} & \textbf{w/o Ph.1} &
				\textbf{Full} & \textbf{w/o Ph.1} &
				\textbf{Full} & \textbf{w/o Ph.1} &
				\textbf{Full} & \textbf{w/o Ph.1} &
				\textbf{Full} & \textbf{w/o Ph.1} &
				\textbf{Avg.\ \% Imp.} \\
				\midrule
				\multirow{16}{*}{\rotatebox{90}{\textbf{RMSE}}}
				& california & 0.2043 & 0.2198 & 0.2052 & 0.2120 & 0.1978 & 0.2142 & 0.2118 & 0.2273 & 0.1974 & 0.2160 & 0.1976 & 0.1995 & 6.15 \\
				& compactive & 0.0490 & 0.0609 & 0.0717 & 0.0698 & 0.0632 & 0.0645 & 0.0744 & 0.0740 & 0.0541 & 0.0859 & 0.0629 & 0.0636 & 13.84 \\
				& cpu\_small & 0.0659 & 0.0670 & 0.0477 & 0.0670 & 0.0645 & 0.0659 & 0.0594 & 0.0690 & 0.0606 & 0.0797 & 0.0629 & 0.0641 & 15.65 \\
				& heat & 0.0268 & 0.0275 & 0.0176 & 0.1860 & 0.0064 & 0.1869 & 0.0172 & 0.1979 & 0.0512 & 0.1958 & 0.0074 & 0.1854 & 1253.03 \\
				& wine\_quality & 0.2148 & 0.2299 & 0.2058 & 0.2098 & 0.2093 & 0.2322 & 0.2164 & 0.2370 & 0.2085 & 0.2322 & 0.2084 & 0.2309 & 8.60 \\
				& abalone & 0.1652 & 0.1892 & 0.1668 & 0.1859 & 0.1646 & 0.1869 & 0.1856 & 0.1979 & 0.1676 & 0.1958 & 0.1636 & 0.1855 & 12.73 \\
				& space\_ga & 0.0729 & 0.0784 & 0.0745 & 0.1501 & 0.0733 & 0.1535 & 0.0773 & 0.1519 & 0.0711 & 0.1519 & 0.0718 & 0.1553 & 90.81 \\
				& debutanizer & 0.1618 & 0.1651 & 0.1751 & 0.1502 & 0.1753 & 0.1835 & 0.1358 & 0.1519 & 0.1697 & 0.1718 & 0.1719 & 0.1852 & 2.22 \\
				& available\_power & 0.0512 & 0.0564 & 0.0565 & 0.0639 & 0.0476 & 0.0584 & 0.0578 & 0.0581 & 0.0595 & 0.0710 & 0.0466 & 0.0478 & 11.39 \\
				& maximal\_torque & 0.0522 & 0.0528 & 0.0562 & 0.0583 & 0.0582 & 0.0615 & 0.0867 & 0.9040 & 0.0695 & 0.0825 & 0.0397 & 0.0421 & 163.00 \\
				& fuel\_consumption\_country & 0.0859 & 0.0863 & 0.0779 & 0.0782 & 0.0871 & 0.0890 & 0.1055 & 0.1074 & 0.0792 & 0.0951 & 0.0855 & 0.0866 & 4.37 \\
				& acceleration & 0.0874 & 0.0895 & 0.0582 & 0.0750 & 0.0880 & 0.0883 & 0.0988 & 0.0995 & 0.0742 & 0.0871 & 0.0854 & 0.0894 & 9.06 \\
				& airfoild & 0.2205 & 0.2244 & 0.1872 & 0.1882 & 0.2292 & 0.2297 & 0.1814 & 0.1285 & 0.2449 & 0.2488 & 0.2301 & 0.2318 & -4.05 \\
				& mortgage & 0.0340 & 0.0342 & 0.0319 & 0.0414 & 0.0165 & 0.0167 & 0.0268 & 0.0281 & 0.0497 & 0.0684 & 0.0236 & 0.0246 & 13.05 \\
				& treasury & 0.0458 & 0.0468 & 0.0501 & 0.0538 & 0.0501 & 0.0523 & 0.0393 & 0.0398 & 0.0587 & 0.0718 & 0.0324 & 0.0330 & 6.57 \\
				& concreteStrength & 0.1329 & 0.1497 & 0.1203 & 0.1399 & 0.1738 & 0.1771 & 0.1842 & 0.1845 & 0.1549 & 0.1580 & 0.2379 & 0.2405 & 5.68 \\
				& \multicolumn{14}{r}{\textbf{Avg.\ 100.76}} \\
				\midrule\midrule
				& \textbf{Dataset}
				& \multicolumn{2}{c|}{\textbf{MLP}}
				& \multicolumn{2}{c|}{\textbf{XGB}}
				& \multicolumn{2}{c|}{\textbf{LR}}
				& \multicolumn{2}{c|}{\textbf{KNN}}
				& \multicolumn{2}{c|}{\textbf{SVR}}
				& \multicolumn{2}{c|}{\textbf{Ridge}}
				& \\
				\cmidrule(lr){3-4}\cmidrule(lr){5-6}\cmidrule(lr){7-8}\cmidrule(lr){9-10}\cmidrule(lr){11-12}\cmidrule(lr){13-14}
				& &
				\textbf{Full} & \textbf{w/o Ph.1} &
				\textbf{Full} & \textbf{w/o Ph.1} &
				\textbf{Full} & \textbf{w/o Ph.1} &
				\textbf{Full} & \textbf{w/o Ph.1} &
				\textbf{Full} & \textbf{w/o Ph.1} &
				\textbf{Full} & \textbf{w/o Ph.1} &
				\textbf{Avg.\ \% Imp.} \\
				\midrule
				\multirow{16}{*}{\rotatebox{90}{\textbf{$R^{2}$}}}
				& california & 0.8168 & 0.7943 & 0.8110 & 0.8036 & 0.8298 & 0.7996 & 0.8080 & 0.7741 & 0.8312 & 0.7962 & 0.8307 & 0.8297 & -2.64 \\
				& compactive & 0.9847 & 0.9739 & 0.9657 & 0.9747 & 0.9719 & 0.9656 & 0.9612 & 0.9604 & 0.9759 & 0.9481 & 0.9723 & 0.9711 & -0.64 \\
				& cpu\_small & 0.9709 & 0.9675 & 0.9797 & 0.9632 & 0.9658 & 0.9647 & 0.9714 & 0.9610 & 0.9675 & 0.9480 & 0.9680 & 0.9700 & -0.84 \\
				& heat & 0.9892 & 0.9835 & 0.9952 & 0.3738 & 0.9963 & 0.3674 & 0.9963 & 0.2907 & 0.9519 & 0.3061 & 0.9992 & 0.3772 & -63.99 \\
				& wine\_quality & 0.4375 & 0.3557 & 0.4837 & 0.4637 & 0.4659 & 0.3431 & 0.4294 & 0.3151 & 0.4705 & 0.3429 & 0.4707 & 0.3505 & -23.79 \\
				& abalone & 0.5060 & 0.3517 & 0.4962 & 0.3738 & 0.5093 & 0.3674 & 0.3765 & 0.2907 & 0.4913 & 0.3061 & 0.5152 & 0.3772 & -28.38 \\
				& space\_ga & 0.7123 & 0.7028 & 0.6951 & 0.6949 & 0.7052 & 0.7022 & 0.6715 & 0.6928 & 0.7164 & 0.7179 & 0.7168 & 0.7259 & 0.48 \\
				& debutanizer & 0.7024 & 0.7016 & 0.6514 & 0.6437 & 0.6505 & 0.6322 & 0.7689 & 0.7378 & 0.6727 & 0.6669 & 0.6643 & 0.6259 & -2.47 \\
				& available\_power & 0.9706 & 0.9657 & 0.9642 & 0.9570 & 0.9755 & 0.9616 & 0.9625 & 0.9559 & 0.9671 & 0.9433 & 0.9756 & 0.9713 & -1.04 \\
				& maximal\_torque & 0.9646 & 0.9645 & 0.9598 & 0.9557 & 0.9816 & 0.9807 & 0.9026 & 0.8940 & 0.9343 & 0.9116 & 0.9796 & 0.9757 & -0.72 \\
				& fuel\_consumption\_country & 0.8849 & 0.8794 & 0.8997 & 0.8950 & 0.8827 & 0.8798 & 0.8297 & 0.8218 & 0.8997 & 0.8599 & 0.8870 & 0.8848 & -1.18 \\
				& acceleration & 0.8892 & 0.8704 & 0.9186 & 0.9025 & 0.8657 & 0.8636 & 0.8307 & 0.8307 & 0.8865 & 0.8685 & 0.8737 & 0.8708 & -1.08 \\
				& airfoild & 0.7963 & 0.7901 & 0.8413 & 0.8392 & 0.7800 & 0.7798 & 0.8622 & 0.8607 & 0.7487 & 0.7467 & 0.7783 & 0.7770 & -0.28 \\
				& mortgage & 0.9934 & 0.9933 & 0.9948 & 0.9912 & 0.9984 & 0.9982 & 0.9960 & 0.9959 & 0.9873 & 0.9761 & 0.9972 & 0.9969 & -0.26 \\
				& treasury & 0.9878 & 0.9803 & 0.9854 & 0.9833 & 0.9936 & 0.9927 & 0.9928 & 0.9907 & 0.9796 & 0.9700 & 0.9935 & 0.9934 & -0.38 \\
				& concreteStrength & 0.8751 & 0.8598 & 0.8364 & 0.8278 & 0.8112 & 0.8025 & 0.7879 & 0.7873 & 0.8316 & 0.8208 & 0.8244 & 0.8165 & -1.03 \\
				& \multicolumn{14}{r}{\textbf{Avg.\ -8.02}} \\
				\bottomrule
			\end{tabular}
		}
	\end{table}

	\begin{table}[p]
		\centering
		\caption{Ablation study \emph{without Phase~2 (Data-Level Imbalanced Learning)}: for each dataset and regressor we report \emph{Full framework} vs.\ \emph{w/o Phase~2}. “Avg.\ Percent Improvement” is the average over regressors of $((\text{w/o}-\text{Full})/\text{Full})\times100$ (positive means worse without the phase for MAE and RMSE; negative means worse without the phase for $R^2$).}
		\label{tab:detailed_Ablation_Ph2}
		\small
		\setlength{\tabcolsep}{3pt}
		\renewcommand{\arraystretch}{0.92}
		\resizebox{\textwidth}{!}{
			\begin{tabular}{
					l
					l|
					>{\columncolor{lightgray}}c c|
					>{\columncolor{lightgray}}c c|
					>{\columncolor{lightgray}}c c|
					>{\columncolor{lightgray}}c c|
					>{\columncolor{lightgray}}c c|
					>{\columncolor{lightgray}}c c|c}
				\toprule
				& \textbf{Dataset}
				& \multicolumn{2}{c|}{\textbf{MLP}}
				& \multicolumn{2}{c|}{\textbf{XGB}}
				& \multicolumn{2}{c|}{\textbf{LR}}
				& \multicolumn{2}{c|}{\textbf{KNN}}
				& \multicolumn{2}{c|}{\textbf{SVR}}
				& \multicolumn{2}{c|}{\textbf{Ridge}}
				& \\
				\cmidrule(lr){3-4}\cmidrule(lr){5-6}\cmidrule(lr){7-8}\cmidrule(lr){9-10}\cmidrule(lr){11-12}\cmidrule(lr){13-14}
				& &
				\textbf{Full} & \textbf{w/o Ph.2} &
				\textbf{Full} & \textbf{w/o Ph.2} &
				\textbf{Full} & \textbf{w/o Ph.2} &
				\textbf{Full} & \textbf{w/o Ph.2} &
				\textbf{Full} & \textbf{w/o Ph.2} &
				\textbf{Full} & \textbf{w/o Ph.2} &
				\textbf{Avg.\ \% Imp.} \\
				\midrule
				\multirow{16}{*}{\rotatebox{90}{\textbf{MAE}}}
				& california & 0.1429 & 0.1433 & 0.1446 & 0.1455 & 0.1410 & 0.1428 & 0.1473 & 0.1482 & 0.1397 & 0.1489 & 0.1407 & 0.1574 & 3.54 \\
				& compactive & 0.0311 & 0.0387 & 0.0415 & 0.0553 & 0.0393 & 0.0393 & 0.0433 & 0.0460 & 0.0479 & 0.0587 & 0.0393 & 0.0397 & 14.58 \\
				& cpu\_small & 0.0462 & 0.0476 & 0.0348 & 0.0496 & 0.0463 & 0.0491 & 0.0422 & 0.0511 & 0.0519 & 0.0614 & 0.0446 & 0.0462 & 15.76 \\
				& heat & 0.0169 & 0.0175 & 0.0118 & 0.0152 & 0.0040 & 0.0049 & 0.0117 & 0.0128 & 0.0395 & 0.0598 & 0.0044 & 0.0049 & 21.17 \\
				& wine\_quality & 0.1595 & 0.1634 & 0.1604 & 0.1622 & 0.1648 & 0.1651 & 0.1640 & 0.1676 & 0.1644 & 0.1651 & 0.1640 & 0.1649 & 1.15 \\
				& abalone & 0.1227 & 0.1239 & 0.1247 & 0.1265 & 0.1257 & 0.1261 & 0.1342 & 0.1322 & 0.1217 & 0.1267 & 0.1249 & 0.1263 & 1.08 \\
				& space\_ga & 0.0531 & 0.0577 & 0.0567 & 0.0591 & 0.0568 & 0.0578 & 0.0543 & 0.0561 & 0.0509 & 0.5900 & 0.0556 & 0.0568 & 179.88 \\
				& debutanizer & 0.1155 & 0.1189 & 0.1219 & 0.1231 & 0.1309 & 0.1323 & 0.0785 & 0.1067 & 0.1223 & 0.1299 & 0.1277 & 0.1284 & 7.95 \\
				& available\_power & 0.0336 & 0.0336 & 0.0349 & 0.0352 & 0.0305 & 0.0309 & 0.0197 & 0.0355 & 0.0423 & 0.0594 & 0.0301 & 0.0304 & 20.63 \\
				& maximal\_torque & 0.0111 & 0.0418 & 0.0304 & 0.0338 & 0.0395 & 0.0409 & 0.0500 & 0.0508 & 0.0487 & 0.0770 & 0.0229 & 0.0234 & 58.87 \\
				& fuel\_consumption\_country & 0.0581 & 0.0612 & 0.0586 & 0.0619 & 0.0653 & 0.0659 & 0.0709 & 0.0741 & 0.0568 & 0.0792 & 0.0644 & 0.0681 & 10.26 \\
				& acceleration & 0.0589 & 0.0694 & 0.0452 & 0.0651 & 0.0678 & 0.0691 & 0.0682 & 0.0740 & 0.0563 & 0.0765 & 0.0658 & 0.0671 & 18.36 \\
				& airfoild & 0.0801 & 0.0809 & 0.1095 & 0.1106 & 0.1396 & 0.1413 & 0.0715 & 0.0758 & 0.1305 & 0.1356 & 0.1362 & 0.1378 & 2.39 \\
				& mortgage & 0.0254 & 0.0316 & 0.0215 & 0.2310 & 0.0110 & 0.0153 & 0.0152 & 0.0235 & 0.0420 & 0.0611 & 0.0115 & 0.0159 & 196.04 \\
				& treasury & 0.0290 & 0.0299 & 0.0241 & 0.0256 & 0.0112 & 0.0223 & 0.0119 & 0.0209 & 0.0490 & 0.0611 & 0.0190 & 0.0212 & 36.72 \\
				& concreteStrength & 0.1103 & 0.1399 & 0.1257 & 0.1361 & 0.1384 & 0.1393 & 0.1417 & 0.1467 & 0.1215 & 0.1378 & 0.1336 & 0.1350 & 8.96 \\
				& \multicolumn{14}{r}{\textbf{Avg.\ 37.33}} \\
				\midrule\midrule
				& \textbf{Dataset}
				& \multicolumn{2}{c|}{\textbf{MLP}}
				& \multicolumn{2}{c|}{\textbf{XGB}}
				& \multicolumn{2}{c|}{\textbf{LR}}
				& \multicolumn{2}{c|}{\textbf{KNN}}
				& \multicolumn{2}{c|}{\textbf{SVR}}
				& \multicolumn{2}{c|}{\textbf{Ridge}}
				& \\
				\cmidrule(lr){3-4}\cmidrule(lr){5-6}\cmidrule(lr){7-8}\cmidrule(lr){9-10}\cmidrule(lr){11-12}\cmidrule(lr){13-14}
				& &
				\textbf{Full} & \textbf{w/o Ph.2} &
				\textbf{Full} & \textbf{w/o Ph.2} &
				\textbf{Full} & \textbf{w/o Ph.2} &
				\textbf{Full} & \textbf{w/o Ph.2} &
				\textbf{Full} & \textbf{w/o Ph.2} &
				\textbf{Full} & \textbf{w/o Ph.2} &
				\textbf{Avg.\ \% Imp.} \\
				\midrule
				\multirow{16}{*}{\rotatebox{90}{\textbf{RMSE}}}
				& california & 0.2043 & 0.2056 & 0.2052 & 0.2063 & 0.1978 & 0.1988 & 0.2118 & 0.2141 & 0.1974 & 0.2091 & 0.1976 & 0.2011 & 1.74 \\
				& compactive & 0.0490 & 0.0593 & 0.0717 & 0.0758 & 0.0632 & 0.0651 & 0.0744 & 0.0763 & 0.0541 & 0.0860 & 0.0629 & 0.0689 & 16.80 \\
				& cpu\_small & 0.0659 & 0.0679 & 0.0477 & 0.0785 & 0.0645 & 0.0648 & 0.0594 & 0.0725 & 0.0606 & 0.0841 & 0.0629 & 0.0665 & 22.44 \\
				& heat & 0.0268 & 0.0285 & 0.0176 & 0.0232 & 0.0064 & 0.0070 & 0.0172 & 0.0201 & 0.0512 & 0.0668 & 0.0074 & 0.0082 & 17.61 \\
				& wine\_quality & 0.2148 & 0.2172 & 0.2058 & 0.2072 & 0.2093 & 0.2097 & 0.2164 & 0.2163 & 0.2085 & 0.2099 & 0.2084 & 0.2089 & 0.48 \\
				& abalone & 0.1652 & 0.1692 & 0.1668 & 0.1678 & 0.1646 & 0.1649 & 0.1856 & 0.1819 & 0.1676 & 0.1692 & 0.1636 & 0.1612 & 0.12 \\
				& space\_ga & 0.0729 & 0.0759 & 0.0745 & 0.0798 & 0.0733 & 0.0799 & 0.0773 & 0.0812 & 0.0711 & 0.0759 & 0.0718 & 0.0721 & 5.40 \\
				& debutanizer & 0.1618 & 0.1649 & 0.1751 & 0.1762 & 0.1753 & 0.1765 & 0.1358 & 0.1548 & 0.1697 & 0.1775 & 0.1719 & 0.1729 & 3.73 \\
				& available\_power & 0.0512 & 0.0516 & 0.0565 & 0.0564 & 0.0476 & 0.0482 & 0.0578 & 0.0574 & 0.0595 & 0.0602 & 0.0466 & 0.0468 & 0.46 \\
				& maximal\_torque & 0.0522 & 0.0655 & 0.0562 & 0.0598 & 0.0582 & 0.0595 & 0.0867 & 0.0862 & 0.0695 & 0.0931 & 0.0397 & 0.0399 & 11.33 \\
				& fuel\_consumption\_country & 0.0859 & 0.0868 & 0.0779 & 0.0851 & 0.0871 & 0.0889 & 0.1055 & 0.1103 & 0.0792 & 0.1086 & 0.0855 & 0.0892 & 9.73 \\
				& acceleration & 0.0874 & 0.0902 & 0.0582 & 0.0849 & 0.0880 & 0.0891 & 0.0988 & 0.0994 & 0.0742 & 0.0950 & 0.0854 & 0.0865 & 13.38 \\
				& airfoild & 0.2205 & 0.2235 & 0.1872 & 0.2535 & 0.2292 & 0.2321 & 0.1814 & 0.1896 & 0.2449 & 0.2511 & 0.2301 & 0.2308 & 7.57 \\
				& mortgage & 0.0340 & 0.0413 & 0.0319 & 0.0329 & 0.0165 & 0.0222 & 0.0268 & 0.0361 & 0.0497 & 0.0721 & 0.0236 & 0.0252 & 24.28 \\
				& treasury & 0.0458 & 0.0462 & 0.0501 & 0.0509 & 0.0501 & 0.0564 & 0.0393 & 0.0432 & 0.0587 & 0.0756 & 0.0324 & 0.0386 & 12.15 \\
				& concreteStrength & 0.1329 & 0.1869 & 0.1203 & 0.1744 & 0.1738 & 0.1799 & 0.1842 & 0.1872 & 0.1549 & 0.1768 & 0.2379 & 0.2452 & 17.99 \\
				& \multicolumn{14}{r}{\textbf{Avg.\ 10.33}} \\
				\midrule\midrule
				& \textbf{Dataset}
				& \multicolumn{2}{c|}{\textbf{MLP}}
				& \multicolumn{2}{c|}{\textbf{XGB}}
				& \multicolumn{2}{c|}{\textbf{LR}}
				& \multicolumn{2}{c|}{\textbf{KNN}}
				& \multicolumn{2}{c|}{\textbf{SVR}}
				& \multicolumn{2}{c|}{\textbf{Ridge}}
				& \\
				\cmidrule(lr){3-4}\cmidrule(lr){5-6}\cmidrule(lr){7-8}\cmidrule(lr){9-10}\cmidrule(lr){11-12}\cmidrule(lr){13-14}
				& &
				\textbf{Full} & \textbf{w/o 2} &
				\textbf{Full} & \textbf{w/o 2} &
				\textbf{Full} & \textbf{w/o 2} &
				\textbf{Full} & \textbf{w/o 2} &
				\textbf{Full} & \textbf{w/o 2} &
				\textbf{Full} & \textbf{w/o 2} &
				\textbf{Avg.\ \% Imp.} \\
				\midrule
				\multirow{16}{*}{\rotatebox{90}{\textbf{$R^{2}$}}}
				& california & 0.8168 & 0.8155 & 0.8110 & 0.8103 & 0.8298 & 0.8260 & 0.8080 & 0.7997 & 0.8312 & 0.8090 & 0.8307 & 0.8219 & -0.91 \\
				& compactive & 0.9847 & 0.9752 & 0.9657 & 0.9573 & 0.9719 & 0.9655 & 0.9612 & 0.9591 & 0.9759 & 0.9480 & 0.9723 & 0.9645 & -1.06 \\
				& cpu\_small & 0.9709 & 0.9644 & 0.9797 & 0.9496 & 0.9658 & 0.9650 & 0.9714 & 0.9569 & 0.9675 & 0.9434 & 0.9680 & 0.9660 & -1.34 \\
				& heat & 0.9892 & 0.9835 & 0.9952 & 0.9917 & 0.9963 & 0.9919 & 0.9963 & 0.9938 & 0.9519 & 0.9310 & 0.9992 & 0.9984 & -27.46 \\
				& wine\_quality & 0.4375 & 0.4268 & 0.4837 & 0.4769 & 0.4659 & 0.4652 & 0.4294 & 0.4212 & 0.4705 & 0.4671 & 0.4707 & 0.4623 & 2.44 \\
				& abalone & 0.5060 & 0.4820 & 0.4962 & 0.4876 & 0.5093 & 0.5036 & 0.3765 & 0.3732 & 0.4913 & 0.4906 & 0.5152 & 0.5134 & -1.49 \\
				& space\_ga & 0.7123 & 0.6920 & 0.6951 & 0.6847 & 0.7052 & 0.7021 & 0.6715 & 0.6623 & 0.7164 & 0.6852 & 0.7168 & 0.7108 & -1.89 \\
				& debutanizer & 0.7024 & 0.6950 & 0.6514 & 0.6462 & 0.6505 & 0.6487 & 0.7689 & 0.7295 & 0.6727 & 0.6447 & 0.6643 & 0.6623 & -1.95 \\
				& available\_power & 0.9706 & 0.9689 & 0.9642 & 0.9640 & 0.9755 & 0.9752 & 0.9625 & 0.9620 & 0.9671 & 0.9296 & 0.9756 & 0.9752 & -0.70 \\
				& maximal\_torque & 0.9646 & 0.9443 & 0.9598 & 0.9536 & 0.9816 & 0.9797 & 0.9026 & 0.9019 & 0.9343 & 0.8878 & 0.9796 & 0.9784 & -1.35 \\
				& fuel\_consumption\_country & 0.8849 & 0.8834 & 0.8997 & 0.8881 & 0.8827 & 0.8804 & 0.8297 & 0.8271 & 0.8997 & 0.8178 & 0.8870 & 0.8792 & -2.00 \\
				& acceleration & 0.8892 & 0.8591 & 0.9186 & 0.8753 & 0.8657 & 0.8626 & 0.8307 & 0.8298 & 0.8865 & 0.8439 & 0.8737 & 0.8704 & -2.29 \\
				& airfoild & 0.7963 & 0.7798 & 0.8413 & 0.7306 & 0.7800 & 0.7719 & 0.8622 & 0.8608 & 0.7487 & 0.7358 & 0.7783 & 0.7768 & -3.06 \\
				& mortgage & 0.9934 & 0.9913 & 0.9948 & 0.9935 & 0.9984 & 0.9974 & 0.9960 & 0.9933 & 0.9873 & 0.9735 & 0.9972 & 0.9971 & -0.35 \\
				& treasury & 0.9878 & 0.9853 & 0.9854 & 0.9843 & 0.9936 & 0.9923 & 0.9928 & 0.9891 & 0.9796 & 0.9667 & 0.9935 & 0.9913 & -0.40 \\
				& concreteStrength & 0.8751 & 0.7814 & 0.8364 & 0.8097 & 0.8112 & 0.7977 & 0.7879 & 0.7811 & 0.8316 & 0.8045 & 0.8244 & 0.8080 & -3.61 \\
				& \multicolumn{14}{r}{\textbf{Avg.\ -2.96}} \\
				\bottomrule
			\end{tabular}
		}
	\end{table}

	\begin{table}[p]
		\centering
		\caption{Ablation study \emph{without Phase~3 (Algorithm-Level Imbalanced Learning)}: for each dataset and regressor we report \emph{Full framework} vs.\ \emph{w/o Phase~3}. “Avg.\ Percent Improvement” is the average over regressors of $((\text{w/o}-\text{Full})/\text{Full})\times100$ (positive means worse without the phase for MAE and RMSE; negative means worse without the phase for $R^2$).}
		\label{tab:detailed_Ablation_Ph3}
		\small
		\setlength{\tabcolsep}{3pt}
		\renewcommand{\arraystretch}{0.92}
		\resizebox{\textwidth}{!}{
			\begin{tabular}{
					l
					l|
					>{\columncolor{lightgray}}c c|
					>{\columncolor{lightgray}}c c|
					>{\columncolor{lightgray}}c c|
					>{\columncolor{lightgray}}c c|
					>{\columncolor{lightgray}}c c|
					>{\columncolor{lightgray}}c c|c}
				\toprule
				& \textbf{Dataset}
				& \multicolumn{2}{c|}{\textbf{MLP}}
				& \multicolumn{2}{c|}{\textbf{XGB}}
				& \multicolumn{2}{c|}{\textbf{LR}}
				& \multicolumn{2}{c|}{\textbf{KNN}}
				& \multicolumn{2}{c|}{\textbf{SVR}}
				& \multicolumn{2}{c|}{\textbf{Ridge}}
				& \\
				\cmidrule(lr){3-4}\cmidrule(lr){5-6}\cmidrule(lr){7-8}\cmidrule(lr){9-10}\cmidrule(lr){11-12}\cmidrule(lr){13-14}
				& &
				\textbf{Full} & \textbf{w/o Ph.3} &
				\textbf{Full} & \textbf{w/o Ph.3} &
				\textbf{Full} & \textbf{w/o Ph.3} &
				\textbf{Full} & \textbf{w/o Ph.3} &
				\textbf{Full} & \textbf{w/o Ph.3} &
				\textbf{Full} & \textbf{w/o Ph.3} &
				\textbf{Avg.\ \% Imp.} \\
				\midrule
				\multirow{16}{*}{\rotatebox{90}{\textbf{MAE}}}
				& california & 0.1429 & 0.1466 & 0.1446 & 0.1451 & 0.1410 & 0.1418 & 0.1473 & 0.1483 & 0.1397 & 0.1514 & 0.1407 & 0.1414 & 2.18 \\
				& compactive & 0.0311 & 0.0385 & 0.0415 & 0.0428 & 0.0393 & 0.0399 & 0.0433 & 0.0498 & 0.0479 & 0.0597 & 0.0393 & 0.0467 & 14.49 \\
				& cpu\_small & 0.0462 & 0.0497 & 0.0348 & 0.0538 & 0.0463 & 0.0493 & 0.0422 & 0.0496 & 0.0519 & 0.0678 & 0.0446 & 0.0498 & 21.41 \\
				& heat & 0.0169 & 0.0174 & 0.0118 & 0.0136 & 0.0040 & 0.0045 & 0.0117 & 0.0129 & 0.0395 & 0.0603 & 0.0044 & 0.0053 & 19.01 \\
				& wine\_quality & 0.1595 & 0.1681 & 0.1604 & 0.1719 & 0.1648 & 0.1649 & 0.1640 & 0.1672 & 0.1644 & 0.1699 & 0.1640 & 0.1607 & 2.65 \\
				& abalone & 0.1227 & 0.1277 & 0.1247 & 0.1278 & 0.1257 & 0.1312 & 0.1342 & 0.1320 & 0.1217 & 0.1289 & 0.1249 & 0.1267 & 2.78 \\
				& space\_ga & 0.0531 & 0.0563 & 0.0567 & 0.0576 & 0.0568 & 0.0573 & 0.0543 & 0.0569 & 0.0509 & 0.0609 & 0.0556 & 0.0542 & 5.07 \\
				& debutanizer & 0.1155 & 0.1159 & 0.1219 & 0.1236 & 0.1309 & 0.1325 & 0.0785 & 0.1018 & 0.1223 & 0.1232 & 0.1277 & 0.1282 & 5.63 \\
				& available\_power & 0.0336 & 0.0341 & 0.0349 & 0.0349 & 0.0305 & 0.0308 & 0.0197 & 0.0360 & 0.0423 & 0.0594 & 0.0301 & 0.0306 & 21.22 \\
				& maximal\_torque & 0.0111 & 0.0418 & 0.0304 & 0.0367 & 0.0395 & 0.0428 & 0.0500 & 0.0504 & 0.0487 & 0.0770 & 0.0229 & 0.0234 & 61.12 \\
				& fuel\_consumption\_country & 0.0581 & 0.0612 & 0.0586 & 0.0620 & 0.0653 & 0.0672 & 0.0709 & 0.0710 & 0.0568 & 0.0793 & 0.0644 & 0.0644 & 8.97 \\
				& acceleration & 0.0589 & 0.0686 & 0.0452 & 0.0665 & 0.0678 & 0.0678 & 0.0682 & 0.0740 & 0.0563 & 0.0747 & 0.0658 & 0.0660 & 17.51 \\
				& airfoild & 0.0801 & 0.0803 & 0.1095 & 0.1104 & 0.1396 & 0.1396 & 0.0715 & 0.0715 & 0.1305 & 0.1305 & 0.1362 & 0.1362 & 0.18 \\
				& mortgage & 0.0254 & 0.0317 & 0.0215 & 0.0215 & 0.0110 & 0.0153 & 0.0152 & 0.0235 & 0.0420 & 0.0611 & 0.0115 & 0.0158 & 33.56 \\
				& treasury & 0.0290 & 0.0290 & 0.0241 & 0.0241 & 0.0112 & 0.0222 & 0.0119 & 0.0209 & 0.0490 & 0.0611 & 0.0190 & 0.0212 & 35.02 \\
				& concreteStrength & 0.1103 & 0.1326 & 0.1257 & 0.1275 & 0.1384 & 0.1384 & 0.1417 & 0.1417 & 0.1215 & 0.1372 & 0.1336 & 0.1336 & 5.76 \\
				& \multicolumn{14}{r}{\textbf{Avg.\ 16.03}} \\
				\midrule\midrule
				& \textbf{Dataset}
				& \multicolumn{2}{c|}{\textbf{MLP}}
				& \multicolumn{2}{c|}{\textbf{XGB}}
				& \multicolumn{2}{c|}{\textbf{LR}}
				& \multicolumn{2}{c|}{\textbf{KNN}}
				& \multicolumn{2}{c|}{\textbf{SVR}}
				& \multicolumn{2}{c|}{\textbf{Ridge}}
				& \\
				\cmidrule(lr){3-4}\cmidrule(lr){5-6}\cmidrule(lr){7-8}\cmidrule(lr){9-10}\cmidrule(lr){11-12}\cmidrule(lr){13-14}
				& &
				\textbf{Full} & \textbf{w/o Ph.3} &
				\textbf{Full} & \textbf{w/o Ph.3} &
				\textbf{Full} & \textbf{w/o Ph.3} &
				\textbf{Full} & \textbf{w/o Ph.3} &
				\textbf{Full} & \textbf{w/o Ph.3} &
				\textbf{Full} & \textbf{w/o Ph.3} &
				\textbf{Avg.\ \% Imp.} \\
				\midrule
				\multirow{16}{*}{\rotatebox{90}{\textbf{RMSE}}}
				& california & 0.2043 & 0.2058 & 0.2052 & 0.2064 & 0.1978 & 0.1998 & 0.2118 & 0.2134 & 0.1974 & 0.2120 & 0.1976 & 0.1993 & 1.89 \\
				& compactive & 0.0490 & 0.0617 & 0.0717 & 0.0756 & 0.0632 & 0.0658 & 0.0744 & 0.0826 & 0.0541 & 0.0978 & 0.0629 & 0.0723 & 23.70 \\
				& cpu\_small & 0.0659 & 0.0698 & 0.0477 & 0.0772 & 0.0645 & 0.0693 & 0.0594 & 0.0729 & 0.0606 & 0.0918 & 0.0629 & 0.0699 & 26.76 \\
				& heat & 0.0268 & 0.0278 & 0.0176 & 0.0196 & 0.0064 & 0.0071 & 0.0172 & 0.0195 & 0.0512 & 0.0682 & 0.0074 & 0.0083 & 14.23 \\
				& wine\_quality & 0.2148 & 0.2273 & 0.2058 & 0.2193 & 0.2093 & 0.2098 & 0.2164 & 0.2193 & 0.2085 & 0.2134 & 0.2084 & 0.2014 & 2.16 \\
				& abalone & 0.1652 & 0.1691 & 0.1668 & 0.1692 & 0.1646 & 0.1742 & 0.1856 & 0.1847 & 0.1676 & 0.1685 & 0.1636 & 0.1624 & 1.49 \\
				& space\_ga & 0.0729 & 0.0791 & 0.0745 & 0.0796 & 0.0733 & 0.0746 & 0.0773 & 0.0784 & 0.0711 & 0.0793 & 0.0718 & 0.0708 & 4.78 \\
				& debutanizer & 0.1618 & 0.1624 & 0.1751 & 0.1768 & 0.1753 & 0.1763 & 0.1358 & 0.1509 & 0.1697 & 0.1708 & 0.1719 & 0.1739 & 2.47 \\
				& available\_power & 0.0512 & 0.0519 & 0.0565 & 0.0566 & 0.0476 & 0.0479 & 0.0578 & 0.0579 & 0.0595 & 0.0792 & 0.0466 & 0.0477 & 156.75 \\
				& maximal\_torque & 0.0522 & 0.0658 & 0.0562 & 0.0598 & 0.0582 & 0.0594 & 0.0867 & 0.0869 & 0.0695 & 0.0931 & 0.0397 & 0.0398 & 11.49 \\
				& fuel\_consumption\_country & 0.0859 & 0.0886 & 0.0779 & 0.0850 & 0.0871 & 0.0882 & 0.1055 & 0.1067 & 0.0792 & 0.1086 & 0.0855 & 0.0858 & 8.69 \\
				& acceleration & 0.0874 & 0.0899 & 0.0582 & 0.0879 & 0.0880 & 0.0880 & 0.0988 & 0.0996 & 0.0742 & 0.0927 & 0.0854 & 0.0854 & 13.27 \\
				& airfoild & 0.2205 & 0.2205 & 0.1872 & 0.2386 & 0.2292 & 0.2292 & 0.1814 & 0.1814 & 0.2449 & 0.2449 & 0.2301 & 0.2301 & 4.58 \\
				& mortgage & 0.0340 & 0.0413 & 0.0319 & 0.0319 & 0.0165 & 0.0223 & 0.0268 & 0.0361 & 0.0497 & 0.0721 & 0.0236 & 0.0236 & 22.73 \\
				& treasury & 0.0458 & 0.0458 & 0.0501 & 0.0501 & 0.0501 & 0.0564 & 0.0393 & 0.0432 & 0.0587 & 0.0757 & 0.0324 & 0.0386 & 11.77 \\
				& concreteStrength & 0.1329 & 0.1424 & 0.1203 & 0.1720 & 0.1738 & 0.1738 & 0.1842 & 0.1842 & 0.1549 & 0.1755 & 0.2379 & 0.2379 & 10.57 \\
				& \multicolumn{14}{r}{\textbf{Avg.\ 19.83}} \\
				\midrule\midrule
				& \textbf{Dataset}
				& \multicolumn{2}{c|}{\textbf{MLP}}
				& \multicolumn{2}{c|}{\textbf{XGB}}
				& \multicolumn{2}{c|}{\textbf{LR}}
				& \multicolumn{2}{c|}{\textbf{KNN}}
				& \multicolumn{2}{c|}{\textbf{SVR}}
				& \multicolumn{2}{c|}{\textbf{Ridge}}
				& \\
				\cmidrule(lr){3-4}\cmidrule(lr){5-6}\cmidrule(lr){7-8}\cmidrule(lr){9-10}\cmidrule(lr){11-12}\cmidrule(lr){13-14}
				& &
				\textbf{Full} & \textbf{w/o Ph.3} &
				\textbf{Full} & \textbf{w/o Ph.3} &
				\textbf{Full} & \textbf{w/o Ph.3} &
				\textbf{Full} & \textbf{w/o Ph.3} &
				\textbf{Full} & \textbf{w/o Ph.3} &
				\textbf{Full} & \textbf{w/o Ph.3} &
				\textbf{Avg.\ \% Imp.} \\
				\midrule
				\multirow{16}{*}{\rotatebox{90}{\textbf{$R^{2}$}}}
				& california & 0.8168 & 0.8097 & 0.8110 & 0.8095 & 0.8298 & 0.8258 & 0.8080 & 0.8097 & 0.8312 & 0.8036 & 0.8307 & 0.8265 & -0.86 \\
				& compactive & 0.9847 & 0.9632 & 0.9657 & 0.9521 & 0.9719 & 0.9687 & 0.9612 & 0.9480 & 0.9759 & 0.9332 & 0.9723 & 0.9616 & -1.80 \\
				& cpu\_small & 0.9709 & 0.9515 & 0.9797 & 0.9519 & 0.9658 & 0.9551 & 0.9714 & 0.9454 & 0.9675 & 0.9398 & 0.9680 & 0.9556 & -2.13 \\
				& heat & 0.9892 & 0.9819 & 0.9952 & 0.9940 & 0.9963 & 0.9960 & 0.9963 & 0.9940 & 0.9519 & 0.9301 & 0.9992 & 0.9987 & -27.84 \\
				& wine\_quality & 0.4375 & 0.4109 & 0.4837 & 0.4712 & 0.4659 & 0.4648 & 0.4294 & 0.4223 & 0.4705 & 0.4654 & 0.4707 & 0.4627 & -2.22 \\
				& abalone & 0.5060 & 0.4842 & 0.4962 & 0.4886 & 0.5093 & 0.4931 & 0.3765 & 0.3760 & 0.4913 & 0.4845 & 0.5152 & 0.5141 & -1.79 \\
				& space\_ga & 0.7123 & 0.6924 & 0.6951 & 0.6825 & 0.7052 & 0.7021 & 0.6715 & 0.6701 & 0.7164 & 0.6632 & 0.7168 & 0.7154 & -2.15 \\
				& debutanizer & 0.7024 & 0.7031 & 0.6514 & 0.6488 & 0.6505 & 0.6501 & 0.7689 & 0.7410 & 0.6727 & 0.6712 & 0.6643 & 0.6632 & -0.73 \\
				& available\_power & 0.9706 & 0.9700 & 0.9642 & 0.9640 & 0.9755 & 0.9732 & 0.9625 & 0.9625 & 0.9671 & 0.9295 & 0.9756 & 0.9749 & -0.71 \\
				& maximal\_torque & 0.9646 & 0.9443 & 0.9598 & 0.9536 & 0.9816 & 0.9797 & 0.9026 & 0.9022 & 0.9343 & 0.8876 & 0.9796 & 0.9795 & -1.33 \\
				& fuel\_consumption\_country & 0.8849 & 0.8826 & 0.8997 & 0.8881 & 0.8827 & 0.8816 & 0.8297 & 0.8279 & 0.8997 & 0.8178 & 0.8870 & 0.8870 & -1.83 \\
				& acceleration & 0.8892 & 0.8599 & 0.9186 & 0.8663 & 0.8657 & 0.8657 & 0.8307 & 0.8301 & 0.8865 & 0.8512 & 0.8737 & 0.8736 & -2.18 \\
				& airfoild & 0.7963 & 0.7963 & 0.8413 & 0.7614 & 0.7800 & 0.7800 & 0.8622 & 0.8622 & 0.7487 & 0.7487 & 0.7783 & 0.7783 & -1.58 \\
				& mortgage & 0.9934 & 0.9913 & 0.9948 & 0.9948 & 0.9984 & 0.9974 & 0.9960 & 0.9933 & 0.9873 & 0.9735 & 0.9972 & 0.9971 & -0.33 \\
				& treasury & 0.9878 & 0.9878 & 0.9854 & 0.9854 & 0.9936 & 0.9885 & 0.9928 & 0.9891 & 0.9796 & 0.9667 & 0.9935 & 0.9914 & -0.40 \\
				& concreteStrength & 0.8751 & 0.8202 & 0.8364 & 0.8151 & 0.8112 & 0.8112 & 0.7879 & 0.7879 & 0.8316 & 0.8073 & 0.8244 & 0.8244 & -1.96 \\
				& \multicolumn{14}{r}{\textbf{Avg.\ -3.11}} \\
				\bottomrule
			\end{tabular}
		}
	\end{table}

	\begin{table}[p]
		\centering
		\caption{Ablation study \emph{without Phase~4 (Feature Fusion)}: for each dataset and regressor we report \emph{Full framework} vs.\ \emph{w/o Phase~4}. “Avg.\ Percent Improvement” is the average over regressors of $((\text{w/o}-\text{Full})/\text{Full})\times100$ (positive means worse without the phase for MAE and RMSE; negative means worse without the phase for $R^2$).}
		\label{tab:detailed_Ablation_Ph4}
		\small
		\setlength{\tabcolsep}{3pt}
		\renewcommand{\arraystretch}{0.92}
		\resizebox{\textwidth}{!}{
			\begin{tabular}{
					l
					l|
					>{\columncolor{lightgray}}c c|
					>{\columncolor{lightgray}}c c|
					>{\columncolor{lightgray}}c c|
					>{\columncolor{lightgray}}c c|
					>{\columncolor{lightgray}}c c|
					>{\columncolor{lightgray}}c c|c}
				\toprule
				& \textbf{Dataset}
				& \multicolumn{2}{c|}{\textbf{MLP}}
				& \multicolumn{2}{c|}{\textbf{XGB}}
				& \multicolumn{2}{c|}{\textbf{LR}}
				& \multicolumn{2}{c|}{\textbf{KNN}}
				& \multicolumn{2}{c|}{\textbf{SVR}}
				& \multicolumn{2}{c|}{\textbf{Ridge}}
				& \\
				\cmidrule(lr){3-4}\cmidrule(lr){5-6}\cmidrule(lr){7-8}\cmidrule(lr){9-10}\cmidrule(lr){11-12}\cmidrule(lr){13-14}
				& &
				\textbf{Full} & \textbf{w/o Ph.4} &
				\textbf{Full} & \textbf{w/o Ph.4} &
				\textbf{Full} & \textbf{w/o Ph.4} &
				\textbf{Full} & \textbf{w/o Ph.4} &
				\textbf{Full} & \textbf{w/o Ph.4} &
				\textbf{Full} & \textbf{w/o Ph.4} &
				\textbf{Avg.\ \% Imp.} \\
				\midrule
				\multirow{16}{*}{\rotatebox{90}{\textbf{MAE}}}
				& california & 0.1429 & 0.1447 & 0.1446 & 0.1458 & 0.1410 & 0.1711 & 0.1473 & 0.1677 & 0.1397 & 0.1502 & 0.1407 & 0.1723 & 11.21 \\
				& compactive & 0.0311 & 0.0418 & 0.0415 & 0.0483 & 0.0393 & 0.0683 & 0.0433 & 0.0498 & 0.0479 & 0.0555 & 0.0393 & 0.0668 & 37.57 \\
				& cpu\_small & 0.0462 & 0.0494 & 0.0348 & 0.0444 & 0.0463 & 0.0567 & 0.0422 & 0.0473 & 0.0519 & 0.0577 & 0.0446 & 0.0568 & 17.93 \\
				& heat & 0.0169 & 0.0215 & 0.0118 & 0.0142 & 0.0040 & 0.0048 & 0.0117 & 0.0154 & 0.0395 & 0.0662 & 0.0044 & 0.0234 & 99.77 \\
				& wine\_quality & 0.1595 & 0.1601 & 0.1604 & 0.1654 & 0.1648 & 0.1664 & 0.1640 & 0.1760 & 0.1644 & 0.1671 & 0.1640 & 0.1651 & 2.35 \\
				& abalone & 0.1227 & 0.1141 & 0.1247 & 0.1286 & 0.1257 & 0.1264 & 0.1342 & 0.1410 & 0.1217 & 0.1244 & 0.1249 & 0.1249 & 0.66 \\
				& space\_ga & 0.0531 & 0.0538 & 0.0567 & 0.0569 & 0.0568 & 0.0578 & 0.0543 & 0.0560 & 0.0509 & 0.0547 & 0.0556 & 0.0557 & 2.37 \\
				& debutanizer & 0.1155 & 0.1163 & 0.1219 & 0.1238 & 0.1309 & 0.1398 & 0.0785 & 0.0944 & 0.1223 & 0.1320 & 0.1277 & 0.1423 & 8.11 \\
				& available\_power & 0.0336 & 0.0382 & 0.0349 & 0.0425 & 0.0305 & 0.0316 & 0.0197 & 0.0562 & 0.0423 & 0.0583 & 0.0301 & 0.0325 & 45.00 \\
				& maximal\_torque & 0.0111 & 0.0263 & 0.0304 & 0.0325 & 0.0395 & 0.0424 & 0.0500 & 0.0563 & 0.0487 & 0.0583 & 0.0229 & 0.0292 & 35.17 \\
				& fuel\_consumption\_country & 0.0581 & 0.0637 & 0.0586 & 0.0597 & 0.0653 & 0.0665 & 0.0709 & 0.0728 & 0.0568 & 0.0702 & 0.0644 & 0.0659 & 6.98 \\
				& acceleration & 0.0589 & 0.0632 & 0.0452 & 0.0508 & 0.0678 & 0.0680 & 0.0682 & 0.0740 & 0.0563 & 0.0648 & 0.0658 & 0.0661 & 7.34 \\
				& airfoild & 0.0801 & 0.0986 & 0.1095 & 0.1474 & 0.1396 & 0.2502 & 0.0715 & 0.1169 & 0.1305 & 0.1653 & 0.1362 & 0.2408 & 50.65 \\
				& mortgage & 0.0254 & 0.0342 & 0.0215 & 0.0219 & 0.0110 & 0.0128 & 0.0152 & 0.0218 & 0.0420 & 0.0426 & 0.0115 & 0.0168 & 23.97 \\
				& treasury & 0.0290 & 0.0435 & 0.0241 & 0.0253 & 0.0112 & 0.0182 & 0.0119 & 0.0179 & 0.0490 & 0.0552 & 0.0190 & 0.0214 & 32.20 \\
				& concreteStrength & 0.1103 & 0.1213 & 0.1257 & 0.1302 & 0.1384 & 0.1439 & 0.1417 & 0.1572 & 0.1215 & 0.1296 & 0.1336 & 0.1398 & 6.63 \\
				& \multicolumn{14}{r}{\textbf{Avg.\ 24.24}} \\
				\midrule\midrule
				& \textbf{Dataset}
				& \multicolumn{2}{c|}{\textbf{MLP}}
				& \multicolumn{2}{c|}{\textbf{XGB}}
				& \multicolumn{2}{c|}{\textbf{LR}}
				& \multicolumn{2}{c|}{\textbf{KNN}}
				& \multicolumn{2}{c|}{\textbf{SVR}}
				& \multicolumn{2}{c|}{\textbf{Ridge}}
				& \\
				\cmidrule(lr){3-4}\cmidrule(lr){5-6}\cmidrule(lr){7-8}\cmidrule(lr){9-10}\cmidrule(lr){11-12}\cmidrule(lr){13-14}
				& &
				\textbf{Full} & \textbf{w/o Ph.4} &
				\textbf{Full} & \textbf{w/o Ph.4} &
				\textbf{Full} & \textbf{w/o Ph.4} &
				\textbf{Full} & \textbf{w/o Ph.4} &
				\textbf{Full} & \textbf{w/o Ph.4} &
				\textbf{Full} & \textbf{w/o Ph.4} &
				\textbf{Avg.\ \% Imp.} \\
				\midrule
				\multirow{16}{*}{\rotatebox{90}{\textbf{RMSE}}}
				& california & 0.2043 & 0.2060 & 0.2052 & 0.2086 & 0.1978 & 0.2292 & 0.2118 & 0.2460 & 0.1974 & 0.2126 & 0.1976 & 0.2301 & 9.78 \\
				& compactive & 0.0490 & 0.0624 & 0.0717 & 0.0748 & 0.0632 & 0.1126 & 0.0744 & 0.1209 & 0.0541 & 0.0791 & 0.0629 & 0.1073 & 48.19 \\
				& cpu\_small & 0.0659 & 0.0668 & 0.0477 & 0.0638 & 0.0645 & 0.0863 & 0.0594 & 0.0797 & 0.0606 & 0.0761 & 0.0629 & 0.0867 & 27.75 \\
				& heat & 0.0268 & 0.0293 & 0.0176 & 0.0190 & 0.0064 & 0.0067 & 0.0172 & 0.0244 & 0.0512 & 0.0724 & 0.0074 & 0.0343 & 78.16 \\
				& wine\_quality & 0.2148 & 0.2217 & 0.2058 & 0.2266 & 0.2093 & 0.2158 & 0.2164 & 0.2317 & 0.2085 & 0.2144 & 0.2084 & 0.2093 & 4.46 \\
				& abalone & 0.1652 & 0.1654 & 0.1668 & 0.1682 & 0.1646 & 0.1635 & 0.1856 & 0.1966 & 0.1676 & 0.1677 & 0.1636 & 0.1648 & 1.17 \\
				& space\_ga & 0.0729 & 0.0734 & 0.0745 & 0.0750 & 0.0733 & 0.0755 & 0.0773 & 0.0808 & 0.0711 & 0.0711 & 0.0718 & 0.0737 & 1.92 \\
				& debutanizer & 0.1618 & 0.1633 & 0.1751 & 0.1787 & 0.1753 & 0.1959 & 0.1358 & 0.1470 & 0.1697 & 0.1948 & 0.1719 & 0.1906 & 8.11 \\
				& available\_power & 0.0512 & 0.0523 & 0.0565 & 0.0579 & 0.0476 & 0.0482 & 0.0578 & 0.1058 & 0.0595 & 0.0686 & 0.0466 & 0.0488 & 18.16 \\
				& maximal\_torque & 0.0522 & 0.0575 & 0.0562 & 0.0579 & 0.0582 & 0.0612 & 0.0867 & 0.1059 & 0.0695 & 0.0698 & 0.0397 & 0.0448 & 8.96 \\
				& fuel\_consumption\_country & 0.0859 & 0.0873 & 0.0779 & 0.0780 & 0.0871 & 0.0949 & 0.1055 & 0.1067 & 0.0792 & 0.0941 & 0.0855 & 0.0861 & 5.23 \\
				& acceleration & 0.0874 & 0.0898 & 0.0582 & 0.0719 & 0.0880 & 0.0932 & 0.0988 & 0.1019 & 0.0742 & 0.0813 & 0.0854 & 0.0864 & 7.68 \\
				& airfoild & 0.2205 & 0.2250 & 0.1872 & 0.3019 & 0.2292 & 0.3528 & 0.1814 & 0.2613 & 0.2449 & 0.3474 & 0.2301 & 0.3182 & 40.24 \\
				& mortgage & 0.0340 & 0.0449 & 0.0319 & 0.0342 & 0.0165 & 0.0172 & 0.0268 & 0.0374 & 0.0497 & 0.0519 & 0.0236 & 0.0239 & 14.79 \\
				& treasury & 0.0458 & 0.0599 & 0.0501 & 0.0513 & 0.0501 & 0.0531 & 0.0393 & 0.0426 & 0.0587 & 0.0593 & 0.0324 & 0.0391 & 11.54 \\
				& concreteStrength & 0.1329 & 0.1558 & 0.1203 & 0.1374 & 0.1738 & 0.2291 & 0.1842 & 0.1957 & 0.1549 & 0.1651 & 0.2379 & 0.2581 & 14.10 \\
				& \multicolumn{14}{r}{\textbf{Avg.\ 18.76}} \\
				\midrule\midrule
				& \textbf{Dataset}
				& \multicolumn{2}{c|}{\textbf{MLP}}
				& \multicolumn{2}{c|}{\textbf{XGB}}
				& \multicolumn{2}{c|}{\textbf{LR}}
				& \multicolumn{2}{c|}{\textbf{KNN}}
				& \multicolumn{2}{c|}{\textbf{SVR}}
				& \multicolumn{2}{c|}{\textbf{Ridge}}
				& \\
				\cmidrule(lr){3-4}\cmidrule(lr){5-6}\cmidrule(lr){7-8}\cmidrule(lr){9-10}\cmidrule(lr){11-12}\cmidrule(lr){13-14}
				& &
				\textbf{Full} & \textbf{w/o Ph.4} &
				\textbf{Full} & \textbf{w/o Ph.4} &
				\textbf{Full} & \textbf{w/o Ph.4} &
				\textbf{Full} & \textbf{w/o Ph.4} &
				\textbf{Full} & \textbf{w/o Ph.4} &
				\textbf{Full} & \textbf{w/o Ph.4} &
				\textbf{Avg.\ \% Imp.} \\
				\midrule
				\multirow{16}{*}{\rotatebox{90}{\textbf{$R^{2}$}}}
				& california & 0.8168 & 0.8122 & 0.8110 & 0.8099 & 0.8298 & 0.7760 & 0.8080 & 0.7356 & 0.8312 & 0.8025 & 0.8307 & 0.7688 & -4.51 \\
				& compactive & 0.9847 & 0.9726 & 0.9657 & 0.9604 & 0.9719 & 0.9112 & 0.9612 & 0.8973 & 0.9759 & 0.9559 & 0.9723 & 0.9192 & -3.70 \\
				& cpu\_small & 0.9709 & 0.9634 & 0.9797 & 0.9667 & 0.9658 & 0.9391 & 0.9714 & 0.9480 & 0.9675 & 0.9536 & 0.9680 & 0.9385 & -1.96 \\
				& heat & 0.9892 & 0.9867 & 0.9952 & 0.9941 & 0.9963 & 0.9953 & 0.9963 & 0.9907 & 0.9519 & 0.9191 & 0.9992 & 0.9817 & -28.47 \\
				& wine\_quality & 0.4375 & 0.4220 & 0.4837 & 0.4744 & 0.4659 & 0.4443 & 0.4294 & 0.3457 & 0.4705 & 0.4621 & 0.4707 & 0.4651 & -5.43 \\
				& abalone & 0.5060 & 0.4928 & 0.4962 & 0.4867 & 0.5093 & 0.5060 & 0.3765 & 0.3002 & 0.4913 & 0.4825 & 0.5152 & 0.5143 & -4.57 \\
				& space\_ga & 0.7123 & 0.7118 & 0.6951 & 0.6910 & 0.7052 & 0.6870 & 0.6715 & 0.6413 & 0.7164 & 0.7130 & 0.7168 & 0.7123 & -1.47 \\
				& debutanizer & 0.7024 & 0.7019 & 0.6514 & 0.6435 & 0.6505 & 0.5639 & 0.7689 & 0.7546 & 0.6727 & 0.5688 & 0.6643 & 0.5869 & -7.26 \\
				& available\_power & 0.9706 & 0.9676 & 0.9642 & 0.9613 & 0.9755 & 0.9723 & 0.9625 & 0.8546 & 0.9671 & 0.9391 & 0.9756 & 0.9739 & -2.54 \\
				& maximal\_torque & 0.9646 & 0.9606 & 0.9598 & 0.9513 & 0.9816 & 0.9723 & 0.9026 & 0.8543 & 0.9343 & 0.9260 & 0.9796 & 0.9730 & -1.53 \\
				& fuel\_consumption\_country & 0.8849 & 0.8823 & 0.8997 & 0.8905 & 0.8827 & 0.8608 & 0.8297 & 0.8269 & 0.8997 & 0.8630 & 0.8870 & 0.8832 & -1.44 \\
				& acceleration & 0.8892 & 0.8838 & 0.9186 & 0.9105 & 0.8657 & 0.8490 & 0.8307 & 0.8202 & 0.8865 & 0.0813 & 0.8737 & 0.8712 & -15.97 \\
				& airfoild & 0.7963 & 0.8565 & 0.8413 & 0.6182 & 0.7800 & 0.4781 & 0.8622 & 0.7140 & 0.7487 & 0.4948 & 0.7783 & 0.5759 & -22.46 \\
				& mortgage & 0.9934 & 0.9897 & 0.9948 & 0.9943 & 0.9984 & 0.9980 & 0.9960 & 0.9928 & 0.9873 & 0.9862 & 0.9972 & 0.9971 & -0.15 \\
				& treasury & 0.9878 & 0.9791 & 0.9854 & 0.9843 & 0.9936 & 0.9935 & 0.9928 & 0.9894 & 0.9796 & 0.9708 & 0.9935 & 0.9911 & -0.41 \\
				& concreteStrength & 0.8751 & 0.8485 & 0.8364 & 0.8360 & 0.8112 & 0.6717 & 0.7879 & 0.7604 & 0.8316 & 0.8300 & 0.8244 & 0.8236 & -4.01 \\
				& \multicolumn{14}{r}{\textbf{Avg.\ -6.62}} \\
				\bottomrule
			\end{tabular}
		}
	\end{table}

	
	\bibliographystyle{elsarticle-harv} 
	\bibliography{bibliography}  
	
	
\end{document}